\documentclass[runningheads]{llncs}

\usepackage{eccv}

\usepackage[width=122mm,left=12mm,paperwidth=146mm,height=193mm,top=12mm,paperheight=217mm]{geometry}
\usepackage{eccvabbrv}

\usepackage{graphicx}
\usepackage{booktabs}
\usepackage{multirow}
\usepackage{tabularx}
\usepackage{adjustbox}
\usepackage{wrapfig}
\usepackage{makecell}
\usepackage{amsmath}
\usepackage{amssymb}
\usepackage{nicefrac}
\usepackage{xcolor,colortbl}
\definecolor{inchworm}{rgb}{0.7, 0.93, 0.36}
\definecolor{babyblueeyes}{rgb}{0.63, 0.79, 0.95}
\definecolor{columbiablue}{rgb}{0.61, 0.87, 1.0}
\usepackage[accsupp]{axessibility}  %
\usepackage{contour}
\contourlength{0.1pt} %
\usepackage{comment}

\usepackage{hyperref}

\usepackage{orcidlink}

\newcommand{\PAR}[1]{\vskip4pt \noindent{\bf #1~}}
\newlength{\tabcolsepdefault}
\newcommand{\verticalspace}{ \vspace*{-17pt} }
\newcommand{\verticalspacetable}{ \vspace*{-10pt} }
\newenvironment{tight_itemize}{
\begin{itemize}
  \setlength{\itemsep}{0pt}
  \setlength{\parskip}{0pt}
}{\end{itemize}}

\newcommand{\fs}{ \cellcolor{babyblueeyes}\bf }   %
\newcommand{\nd}{ \cellcolor{columbiablue}\underline }      %

\newcommand{\figurespace}{ \vspace*{-5pt} }
\begin{document}

\title{MaRINeR: Enhancing Novel Views by Matching Rendered Images with Nearby References}
\vspace{-25pt}
\titlerunning{MaRINeR: Enhancing Novel Views}

\author{Lukas Bösiger\inst{1}\orcidlink{0000-0001-6694-3560} \and
Mihai Dusmanu\inst{2}\orcidlink{0000-0002-3219-1783} \and
Marc Pollefeys\inst{1, 2}\orcidlink{0000-0003-2448-2318} \and
Zuria Bauer \inst{1}\orcidlink{0000-0001-8447-2344}
}

\authorrunning{L.~Bösiger et al.}

\institute{Department of Computer Science, ETH Zurich, Switzerland \and
Microsoft Mixed Reality \& AI Lab, Zurich, Switzerland \\
\url{https://boelukas.github.io/mariner/}}
\maketitle
\vspace*{-25pt}
\begin{abstract}
  Rendering realistic images from 3D reconstruction is an essential task of many Computer Vision and Robotics pipelines, notably for mixed-reality applications as well as training autonomous agents in simulated environments.
  However, the quality of novel views heavily depends of the source reconstruction which is often imperfect due to noisy or missing geometry and appearance.
  Inspired by the recent success of reference-based super-resolution networks, we propose MaRINeR, a refinement method that leverages information of a nearby mapping image to improve the rendering of a target viewpoint.
  We first establish matches between the raw rendered image of the scene geometry from the target viewpoint and the nearby reference based on deep features, followed by hierarchical detail transfer.
  We show improved renderings in quantitative metrics and qualitative examples from both explicit and implicit scene representations.
   We further employ our method on the downstream tasks of pseudo-ground-truth validation, synthetic data enhancement and detail recovery for renderings of reduced 3D reconstructions.

\end{abstract}
\vspace{-32pt}
\begin{figure}[!htb]
\vspace*{-5pt}
  \centering
  \includegraphics[height=4.7cm]{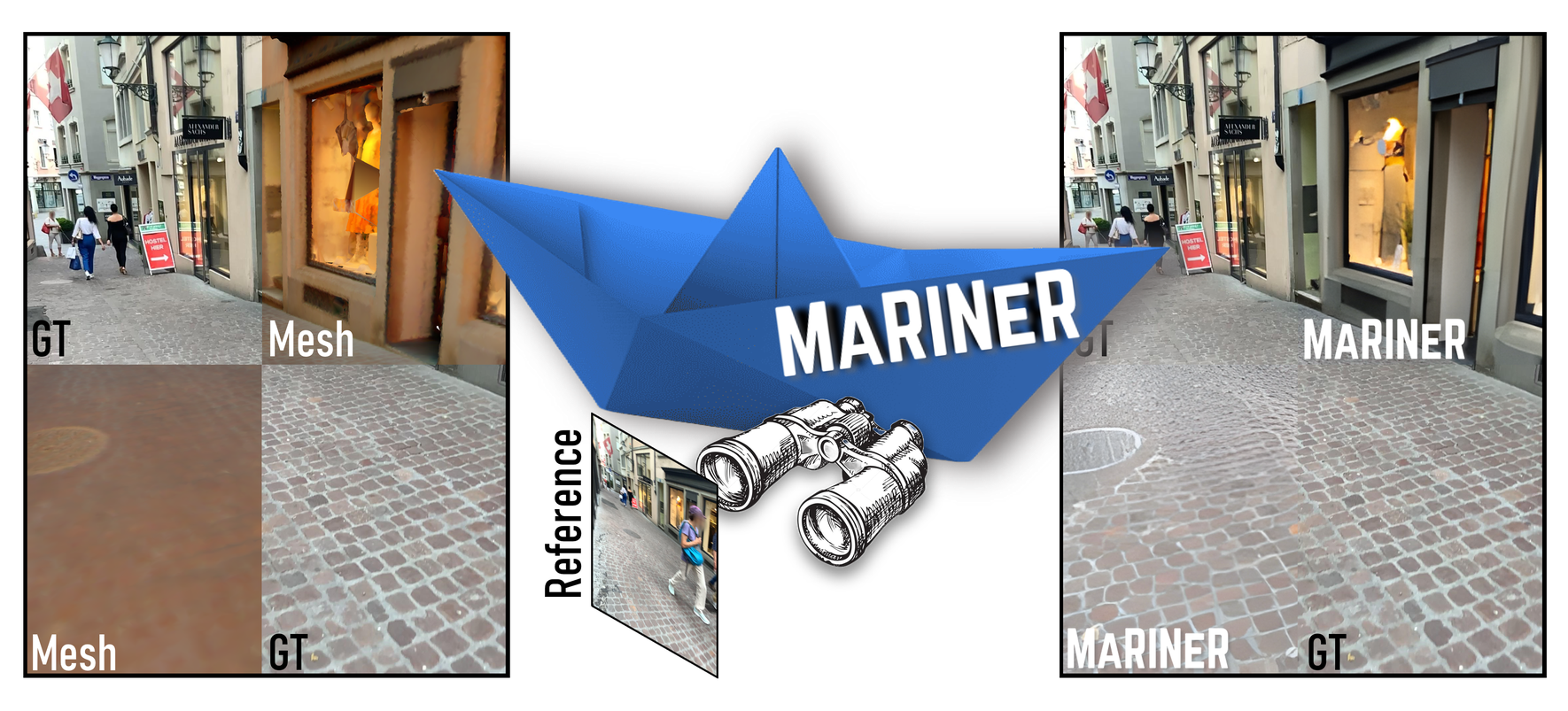}
  \vspace*{-5pt}
  \caption{We introduce \textbf{MaRINeR}: a pipeline taking as input a novel-view obtained from a 3D reconstruction exhibiting geometric and / or appearance artifacts and inaccuracies as well as a nearby reference used during the reconstruction process, and outputting an enhanced version of the novel-view through feature matching and transfer.}
  \label{fig:teaser}
  \vspace{-35pt}
\end{figure}

\section{Introduction}
\label{sec:intro}
\vspace{-5pt}

One of the fundamental problems of computer vision and robotics is reconstructing the environment from sensorial data such as color or depth cameras or LiDAR scanners.
These pipelines produce a computer-friendly representation of the space which can be either explicit (e.g., point-clouds, meshes), implicit (e.g, occupancy nets~\cite{OccupancyNets, ConvolutionalOccupancyNets}, neural radiance fields -- NeRF~\cite{NeRF}), or hybrid (e.g., Gaussian splats~\cite{GaussianSplats}) which serve as starting point for many subsequent tasks, notably novel-view synthesis, environment understanding, planning, and navigation.
All existing methods have limitations: point-clouds are highly dependent on the sensor quality~\cite{PointCloudSurvey}, often contain artifacts due to moving objects~\cite{PointCloudsMovingObjects}, and are not suitable for occlusion checking~\cite{PointCloudsVSMeshes} or pattern rendering.
Meshing algorithms often create both appearance and geometric artifacts and inconsistencies while connecting the vertices and coloring / texturing the polygons~\cite{segment3d, Matterport3D, Scannet, ETH3D_SLAM, ETH3D_STEREO, Habitat_Matterport}.
More modern implicit methods show exemplary rendering performance but often require very densely sampled frames or even depth maps which are not always available~\cite{NeRFLimitations}.
Furthermore, these methods also need extensive per-scene training. 
The performance decreases drastically as the frame-rate and the input modalities are reduced.
Any artifacts or inconsistencies produced by the reconstruction pipeline can lead to significant impact in downstream tasks. To create renderings of novel views without 3D scene representation, image based rendering methods, such as IBRnet~\cite{IBRnet}, learn to interpolate between existing views of the 3D scene. However, the novel view renderings can also contain artifacts such as blurry image parts or noisy geometry. While methods exist to remove such artifacts for a specific type of pipeline, such as NeRFLiX~\cite{NeRFLiX} for neural radiance fields, these techniques often lack the ability to remove artifacts produced by other types of pipelines.

To address these limitations, we propose a post-processing step of novel rendered views entitled \textbf{MaRINeR} by \textit{\textbf{Ma}tching the \textbf{R}endered \textbf{I}mages with \textbf{Ne}arby \textbf{R}eferences}. 
To this end, we make further use of input images to the reconstruction process as reference data.
Our task is strongly connected with Reference-based Super-Resolution (RefSR) since similar to renderings from low-quality or noisy 3D reconstructions, a naively up-scaled version of a low-resolution (low-res.) image lacks details.
RefSR methods use details present in a closely related high-resolution (high-res.) reference image to help super-resolve the low-res. image.
We notice that the methods used to match between low-res. and high-res. image domains for information transfer and fusion can more generally be used to transfer details from a reference to a related image of any nature.
However, the classically used CUFED5~\cite{SRNTT} dataset is not suitable for our task of novel view enhancement.
We therefore generate new training and test datasets building upon the recently released LaMAR~\cite{LaMAR} dataset.

\begin{figure}[!t]
\vspace*{-12pt}
  \centering
    \newcommand{\sz}{0.16}
    \newcommand{\hw}{56.5pt}
    \resizebox{\textwidth}{!}{
    \begin{tabular}{cccccccc}
        & \scriptsize Reference & \scriptsize Rendering & \scriptsize Result & \scriptsize Reference & \scriptsize Rendering & \scriptsize Result &
        \\
        \makecell[c]{a} &       
        \makecell[c]{\includegraphics[width=\sz\linewidth, height=\hw]{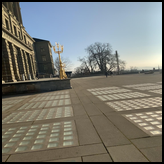}} &
        \makecell[c]{\includegraphics[width=\sz\linewidth, height=\hw]{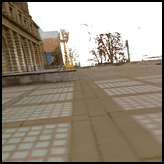}} &
        \makecell[c]{\includegraphics[width=\sz\linewidth, height=\hw]{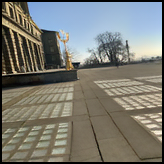}} &
        \hspace*{3pt}
        \makecell[c]{\includegraphics[width=\sz\linewidth, height=\hw]{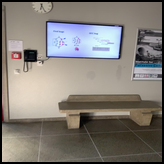}} &
        \makecell[c]{\includegraphics[width=\sz\linewidth, height=\hw]{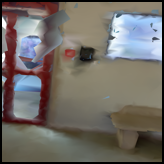}} &
        \makecell[c]{\includegraphics[width=\sz\linewidth, height=\hw]{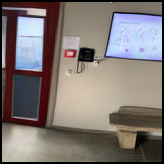}} &
        \makecell[c]{e}
        \\
        \makecell[c]{b} &
        \makecell[c]{\includegraphics[width=\sz\linewidth, height=\hw]{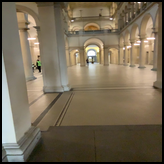}} &
        \makecell[c]{\includegraphics[width=\sz\linewidth, height=\hw]{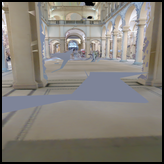}} &
        \makecell[c]{\includegraphics[width=\sz\linewidth, height=\hw]{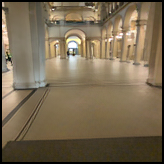}} &
        \hspace*{3pt}
        \makecell[c]{\includegraphics[width=\sz\linewidth, height=\hw]{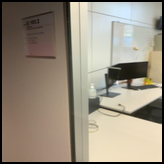}} &
        \makecell[c]{\includegraphics[width=\sz\linewidth, height=\hw]{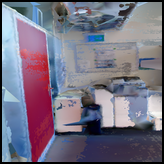}} &
        \makecell[c]{\includegraphics[width=\sz\linewidth, height=\hw]{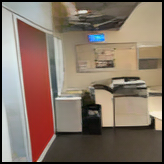}} &
        \makecell[c]{f}
        \\
        \makecell[c]{c} &
        \makecell[c]{\includegraphics[width=\sz\linewidth, height=\hw]{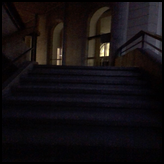}} &
        \makecell[c]{\includegraphics[width=\sz\linewidth, height=\hw]{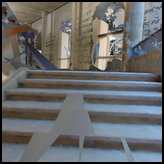}} &
        \makecell[c]{\includegraphics[width=\sz\linewidth, height=\hw]{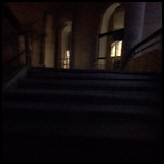}} &
        \hspace*{3pt}
        \makecell[c]{\includegraphics[width=\sz\linewidth, height=\hw]{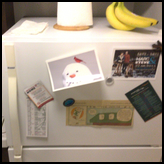}} &
        \makecell[c]{\includegraphics[width=\sz\linewidth, height=\hw]{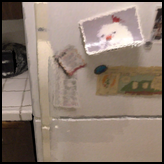}} &
        \makecell[c]{\includegraphics[width=\sz\linewidth, height=\hw]{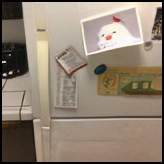}} &
        \makecell[c]{g}
        \\
        \makecell[c]{d} &
        \makecell[c]{\includegraphics[width=\sz\linewidth, height=\hw]{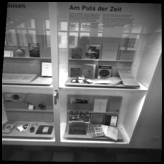}} &
        \makecell[c]{\includegraphics[width=\sz\linewidth, height=\hw]{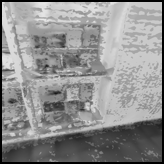}} &
        \makecell[c]{\includegraphics[width=\sz\linewidth, height=\hw]{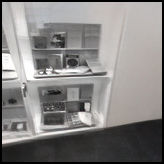}} &
        \hspace*{3pt}
        \makecell[c]{\includegraphics[width=\sz\linewidth, height=\hw]{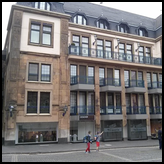}} &
        \makecell[c]{\includegraphics[width=\sz\linewidth, height=\hw]{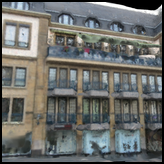}} &
        \makecell[c]{\includegraphics[width=\sz\linewidth, height=\hw]{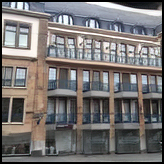}} &
        \makecell[c]{h}
        
    \end{tabular}}
    \vspace{-10pt}
  \caption{\textbf{Robustness of \textbf{MaRINeR}.} Our model recovers missing parts that appear due to rendering artifacts \textbf{a}, \textbf{b}. It adopts the illumination from the reference \textbf{c}, is device agnostic generalizing to gray-scale images \textbf{d}. The model enhances renderings of low triangle meshes \textbf{e} and also improves the rendering even if the reference has little content in common \textbf{f}. It can be applied to unseen scenes such as 12 Scenes~\cite{12scenes} \textbf{g} or Aachen Day-Night~\cite{AACHEN} \textbf{h} without retraining.}
  \label{fig:robustness}
  \verticalspace
\end{figure}

The enhanced novel views show promising results for different downstream tasks.
First, our method quantitatively and qualitatively narrows the gap between renderings and real images, including these of implicit representations. %
As a by-product, this improves the quality of data obtained when using digital twins for training reinforcement agents, following the recent advances in egocentric human synthetic data generation~\cite{Egogen}.
Second, many recent datasets (12 Scenes~\cite{12scenes}, RIO10~\cite{RIO10}, LaMAR~\cite{LaMAR}) designed pseudo-ground-truth pipelines to automatically and accurately register trajectories from various devices, notably in the context of mixed reality experiences.
One common way to validate the accuracy of these pipeline is through qualitative checks between the aligned images and an associated rendering from a 3D reconstruction.
Narrowing the render-to-real gap by removing the artifacts and improving the textural accuracy and realism opens the door to automating this process by taking advantage of existing geometric methods to estimate the accuracy of the ground-truth (GT).
Third, for any downstream application, decisions have to be made to reduce the size of 3D representations in order to efficiently process them at the cost of reduced detail accuracy and realism.
Our method is capable of recovering the lost details and realism of such 3D reconstructions as illustrated in \cref{fig:robustness}.
We will release the source code of our paper upon acceptance.

To summarize, our contributions are:
\begin{tight_itemize}
    \item We introduce \textbf{MaRINeR} which, to the best of our knowledge, is the first method enhancing novel views by using a close-by reference image that is applicable for renderings from a wide range of 3D reconstruction pipelines.
    \item An extensive evaluation of the proposed method is performed, providing not only a qualitative and quantitative analysis of the method but also an overview of the robustness of \textbf{MaRINeR} to different datasets, temporal conditions and temporal changes in scenes, while discussing limitations. 
    \item We showcase the excellent performance of our model in several applications: elimination of manual checks in pseudo-GT pipelines, improvement of synthetic AR trajectories, and enhancing the output of neural renderings.
\end{tight_itemize}

\section{Related work}
We provide an overview of related research fields which also incorporate information from a reference image into a target image, notably reference-based image super-resolution and style transfer.
\PAR{Reference-based image Super-Resolution (RefSR).} The goal of RefSR is to recover high-res. from low-res. images by transferring missing details from high-res. reference images. The methods usually work by aligning and fusing features extracted from low-res. and ref. images. While early work uses hand crafted features~\cite{Landmark}, more recent works use either pre-trained features~\cite{SRNTT, WTRNTIP} or train the feature extraction end-to-end with the task~\cite{SSNET, CrossNet, E2ENT, SSEN, TTSR, c2matching, MASASR, RRSGAN}. The alignment of the ref. and low-res. features proposes a challenge because of the resolution difference. Some methods use implicit alignment: CrossNet~\cite{CrossNet} estimates the optical flow between ref. and low-res. images. Because optical flow fails at capturing long distance correspondences, SSEN~\cite{SSEN} utilizes deformable convolutions which ensure a large receptive field. Other work uses explicit alignment by feature or patch matching. SRNTT~\cite{SRNTT} uses feature similarity and transfers textures from the ref. images at different scales. To reduce the computational complexity, MASA~\cite{MASASR} proposes a coarse-to-fine correspondence matching module. $C^2$-Matching~\cite{c2matching} introduces knowledge distillation and contrastive learning methods to improve the matching between low-res. and ref. despite the resolution gap. WTRN~\cite{WTRNTIP} uses wavelets to separate high and low frequency parts of the images, which helps to more transfer more visually plausible texture patterns. DATSR~\cite{DATSR} uses Swin-Transformers~\cite{SwinTransformer} to replace the commonly used residual blocks~\cite{ResBlock}, leading to more robust matches and texture transfer. HMCF~\cite{HMCF} improves the matching between low-res. and ref. of similar objects with different texture by using high-to-low-level feature matching and complementary information fusion. RRSGAN~\cite{RRSGAN} uses generative adversarial networks and deformable convolutions. FRFSR~\cite{FRFSR} notes that the commonly used perceptual and adversarial loss have an adverse effect on texture transfer and reconstruction. As a solution, they propose the use of a texture reuse framework. RRSR~\cite{RRSR} uses a reciprocal learning strategy to strengthen the training process by using the super resolution result as reference to help super-resolve a low-res. variant of the original high-res. reference. CMRSR~\cite{CMRSR} notes that due to the gap between inputs and reference, the super resolution image often yields distortions and ghosting artifacts and they propose a contrastive attention-guided multi-level feature registration module to mitigate those. There are also methods that use multiple references as input such as CIMR-SR~\cite{CIMRSR}, AMRSR~\cite{AMRSR}, AMSA~\cite{AMSA} or LMR~\cite{LMR}. We notice that many of ideas to align ref. and low-res. features are not limited to align images with resolution differences but can be used more broadly to also align rendered images to real images. 
\PAR{Style Transfer (ST).}
Artistic style-transfer methods transfer the style of a style image to a content image. A subcategory is the universal ST methods~\cite{WCT, AdaIN, ArtFlow}, which transfer any style to the content image. This can be done by separating content and style information in the images. AdaIN~\cite{AdaIN} transfers channel-wise mean and variance feature statistics. WCT~\cite{WCT} uses whitening and coloring transformations, where the whitening transformation can remove the style of the content image and the coloring transformation can incorporate the style of the style image. However the separation of content and style is challenging and some content can be corrupted. ArtFlow~\cite{ArtFlow} calls this issue content leak and introduces a reversible neural flow-based network to avoid it. StyTr2~\cite{StyTr2} uses transformers to extract and maintain global image information, which then help with the content leak problem. For universal ST, the style images usually have little content in common with the content image. The results look like an artistic version of the content image which is however far from being realistic. Semantic ST methods~\cite{NNST, MST} work with style images that contain similar objects as the content image. The goal is to build semantic correspondences between similar objects and map the style region only to the semantically similar content regions~\cite{ReviewStyleTransfer}. NNST~\cite{NNST} matches VGG~\cite{VGG} features between content and style and replaces the content features with the nearest style features. MST~\cite{MST} uses graph cuts for matching between content and style features. While those methods work well at transferring the semantic correspondences, they can introduce distortions and don't produce photo-realistic images. Photo-realistic ST methods aim at transferring the style of the color distribution while preserving the structures of the content image~\cite{ReviewStyleTransfer}. WCT2~\cite{WCT2} adds a wavelet based correction to the whitening and coloring transforms of WCT~\cite{WCT}. This helps to preserve the structural and statistical properties of the VGG features during stylization. The result is a more photo-realistic image without distortions. However, photo-realistic ST assumes that the content image is already photo-realistic. If this image contains artifacts, then those are also carried over to the stylized image.

\section{Method}
\begin{figure}[tb]
    \vspace*{-21pt}
  \centering
  \includegraphics[width=0.9\linewidth]{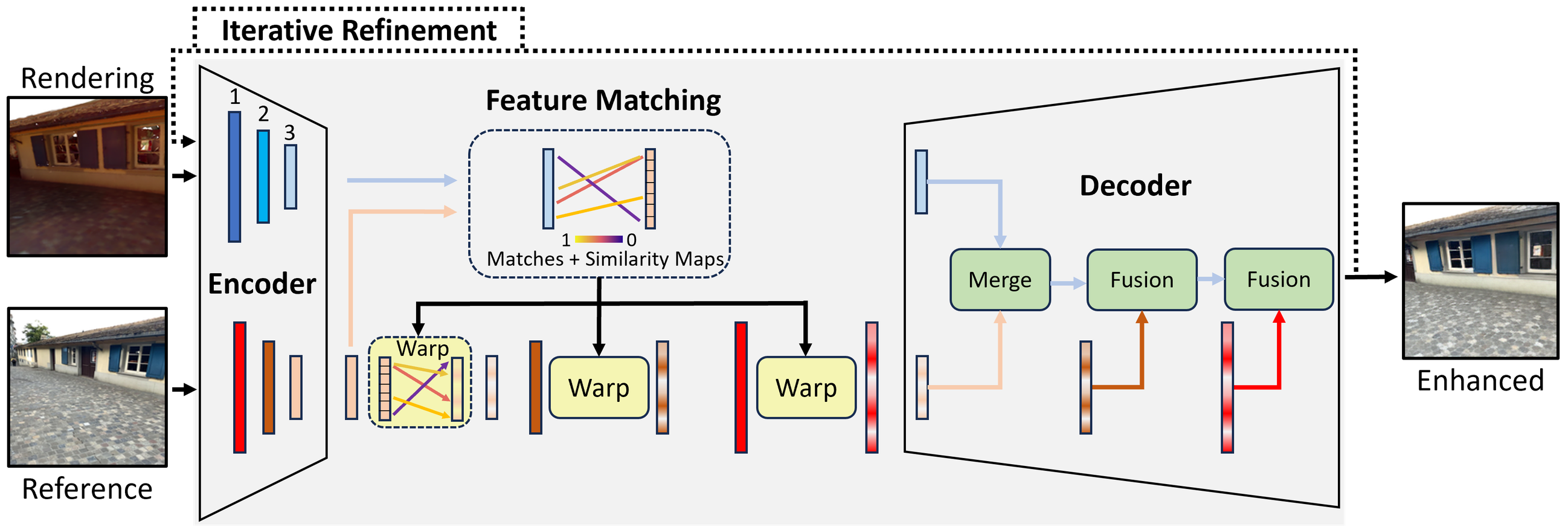}
  \vspace*{-10pt}
  \caption{\textbf{MaRINeR architecture.} The learned features of the encoder are used to for correspondence matching and warping of the reference features. They are fused with the rendering features to create a enhanced rendering, which is iteratively refined.}
  \label{fig:pipline}
  \verticalspace
\end{figure}
The \textbf{MaRINeR} pipeline, illustrated in~\cref{fig:pipline}, takes as input a rendering with noisy appearance and geometry as well as a nearby reference and outputs an enhanced version of the rendering by transferring relevant information from the reference.
The pipeline starts by densely extracting features at multiple levels from both input images with a shared convolutional encoder.
Next, the deepest features extracted from the rendering are matched to those of the reference to retrieve similar content.
These matches are then used to warp the reference features at different levels.
The warped reference features are fused with those of the rendering in the decoder.
Given the severe artifacts sometimes present in novel views, we employ an iterative refinement approach that repeats the process by replacing the input rendering with the enhanced output of the previous iteration.
We start from the MASA~RefSR~\cite{MASASR} pipeline and implement several changes in architecture, loss function as well as data augmentation to make it amenable to the novel task of reference-based rendering enhancement.

\PAR{Encoder.}
As mentioned above, we use a shared convolutional encoder to extract features at multiple levels from both the rendered image $I$ and the reference $R$, for simplicity assumed both of size $H \times W$.
We use three levels, each halving the resolution of the previous one, yielding two sets of dense tensors: $\{\mathcal{F}^I_1 \in \mathbb{R}^{H \times W \times F_1}, \mathcal{F}^I_2 \in \mathbb{R}^{\nicefrac{H}{2} \times \nicefrac{W}{2} \times F_2}, \mathcal{F}^I_3 \in \mathbb{R}^{\nicefrac{H}{4} \times \nicefrac{W}{4} \times F_3}\}$ and $\{\mathcal{F}^R_1, \mathcal{F}^R_2, \mathcal{F}^R_3\}$ for the rendering and the reference, respectively, where $F_i$ is the number of channels of the output of level $i$. %
These features will next be used to find corresponding patches between the rendering and the reference in a coarse-to-fine fashion.

\PAR{Feature matching.}
We use the Matching and Extraction Module (MEM) from MASA~\cite{MASASR} to match the deepest features of both input images, $\mathcal{F}^R_3$ and $\mathcal{F}^I_3$, using cosine similarity.
The MEM performs matching first on a coarse grid with a stride and then densely within a fixed-size window around the resulting matches.
This step yields a mapping $m$ of indices from the level 3 features of the rendering to those of the reference and associated matching scores $s$:
\figurespace
\begin{equation}
    m_{I \rightarrow R}: (x, y) \in \mathcal{F}^I_3 \rightarrow \{ (u, v) \in \mathcal{F}^R_3, s \in \mathbb{R} \} \enspace .
\figurespace
\end{equation}
This mapping is used to warp and weight the reference features at each of the three levels $i$, where blocks of features with size relative to the spatial resolution of the current level are cropped and moved together resulting in warped feature maps $\{\mathcal{F}^{R \rightarrow I}_1, \mathcal{F}^{R \rightarrow I}_2, \mathcal{F}^{R \rightarrow I}_3\}$.
In contrast to RefSR methods which have an input with lower resolution and thus need to perform the matching on the $\mathcal{F}_1$ features of the low-res. input and down-scaled reference~\cite{MASASR}, we use deeper features, allowing us to leverage the increased robustness to find better quality matches.
Weighting the warped features based on the matching scores reduces the impact of features with low confidence matches.
This enables the model to only use the reference features if they have a confident match and otherwise use the rendering features when fusing them in the decoder.

\PAR{Decoder.}
\begin{figure}[tb]
\vspace*{-12pt}
  \centering
  \includegraphics[width=\linewidth]{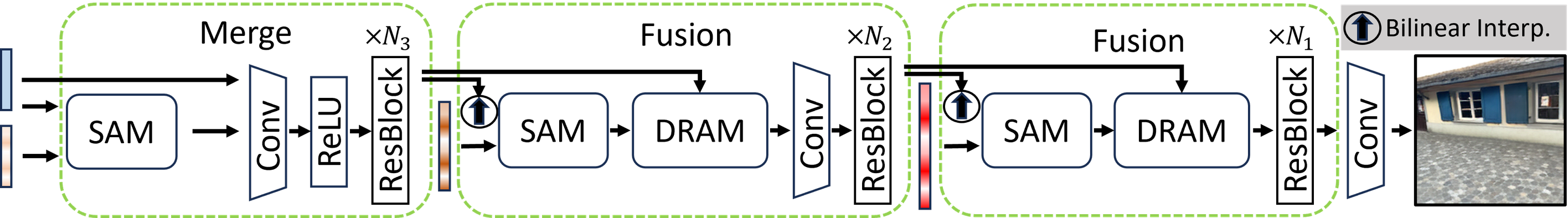}
  \vspace*{-17pt}
  \caption{\textbf{Architecture of the decoder.} We fuse the rendering and warped reference features using SAM~\cite{MASASR}, DRAM~\cite{MASASR} and residual blocks~\cite{ResBlock}.}
  \label{fig:decoder}
  \verticalspace
\end{figure}
Using the deepest features of the rendering $\mathcal{F}^I_3$ and the warped reference features $\mathcal{F}^{R \rightarrow I}_3$, $\mathcal{F}^{R \rightarrow I}_2$ and $\mathcal{F}^{R \rightarrow I}_1$, we fuse them using Spatial Adaptation Modules (SAM)~\cite{MASASR}, Dual Residual Aggregation Modules (DRAM)~\cite{MASASR} and residual blocks~\cite{ResBlock}, as shown in~\cref{fig:decoder}. SAM learns to remap the distribution of the reference features to the one of the rendering features. DRAM fuses features of different spatial resolution aiming to refine and aggregate the details of both branches and up-sample the low-res. features with a transposed convolution. The decoder procedure can be summarized as follows:
\figurespace
\begin{equation}
\figurespace
\begin{split}
        \mathcal{O}_3 & = \text{P}_3(\text{SAM}(\mathcal{F}^I_3, F^{R \rightarrow I}_3) \oplus \mathcal{F}^I_3) \\
        \mathcal{O}_2  & = \text{P}_2(\text{DRAM}(\text{SAM}(\mathcal{O}_3^{\uparrow}, \mathcal{F}^{R \rightarrow I}_2), \mathcal{O}_3)) \\
        \mathcal{O}_1  & = \text{P}_1(\text{DRAM}(\text{SAM}(\mathcal{O}_2^{\uparrow}, \mathcal{F}^{R \rightarrow I}_1), \mathcal{O}_2))
        \label{eq:decoder}
\end{split}
\figurespace
\end{equation}
where $\text{P}_i$ stands for processing the features using a convolution and $N_i$ residual blocks~\cite{ResBlock}, $\oplus$ for a convolution to merge the features and $\uparrow$ for bilinear interpolation. We merge the level 3 rendering features $\mathcal{F}^I_3$ with the warped level 3 reference features $\mathcal{F}^{R \rightarrow I}_3$ by concatenating and processing them using a convolution, leveraging that the rendering and warped reference features are of similar spatial resolution.
The SAM aligns the rendering and warped reference feature distributions such that the DRAM can successfully merge the features.
The output image is then created from $O_1$ using a convolution to reduce the feature dimension.
In contrast, RefSR methods such as MASA deal with features with a different spatial resolution that can not directly be merged. MASA first fuses the low resolution features of level 3 to 1 together before merging the reference features. For the task of RefSR this is beneficial because the result is encouraged to be structurally similar to the input with only additional details from the reference. For the task of rendering enhancement where the rendering can contain structures that come from artifacts, merging the features of similar spatial resolution enables the model to also take structural information from the reference. This is beneficial to remove rendering artifacts or fill in missing image parts caused by gaps in the source 3D reconstruction.

\PAR{Iterative refinement.}
Since the gap between the rendering and the reference can be large due to rendering artifacts that occlude the underlying geometry, we found it beneficial to apply the model several times in an iterative fashion.
The first iteration can thus focus on removing artifacts and enhancing the image.
The following iterations are then more successful in establishing correspondences and transferring the missing details to the enhanced rendering.
To this end, we supervise the model after each iteration, thus obtaining a more general model that can deal with a wide variety of rendering qualities.

\subsection{Loss function}
Our goal is to preserve the spatial information of the rendering while removing artifacts, adding details from the reference, and producing a visually pleasing result. To this end, we combine a reconstruction loss, perceptual loss, and adversarial loss with associated weights $\lambda_{\text{rec}}$, $\lambda_{\text{per}}$, and $\lambda_{\text{adv}}$, written as:
\figurespace
\begin{equation}
        \mathcal{L} = \lambda_{\text{rec}}\mathcal{L}_{\text{rec}} + \lambda_{\text{per}}\mathcal{L}_{\text{per}} + \lambda_{\text{adv}}\mathcal{L}_{\text{adv}} \enspace .
        \label{eq:per_loss} 
\figurespace
\end{equation}

\PAR{Reconstruction loss.}
The enhanced rendering $I_{\text{ER}}$ should be close to the GT image taken at the same pose as the rendering by using the information present in the close-by reference. We adopt the following reconstruction loss:
\figurespace
\begin{equation}
        \mathcal{L}_{\text{rec}} = \|\mathbf{I}_{\text{GT}} - \mathbf{I}_{\text{ER}}\|_1 \enspace ,
        \label{eq:rec_loss} 
\figurespace
\figurespace
\end{equation}
where $\|\cdot\|_1$ is the $\ell_1$ norm.

\PAR{Perceptual loss.}
The perceptual loss is widely used by RefSR models~\cite{MASASR, DATSR, FRFSR, c2matching} to enhance the visual quality of the result by guiding the resulting image to be more semantically similar to the GT. This loss is formulated as:
\figurespace
\figurespace
\begin{equation}
        \mathcal{L}_{\text{per}} = \frac{1}{3}\underset{i=1}{\overset{3}{\sum}}\|\phi_i(\mathbf{I}_{\text{GT}}) - \phi_i(\mathbf{I}_{\text{ER}})\|_2^2 \enspace ,
        \label{eq:per_loss} 
\figurespace
\figurespace
\end{equation}
where $\phi_i(\cdot)$ denotes the outputs of ImageNet~\cite{ImageNet}-pretrained VGG16~\cite{VGG} at layers \texttt{relu1\_1}, \texttt{relu2\_2} and \texttt{relu3\_3}. Contrary to RefSR methods, we chose to use more shallow features~\cite{perceptualLossCoarseNet} since the domain gap between rendering and reference leads to a mismatch between the deeper features and therefore causes increased artifact generation. We show qualitative results in the supplementary. %
 
\PAR{Adversarial loss.}
The drawback of the perceptual loss is that it tends to generate grid like artifacts~\cite{perceptualLossartifacts}. The adversarial loss~\cite{advLoss} helps to remove those artifacts and generate visually pleasing images:
\figurespace
\begin{equation}
    \mathcal{L}_{\text{disc}} = - \mathbb{E}_{\mathbf{I}_{\text{GT}}}[\log(D(\mathbf{I}_{\text{GT}},\mathbf{I}_{\text{ER}}))] - \mathbb{E}_{\mathbf{I}_{\text{ER}}}[\log(1-D(\mathbf{I}_{\text{ER}}, \mathbf{I}_{\text{GT}}))] \enspace ,
    \label{eq:adv_d_loss}
\end{equation}
\vspace{-15pt}
\begin{equation}
    \mathcal{L}_{\text{adv}} = - \mathbb{E}_{\mathbf{I}_{\text{GT}}}[\log(1-D(\mathbf{I}_{\text{GT}}, \mathbf{I}_{\text{ER}})) - \mathbb{E}_{\mathbf{I}_{\text{ER}}}[\log(D(\mathbf{I}_{\text{ER}}, \mathbf{I}_{\text{GT}}))]
    \label{eq:adv_g_loss} \enspace ,
\end{equation}
where $\mathcal{L}_{\text{disc}}$ represents the discriminator loss and ${L}_{\text{adv}}$ is the generator loss. We adopted the Relativistic GAN~\cite{relativisticGAN} formulation following MASA~\cite{MASASR}.

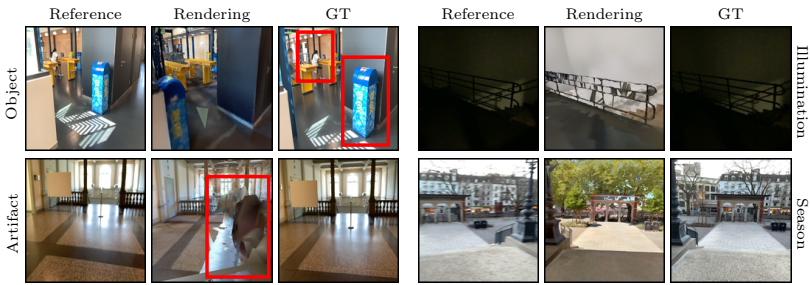
\begin{figure}[!t]
\vspace*{-10pt}
  \centering
    \newcommand{\sz}{0.16}
    \newcommand{\hw}{56.5pt}
    \centering
    \resizebox{0.9\textwidth}{!}{
    \begin{tabular}{cccccccc}
          & \scriptsize Reference & \scriptsize Rendering & \scriptsize GT & \hspace*{3pt} \scriptsize Reference & \scriptsize Rendering & \scriptsize GT & 
        \\
        \rotatebox[origin=c]{90}{\scriptsize Object} & \makecell{\begin{tikzpicture}
        \node[anchor=south west,inner sep=0] at (0,0) {\includegraphics[width=\sz\linewidth, height=\hw]{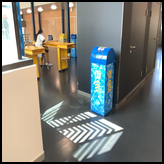}};
        \end{tikzpicture}} &
        \makecell{\includegraphics[width=\sz\linewidth, height=\hw]{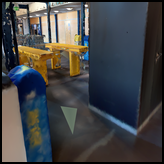}} &
        \makecell{\begin{tikzpicture}
        \node[anchor=south west,inner sep=0] at (0,0) {\includegraphics[width=\sz\linewidth, height=\hw]{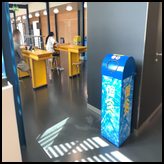} };
        \draw[red,ultra thick] (1.79,1.5) rectangle (1.05,0.1);
        \draw[red,ultra thick] (0.89,1.12) rectangle (0.32,1.9);
        \end{tikzpicture}} &
        \hspace*{3pt}
        \makecell{\includegraphics[width=\sz\linewidth, height=\hw]{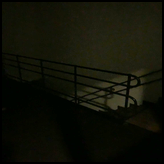}} &
        \makecell{\includegraphics[width=\sz\linewidth, height=\hw]{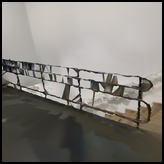}} &
        \makecell{\includegraphics[width=\sz\linewidth, height=\hw]{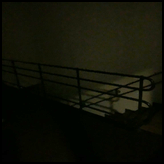}} &
        \rotatebox[origin=c]{270}{\scriptsize Illumination}
        \\   
        \rotatebox[origin=c]{90}{\scriptsize Artifact} & \makecell{\includegraphics[width=\sz\linewidth, height=\hw]{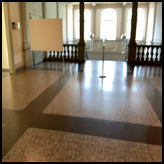}} &
        \makecell{\begin{tikzpicture}
        \node[anchor=south west,inner sep=0] at (0,0) {\includegraphics[width=\sz\linewidth, height=\hw]{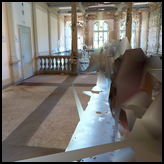} };
        \draw[red,ultra thick] (1.9,1.7) rectangle (0.9,0.1);
        \end{tikzpicture}} &
        \makecell{\includegraphics[width=\sz\linewidth, height=\hw]{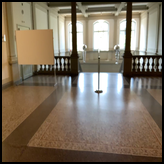}} &
        \hspace*{3pt}
        \makecell{\includegraphics[width=\sz\linewidth, height=\hw]{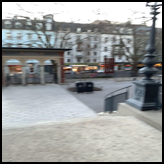}} &
        \makecell{\includegraphics[width=\sz\linewidth, height=\hw]{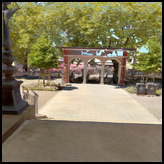}} &
        \makecell{\includegraphics[width=\sz\linewidth, height=\hw]{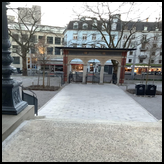}} &
        \rotatebox[origin=c]{270}{\scriptsize Season} 
        \\
    \end{tabular}
    }
    \vspace*{-11pt}
  \caption{\textbf{Common dataset challenges}. There can be different objects present between rendering and GT, some of which can be artifacts. The illumination can also be different because of day time or seasonal changes.}
  \label{fig:dataset}
  \verticalspace
\end{figure}
\vspace{-10pt}
\section{Experiments}
\vspace{-10pt}
\PAR{Datasets.}
We use the recently introduced LaMAR dataset~\cite{LaMAR} to create training and test datasets. LaMAR~\cite{LaMAR} consists of scenes represented by 3D scans and localized AR device trajectories within those scenes. The devices used were iPhones/iPads and HoloLens 2. We use the RGB trajectories from the iPhones and iPads. The dataset consists of three different scenes: CAB, LIN and HGE. CAB is a multi-floor office building, LIN is a few blocks of an old town and HGE the ground floor of a historical university building. We create a training set from CAB and LIN, where CAB contains mostly indoor and LIN outdoor images. Test sets are created from CAB, LIN and HGE. HGE is used to test generalization to novel scenes. For this we take the trajectories, which are a sequence of RGB image and camera pose pairs. The poses are used to render an image from the 3D scan, which we use as the input rendering to be enhanced. The RGB image is used as the GT. As reference we use a nearby image with a different pose than the GT. Example pairs can be seen in \cref{fig:dataset}. The renderings can contain artifacts due to the scan quality. Because the trajectories and the 3D scan were not recorded at the same time, different objects, illumination and seasonal changes can be present within the rendering and the GT. The CAB and LIN training set contains 21350 image pairs. The CAB, LIN and HGE test sets consist of 329, 608 and 492 image pairs. The datasets contain different references of various levels for each rendering: a low level indicates that the reference pose is close to the GT pose (easier) and a high level indicates that the reference is further away (harder). Because RefSR methods usually train on the CUFED~\cite{SRNTT} dataset which consist of images with resolution 160x160, we also rescale our dataset images to this resolution.

\PAR{Implementation details.}
The encoder consists of 3 levels where each level is connected to the next one and consists of 1 convolutional layer and 4 residual blocks~\cite{ResBlock}.
In our experiments, we keep the number of feature channels fixed to $F_i = 64$ for all levels.
We train on 160x160 images following the convention of recent RefSR methods~\cite{MASASR, DATSR, FRFSR}. In the decoder we use $N_3 = 12$, $N_2 = 8$ and $N_1 = 4$ residual blocks in the merge and fusion layers.
We train our model for 60 epochs using only the reconstruction and perceptual loss and fine-tune the model for 20 epochs using additionally the adversarial loss. In our experiments, the weight coefficients $\lambda_{\text{rec}}$, $\lambda_{\text{per}}$ and $\lambda_{\text{adv}}$ are 1, 1 and 0.001. For training we use a NVIDIA Tesla A100 40GB GPU with a batch size of 9 for 37 hours.
We use two data-augmentation strategies specifically targeted to our task.
For generalization to a wide range of rendering qualities, we augment the training data with renderings from down-sampled versions of the meshes containing only $10\%$ of the original triangles.
To ensure that the model removes artifacts and enhances the rendering even if the reference image is far away or has little content in common, we pick the training reference images randomly from within a 5s temporal window in the sequence.

\PAR{Evaluation metrics.}
Because we start from a RefSR method, we use the same metrics for evaluation, notably: Peak Signal Noise Ration (PSNR~$\uparrow$) and Structural Similarity Index Measure (SSIM~$\uparrow$).
To follow the convention~\cite{MASASR, DATSR}, all PSNR and SSIM results are evaluated on the Y channel of the YCbCr color space. Because PSNR and SSIM can not determine visual quality we also report the Learned Perceptual Image Patch Similarity (LPIPS~$\downarrow$)~\cite{LPIPS} and the Edge Restoration Quality Assessment (ERQA~$\uparrow$)~\cite{ERQA}. LPIPS represents the visual quality with respect to the human perception. Because PSNR and SSIM do not align with human perception when it comes to value blurry images against images with details~\cite{LPIPS}, we also use ERQA that measures how well a method performs at restoring edge details.

 \setlength{\tabcolsep}{1.0pt}
    \begin{table}[!t]
    \verticalspacetable
      \caption{\textbf{Quantitative evaluation.} Our model enhances the rendering in common image quality metrics. It does so using the optimal reference $\text{Ref.}=\text{GT}$ as an upper bound or using a close-by reference. It generalizes to the unseen HGE scene and performs better than existing RefSR (RSR) and style transfer (ST) methods.
      }
      \label{tab:quantitaive}
      \centering
      \verticalspacetable
      \tiny
      \resizebox{\textwidth}{!}{
      \begin{tabular}{c|c|cccc|cccc|cccc|cccc}
        \toprule
        &\multirow{2.5}{*}{Method} & \multicolumn{4}{c|}{$\text{CAB}_\text{Ref.=GT}$} & \multicolumn{4}{c|}{CAB} & \multicolumn{4}{c|}{LIN} & \multicolumn{4}{c}{HGE}\\ \cmidrule{3-18}
        &                                                                        & PSNR  & SSIM  & ERQA & LPIPS     & PSNR  & SSIM  & ERQA & LPIPS      & PSNR  & SSIM  & ERQA & LPIPS      & PSNR  & SSIM  & ERQA & LPIPS \\
        \midrule
                                                        & Render                 & 15.60 & 0.559 & 0.564 & 0.380    & 15.60 & \nd{0.559} & 0.564 & 0.380     & 14.39 & 0.529 & 0.549 & 0.392     & 15.84 & 0.575 & 0.619 & 0.364\\
        \midrule
        \multirow{2}{*}{\rotatebox[origin=c]{90}{RSR}}  & MASA~\cite{MASASR}     & 15.55 & 0.555 & 0.568 & 0.347    & 15.47 & 0.524 & 0.544 & 0.367     & 14.17 & 0.419 & 0.523 & 0.397     & 15.62 & 0.478 & 0.576 & 0.360\\ 
                                                        & DATSR~\cite{DATSR}     & 15.65 & 0.568 & 0.553 & 0.349    & 15.63 & 0.557 & 0.530 & \nd{0.364}     & 14.34 & 0.468 & 0.483 & 0.438     & 15.80 & 0.536 & 0.553 & 0.376\\
        \midrule
        \multirow{3}{*}{\rotatebox[origin=c]{90}{ST}}   & Artflow~\cite{ArtFlow} & 16.30 & 0.472 & 0.533 & 0.334    & 15.39 & 0.414 & 0.489 & 0.393     & 17.00 & 0.503 & 0.622 & 0.321     & 16.97 & 0.500 & 0.602 & 0.341\\
                                                        & WCT2~\cite{WCT2}       & 16.48 & 0.559 & 0.569 & 0.357    & 16.08 & 0.554 & \nd{0.567} & 0.367     & 17.44 & \nd{0.569} & 0.565 & 0.332     & 17.52 & 0.589 & 0.623  & 0.324\\
                                                        & NNST~\cite{NNST}       & \nd{18.52} & \nd{0.643} & \nd{0.620} & \nd{0.303}    & \nd{16.33} & \nd{0.559} & 0.566 & 0.370     & \nd{18.53} & 0.568 & \nd{0.661} & \nd{0.315}     & \nd{18.48} & \nd{0.591} & \nd{0.646} & \nd{0.315}\\
        \midrule
         \rotatebox[origin=c]{90}{RE}                   
                                                        & \textbf{MaRINeR} & \fs{23.89} & \fs{0.799} & \fs{0.722} & \fs{0.089}       & \fs{20.03} & \fs{0.697} & \fs{0.643} & \fs{0.180}     & \fs{21.73} & \fs{0.668} & \fs{0.705} & \fs{0.155}     & \fs{20.96} & \fs{0.673} & \fs{0.684}& \fs{0.176} \\
      \bottomrule
      \end{tabular}}
      \verticalspacetable
      \verticalspacetable
    \end{table}
    \setlength{\tabcolsep}{\tabcolsepdefault}
\vspace{-15pt}
\subsection{Comparison with RefSR and ST methods}
\vspace{-10pt}
We conduct quantitative and qualitative comparisons between our method and existing RefSR and ST methods. The RefSR methods we compare with are MASA~\cite{MASASR} and DATSR~\cite{DATSR}. The ST methods are the universal method Artflow~\cite{ArtFlow}, the photorealistic method WCT2~\cite{WCT2} and the semantic method NNST~\cite{NNST}. \cref{tab:quantitaive} shows the results of the quantitative comparison. The metrics are calculated between the GT and the enhanced rendering. The row \textit{Render} is the baseline and shows the scores of the not enhanced rendering and the GT. In the column $\text{CAB}_{\text{Ref.=GT}}$ the GT is used as the reference showing the methods' performance when using the optimal reference image. MaRINeR successfully enhances the rendering leading to a significant improvement in all metrics. It also performs better at the task than existing RefSR and ST methods. Even though HGE is a novel scene, our model performs similarly well as on the CAB and LIN scenes, demonstrating that our model exhibits a strong generalization ability. Because in our dataset different objects can be present in rendering and GT, the enhanced rendering will not necessarily exactly look like the GT. This is also reflected in the scores, which are generally lower then when comparing RefSR methods where the GT and low-res. match content-wise.

\PAR{Qualitative comparison.}
\begin{figure}[!t]
\verticalspacetable
  \centering
    \newcommand{\sz}{0.14}
    \newcommand{\hw}{50.5pt}
    \renewcommand\fbox{\fcolorbox{black}{black}}
    \setlength{\fboxsep}{0.4pt} 
    \setlength{\fboxrule}{0.4pt}
    \vspace{-2pt}
    \resizebox{\textwidth}{!}{
    \begin{tabular}{ccccccc}
        \scriptsize Reference & \scriptsize Rendering & \scriptsize MASA & \scriptsize DATSR & \scriptsize NNST & \scriptsize \textbf{MaRINeR} & \scriptsize GT
        \\  %
        \fbox{\includegraphics[width=\sz\linewidth, height=\hw]{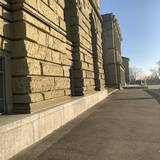}} &
        \fbox{\includegraphics[width=\sz\linewidth, height=\hw]{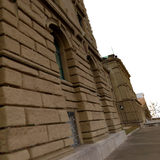}} &
        \fbox{\includegraphics[width=\sz\linewidth, height=\hw]{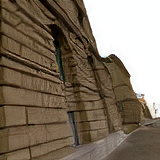}} &
        \fbox{\includegraphics[width=\sz\linewidth, height=\hw]{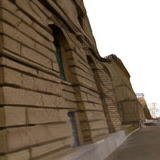}} &
        \fbox{\includegraphics[width=\sz\linewidth, height=\hw]{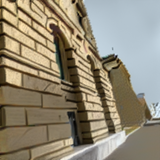}} &
        \fbox{\includegraphics[width=\sz\linewidth, height=\hw]{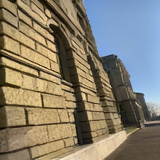}} &
        \fbox{\includegraphics[width=\sz\linewidth, height=\hw]{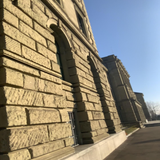}} 
        \\  %
        \fbox{\includegraphics[width=\sz\linewidth, height=\hw]{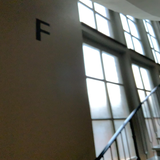}} &
        \fbox{\includegraphics[width=\sz\linewidth, height=\hw]{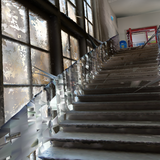}} &
        \fbox{\includegraphics[width=\sz\linewidth, height=\hw]{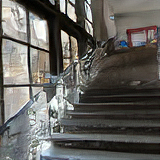}} &
        \fbox{\includegraphics[width=\sz\linewidth, height=\hw]{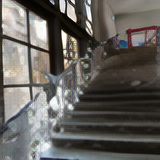}} &
        \fbox{\includegraphics[width=\sz\linewidth, height=\hw]{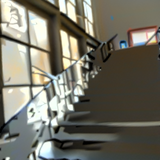}} &
        \fbox{\includegraphics[width=\sz\linewidth, height=\hw]{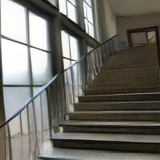}} &
        \fbox{\includegraphics[width=\sz\linewidth, height=\hw]{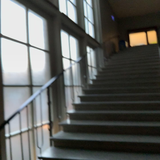}}
        \\  %
        \fbox{\includegraphics[width=\sz\linewidth, height=\hw]{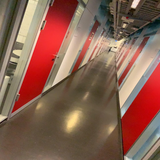}} &
        \fbox{\includegraphics[width=\sz\linewidth, height=\hw]{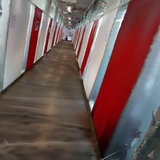}} &
        \fbox{\includegraphics[width=\sz\linewidth, height=\hw]{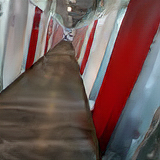}} &
        \fbox{\includegraphics[width=\sz\linewidth, height=\hw]{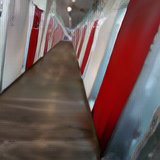}} &
        \fbox{\includegraphics[width=\sz\linewidth, height=\hw]{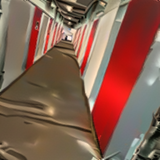}} &
        \fbox{\includegraphics[width=\sz\linewidth, height=\hw]{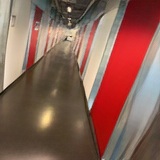}} &
        \fbox{\includegraphics[width=\sz\linewidth, height=\hw]{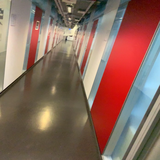}} 
        \\  %
        \fbox{\includegraphics[width=\sz\linewidth, height=\hw]{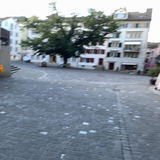}} &
        \fbox{\includegraphics[width=\sz\linewidth, height=\hw]{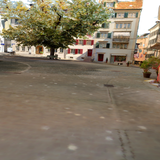}} &
        \fbox{\includegraphics[width=\sz\linewidth, height=\hw]{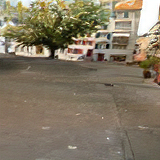}} &
        \fbox{\includegraphics[width=\sz\linewidth, height=\hw]{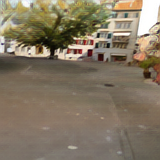}} &
        \fbox{\includegraphics[width=\sz\linewidth, height=\hw]{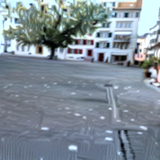}} &
        \fbox{\includegraphics[width=\sz\linewidth, height=\hw]{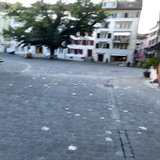}} &
        \fbox{\includegraphics[width=\sz\linewidth, height=\hw]{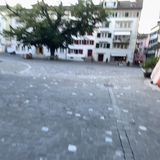}}
        \\  %
        \fbox{\includegraphics[width=\sz\linewidth, height=\hw]{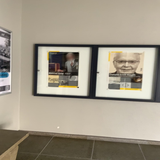}} &
        \fbox{\includegraphics[width=\sz\linewidth, height=\hw]{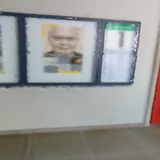}} &
        \fbox{\includegraphics[width=\sz\linewidth, height=\hw]{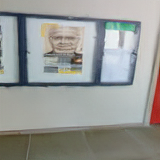}} &
        \fbox{\includegraphics[width=\sz\linewidth, height=\hw]{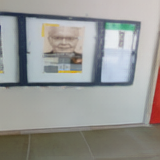}} &
        \fbox{\includegraphics[width=\sz\linewidth, height=\hw]{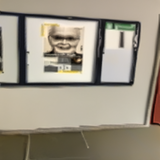}} &
        \fbox{\includegraphics[width=\sz\linewidth, height=\hw]{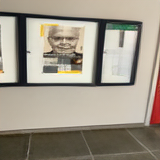}} &
        \fbox{\includegraphics[width=\sz\linewidth, height=\hw]{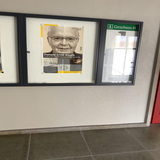}}
        \\ %
        \fbox{\includegraphics[width=\sz\linewidth, height=\hw]{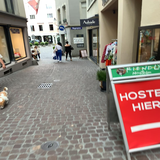}} &
        \fbox{\includegraphics[width=\sz\linewidth, height=\hw]{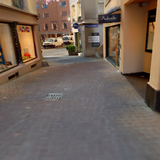}} &
        \fbox{\includegraphics[width=\sz\linewidth, height=\hw]{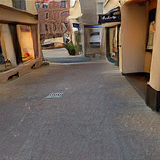}} &
        \fbox{\includegraphics[width=\sz\linewidth, height=\hw]{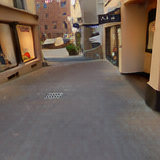}} &
        \fbox{\includegraphics[width=\sz\linewidth, height=\hw]{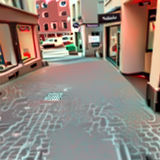}} &
        \fbox{\includegraphics[width=\sz\linewidth, height=\hw]{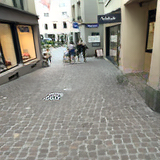}} &
        \fbox{\includegraphics[width=\sz\linewidth, height=\hw]{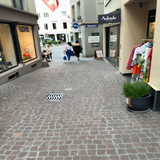}}
        \\ %
        
    \end{tabular}}
    \vspace{-10pt}
  \caption{\textbf{Qualitative comparison.} We compare MASA~\cite{MASASR}, DATSR~\cite{DATSR} (RefSR) and NNST~\cite{NNST} (ST) with \textbf{MaRINeR} on the task of novel view enhancement.}
\verticalspacetable
\verticalspacetable
  \label{fig:qualitative}
\end{figure}
\cref{fig:qualitative} shows a visual comparison with RefSR and ST methods. RefSR methods stay close to their low-res. input structure and color wise. They add details from the reference which are first transformed to the low-res. color distribution. Therefore they can not remove artifacts and the color distribution is not adapted to the one of the reference. Style-transfer methods on the other hand have a built in trade-off between content preservation and style transfer. Photo-realistic methods successfully adapt the color distribution of the reference while not changing the content of the rendering, which inadvertently also preserve the artifacts. Universal style transfer methods transfer both the color and content from the reference to the output so some artifacts can disappear. However, they do not distinguish between real content and artifacts and therefore introduce unrealistic distortions into the image. Semantic style transfer methods match between content and style and successfully transfer the style of matching objects. If no matches are found, then the methods also introduce distortions. Our model successfully transfers the colors, removes rendering artifacts while preserving the underlying content and fills in missing parts.

\vspace{-10pt}
\subsection{Applications}
\vspace{-5pt}
Novel view enhancement can be applied to a variety of situations. We showcase the benefits of using our model for eliminating manual sanity checks, enhancing synthetic trajectories and as post-processing tool for renderings of NeRFs.

\PAR{Validation of localization pseudo-ground-truth.}
A limitation of automatic pseudo-GT pipelines for localization is that they often require manual validation.
For instance, the LaMAR~\cite{LaMAR} pipeline registers sequences of images recorded by AR devices, into a common 3D reconstruction based on a high quality LiDAR scanner. 
To check if the generated alignment is accurate, manual checks between renderings from mesh and input images are needed.
To automate those, an option is to estimate a homography between the rendering of the scene at the estimated pose and the associated image of the input sequence.
The homography should be identity if the localization of the pipeline was successful.
However, estimating a homography between the rendered and real image is not accurate because of the domain gap.
Using our method, we can enhance the renderings and improve the accuracy of the estimated homography.
We use SuperPoint~\cite{Superpoint} for feature extraction and SuperGlue~\cite{Superglue} for matching. \cref{fig:homography_app} and \cref{tab:homography} show that with our method we increase the number of matches and the inlier ratio supporting the homography.
This leads to a more accurate homography estimation, characterized by the homography error~\cite{Superpoint} in the table which is optimally zero. Because of the increased accuracy of the homography we can therefore eliminate the manual sanity checks and replace them with an automated tool.

\begin{table}[tb]
      \caption{\textbf{Applications. Left -- Improved homography estimation.} Using enhanced renderings improves the matching and homography estimation between the rendering and the raw image. \textbf{Right -- NeRFs.} Post-processing NeRF renderings leads to improved visual image quality reflected by better ERQA and LPIPS scores.}
      \verticalspacetable
      \begin{minipage}{0.6\textwidth}
          \label{tab:homography}
          \centering
          \tiny
          \begin{tabular}{lccc}
            \toprule
             & \# matches & inlier ratio & homography error \\ 
            \midrule
            Render $\leftrightarrow$ Image & 39.21 & 61.24\% & 4.86\\
            + \textbf{MaRINeR} $\leftrightarrow$ Image & \fs{58.89} & \fs{78.16\%} &  \fs{1.88}\\
          \bottomrule
          \end{tabular}
        \end{minipage}~
        \begin{minipage}{0.3\textwidth}
            \label{tab:nerf_table}
            \centering
            \tiny
            \begin{tabular}{lcccc}
            \toprule
                   & PSNR & SSIM & ERQA & LPIPS \\
        \midrule
         NeRF output & \fs{21.14} & \fs{0.622}  & 0.646 & 0.238 \\
         + \textbf{MaRINeR}      & 20.45 & 0.592  & \fs{0.701} & \fs{0.167} \\
          \bottomrule
        \end{tabular}
            
        \end{minipage}
    \end{table}

\begin{figure}[!t]
    \vspace{-10pt}
  \centering
    \newcommand{\sz}{0.305}
    \resizebox{0.9\textwidth}{!}{
    \begin{tabular}{cccc}
        \rotatebox[origin=c]{90}{\tiny Render vs GT} &       
        \makecell{\includegraphics[width=\sz\linewidth]{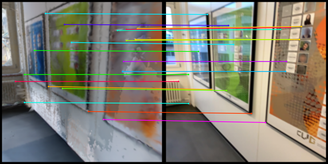}} &
        \makecell{\includegraphics[width=\sz\linewidth]{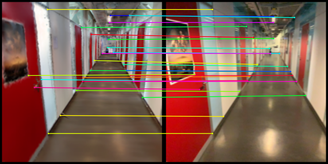}} &
        \makecell{\includegraphics[width=\sz\linewidth]{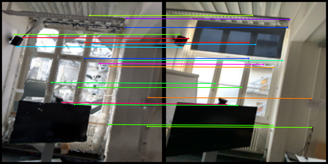}} 
        \\
        \rotatebox[origin=c]{90}{\tiny MaRINeR vs GT} &
        \makecell{\includegraphics[width=\sz\linewidth]{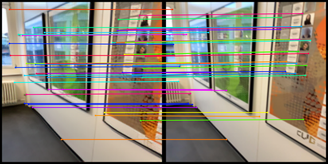}} &
        \makecell{\includegraphics[width=\sz\linewidth]{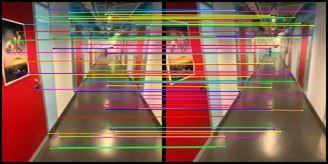}} &
        \makecell{\includegraphics[width=\sz\linewidth]{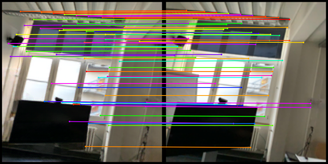}} 
        \\
        
    \end{tabular}}
    \vspace{-10pt}
    \caption{\textbf{Better homography estimation.} Using enhanced renderings of MARINeR, estimating a homography to the aligned source image is more accurate and can be used to automate manual sanity checks in the LaMAR~\cite{LaMAR} pipeline.}
    \label{fig:homography_app}
    \verticalspacetable
    \verticalspacetable
\end{figure}

\PAR{Enhancing synthetic trajectories.}
When creating large AR datasets, substantial human effort is needed to record AR devices trajectories. With recent advances in simulating natural human body movements in 3D scenes such as EgoGen~\cite{Egogen}, synthetic trajectories can be generated effortlessly to extend the existing datasets. However, because of the quality of the underlying 3D representations, there is a gap between synthetic and real images. \cref{fig:domain_gap} shows that with our method we can take a synthetic trajectory and enhance it using a nearby existing reference image from previously recorded trajectories.

\begin{figure}[!b]
    \vspace{-23pt}
    \newcommand{\sz}{0.155}
    \centering
    \hspace*{-10pt}
    \resizebox{0.9\textwidth}{!}{
    \begin{tabular}{ccccccc}
        & \scriptsize t=0 & \scriptsize t=1 & \scriptsize t=2 & \scriptsize t=3 & \scriptsize t=4 & \scriptsize t=5
        \\
        \scriptsize \rotatebox[origin=c]{90}{Synthetic} &
        \makecell{\includegraphics[width=\sz\linewidth]{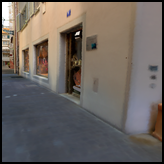}} &
        \makecell{\includegraphics[width=\sz\linewidth]{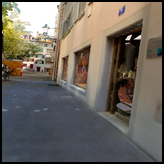}} &
        \makecell{\includegraphics[width=\sz\linewidth]{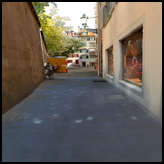}} &
        \makecell{\includegraphics[width=\sz\linewidth]{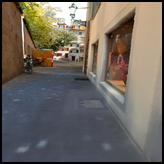}} &
        \makecell{\includegraphics[width=\sz\linewidth]{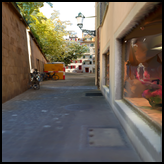}} &
        \makecell{\includegraphics[width=\sz\linewidth]{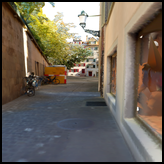}}
        \\
        \scriptsize \rotatebox[origin=c]{90}{Enhanced} &
        \makecell{\includegraphics[width=\sz\linewidth]{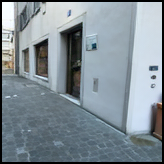}} &
        \makecell{\includegraphics[width=\sz\linewidth]{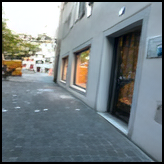}} &
        \makecell{\includegraphics[width=\sz\linewidth]{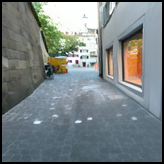}} &
        \makecell{\includegraphics[width=\sz\linewidth]{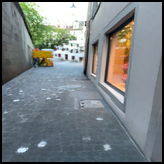}} &
        \makecell{\includegraphics[width=\sz\linewidth]{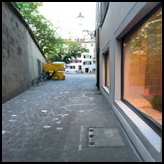}} &
        \makecell{\includegraphics[width=\sz\linewidth]{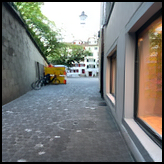}}     
        \\ [-5pt]
        
    \end{tabular}}
    \vspace{-3pt}
    \caption{\textbf{Enhancing synthetic trajectories with nearby localized images.} The result exhibits increased realism and can extend the current dataset without introducing a gap between synthetic and human recorded trajectories.}
    \label{fig:domain_gap}\
      \verticalspace
      \vspace{-7pt}
\end{figure}

\begin{figure}[!t]
\verticalspacetable
    \hspace{-14pt}
    \newcommand{\sz}{0.29}
    \begin{tabular}{cc}
    \begin{minipage}{.5\linewidth}
    \begin{tabular}{cccc}
         & \scriptsize Reference & \scriptsize Rendering & \scriptsize Result
        \\
        \multirow{2}{*}{\rotatebox[origin=c]{90}{\scriptsize Flt. Tree}} & 
        \includegraphics[width=\sz\linewidth]{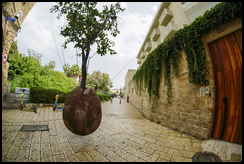} &
        \includegraphics[width=\sz\linewidth]{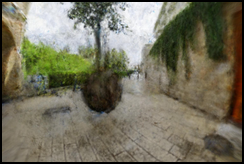} &
        \includegraphics[width=\sz\linewidth]{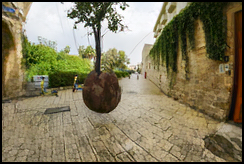}
        \\
        & \includegraphics[width=\sz\linewidth]{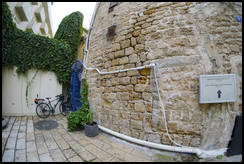} &
        \includegraphics[width=\sz\linewidth]{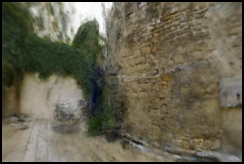} &
        \includegraphics[width=\sz\linewidth]{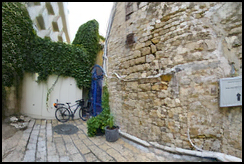}     
        
    \end{tabular}
    \end{minipage} 
    &
    
    \begin{minipage}{.6\linewidth}
    \hspace{-14pt}
    \begin{tabular}{cccc}
        & \scriptsize Reference & \scriptsize Rendering & \scriptsize Result
        \\
        \multirow{2}{*}{\rotatebox[origin=c]{90}{\scriptsize Egypt}} & 
        \includegraphics[width=\sz\linewidth]{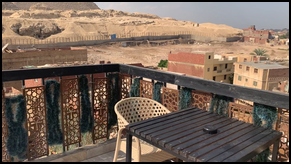} &
        \includegraphics[width=\sz\linewidth]{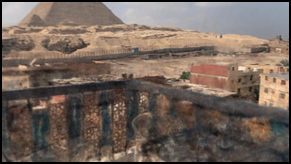} &
        \includegraphics[width=\sz\linewidth]{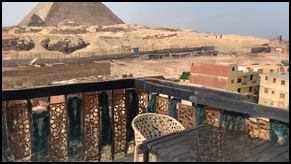}
        \\
        & \includegraphics[width=\sz\linewidth]{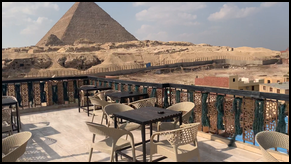} &
        \includegraphics[width=\sz\linewidth]{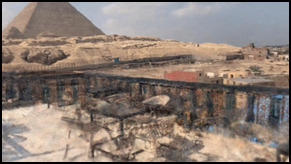} &
        \includegraphics[width=\sz\linewidth]{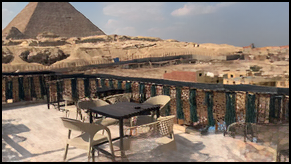}     
    \\
    \end{tabular}
    \end{minipage}
    \\
    \end{tabular}
    \vspace{-7pt}
    \caption{\textbf{NeRF postprocessing results.} Training a nerfacto~\cite{NerfstudioNerfacto} model on the Floating tree and Egypt data. We use the smallest nerfacto model and the result contains artifacts which our model can successfully remove.}
    \label{fig:nerf_floating_tree}
    \verticalspacetable
    \verticalspacetable
\end{figure}

\PAR{NeRF postprocessing.}
Training NeRFs can be computationally expensive and requires a large number of images to generate accurate 3D representations~\cite{NeRFLimitations}. A sufficient number of images may not always be available and training a small model on a large scene with not enough data can lead to noisy representations. \cref{fig:nerf_floating_tree} shows how our off-the-shelf model removes the artifacts created by the nerfacto~\cite{NerfstudioNerfacto} model on the Floating tree and Egypt scenes. Both scenes are large and detail-rich outdoor scenes.
We use the smallest nerfacto model with default parameters and enhance the evaluation images using the closest training image as reference. \cref{tab:nerf_table} shows that our model successfully enhances the nerfacto rendering with respect to ERQA and LPIPS.

\subsection{Ablation study}
We refer to the supplemental for more ablations of the architecture and loss function weights.

\begin{wraptable}{r}{5cm}
\vspace*{-36pt}
      \caption{\textbf{\small 12 Scenes evaluation.} \small %
      }
      \label{tab:12scenes}
      \centering
      \tiny
      \begin{tabular}{c|cccc}
        \toprule
        \multirow{2.5}{*}{Method} & \multicolumn{4}{c}{12 Scenes} \\ \cmidrule{2-5}
                                                                                & PSNR  & SSIM  & ERQA & LPIPS          \\
        \midrule
                                                         Render                 & 20.59 & 0.732 & 0.640 & 0.164                                \\
                           
                                                         \textbf{MaRINeR}                   & \fs{22.99} & \fs{0.775} & \fs{0.703} & \fs{0.071}               \\
      \bottomrule
      \end{tabular}
       \verticalspace
       \vspace{-10pt}
\end{wraptable}
\PAR{Reconstruction method.} %
Contrary to LaMAR~\cite{LaMAR} which uses a NavVis scanner running LiDAR-inertial SLAM followed by the Advancing Front~\cite{AdvancingFront} algorithm for meshing, we report results on the 12 Scenes~\cite{12scenes} dataset which uses RGB-D SLAM on Kinect data and BundleFusion~\cite{BundleFusion} in \cref{tab:12scenes}. %

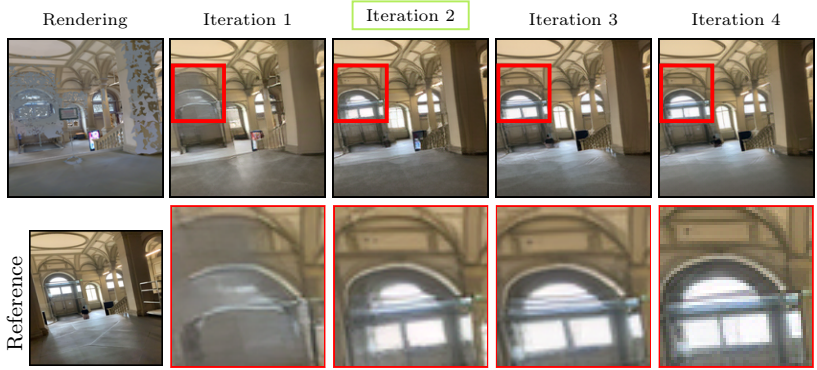
\begin{figure}[!b]
\vspace*{-17pt}
\centering
    \newcommand{\sz}{0.185}
    \renewcommand\fbox{\fcolorbox{red}{red}}
    \setlength{\fboxsep}{0.4pt} 
    \setlength{\fboxrule}{0.4pt}
    \resizebox{.9\textwidth}{!}{
    \begin{tabular}{ccccc}
        \scriptsize Rendering & \scriptsize Iteration 1 & \makecell[c]{\begin{tikzpicture} 
        \draw[draw=inchworm, thick] (0,0) rectangle (1.7,0.4) node[pos=.5] {\scriptsize Iteration 2};
        \end{tikzpicture}} & \scriptsize Iteration 3 & \scriptsize Iteration 4
        \\
        \includegraphics[width=65pt, height=65pt]{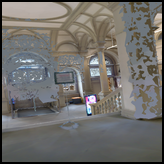} &
        \begin{tikzpicture}
        \node[anchor=south west,inner sep=0] at (0,0) {\includegraphics[width=65pt, height=65pt]{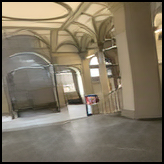}};
        \draw[red,ultra thick] (0.8,1.9) rectangle (0.05,1.1);
        \end{tikzpicture} &
        \begin{tikzpicture}
        \node[anchor=south west,inner sep=0] at (0,0) {\includegraphics[width=65pt, height=65pt]{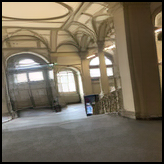}};
        \draw[red,ultra thick] (0.8,1.9) rectangle (0.05,1.1);
        \end{tikzpicture} &
        \begin{tikzpicture}
        \node[anchor=south west,inner sep=0] at (0,0) {\includegraphics[width=65pt, height=65pt]{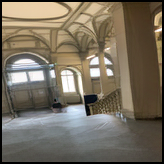}};
        \draw[red,ultra thick] (0.8,1.9) rectangle (0.05,1.1);
        \end{tikzpicture} &
        \begin{tikzpicture}
        \node[anchor=south west,inner sep=0] at (0,0) {\includegraphics[width=65pt, height=65pt]{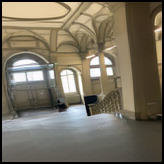}};
        \draw[red,ultra thick] (0.8,1.9) rectangle (0.05,1.1);
        \end{tikzpicture} 
        \\
        \rotatebox[origin=l]{90}{ Reference} \includegraphics[width=0.16\linewidth]{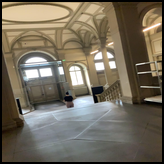} &
        
        \fbox{\includegraphics[width=63pt, height=65pt]{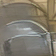}} &
        \fbox{\includegraphics[width=63pt, height=65pt]{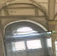}} &
        \fbox{\includegraphics[width=63pt, height=65pt]{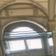}} &
        \fbox{\includegraphics[width=63pt, height=65pt]{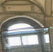}} 
        \\ [-5pt]
        
    \end{tabular}}
    \vspace{-3pt}
  \caption{\textbf{Visual improvement with iterative refinement.} Refining the result over several iterations helps the correspondence matching in presence of large artifacts. }
\label{fig:iterations}
\verticalspacetable
\end{figure}

\PAR{Iterative refinement.}
\cref{fig:iterations} shows the effect of refining the rendering over subsequent iterations. With one iteration the model fails matching regions with large artifacts. When using two or more iterations, the first iteration removes the artifacts, allowing the correspondence matching to succeed in the next ones. We see the largest improvement when using 2 iterations. More iterations gives only a small improvement but linearly increases the inference time.

\PAR{Data augmentation.}
We analyze the effectiveness of our data augmentation strategies. %
\cref{tab:mesh_res}~left shows the model trained without the augmentation performs worse on meshes with different resolutions.
\cref{tab:mesh_res}~right shows that for the case without random reference augmentation, the performance is better when the reference is the GT image.
As soon as the reference level increases, the performance drastically drops compared to the augmented model.

\begin{table}[!t]
  \caption{\textbf{Robustness. Left -- Mesh quality.} Augmenting the data with renderings of a down-sampled mesh increases robustness to changes in the mesh resolution. \textbf{Right -- Reference level.} A higher level indicates a larger temporal distance to the rendering within a sequence, roughly one meter / $20$ degrees / one second per level, which generally correlates with less content in common.}
  \centering
      \verticalspacetable
    \begin{minipage}{0.45\textwidth}
      \label{tab:mesh_res}
      \centering
      \tiny
      \begin{tabular}{ccccc}
        \toprule
        \multirow{2}{*}{Mesh size} & \multicolumn{2}{c}{$\mathrm{CAB}_{\text{aug}}$} & \multicolumn{2}{c}{CAB} \\ \cmidrule(lr){2-3} \cmidrule(lr){4-5} 
        & PSNR & SSIM & PSNR & SSIM \\
        \midrule
          100\% &\fs{19.80} & \fs{0.687} & 19.45 & 0.686\\
          75\% & \fs{19.46} & \fs{0.680} & 18.92 & 0.678\\
          50\% & \fs{19.41} & \fs{0.677} & 18.83 & 0.673\\
          25\% & \fs{19.33} & \fs{0.669} & 18.56 & 0.660\\
          10\% & \fs{19.10} & \fs{0.650} & 17.98 & 0.626\\
      \bottomrule
      \end{tabular}
    \end{minipage}
    \begin{minipage}{0.45\textwidth}
     \label{tab:ref_distance}
      \centering
      \tiny
      \begin{tabular}{ccccc}
        \toprule
        \multirow{2}{*}{Ref. Level} & \multicolumn{2}{c}{$\mathrm{CAB}_{\text{aug}}$} & \multicolumn{2}{c}{CAB} \\ \cmidrule(lr){2-3} \cmidrule(lr){4-5} 
        & PSNR & SSIM & PSNR & SSIM \\
        \midrule
          0 = GT & 22.91 & 0.777& \fs{23.51} & \fs{0.783}\\
          1 & \fs{19.88} & \fs{0.687}& 19.85 & 0.681\\
          2 & \fs{18.99} & \fs{0.664}& 18.48 & 0.644\\
          5 & \fs{18.25} & \fs{0.646}& 17.13 & 0.607\\
          8 & \fs{18.04} & \fs{0.643}& 16.79 & 0.600\\
      \bottomrule
      \end{tabular}
    \end{minipage}
    \verticalspacetable
    \verticalspacetable
\end{table}

\section{Limitations}
\vspace{-7pt}
While the model detects and removes rendering artifacts, it is also possible that some content is wrongly detected as an artifact and removed. This can lead to blurry or smeared out image parts. The model preserves the content of the rendering, but may transfer additional content from the reference. Currently the model works best on images with resolution in the order of 160 with any aspect ratio. Larger resolutions are only indirectly supported, by first down-scaling the rendering, running our model and then up-scaling the image again using a super resolution method, such as Real-ERSGAN~\cite{realERSGAN}. This could be addressed in the future by transitioning to a more advanced matching pipeline such as LoFTR~\cite{LoFTR} or CroCro~\cite{croco} which would come at the cost of more inference time. The model is targeted to enhance low quality renderings, thus high quality renderings are only improved with very close references. The method matches objects on a texture level and not on a semantic level. This means that the objects should have similar texture, where the rendering is a low quality version. The current model may introduce flickering between neighboring frames of a sequence. For video prediction, the pipeline may be further extended to ensure temporal consistency between the generated frames.

\section{Conclusions}
\vspace{-7pt}
In this work, we propose a novel method to enhance renderings of 3D reconstructions. Specifically, we use localized images in the 3D scene to enhance renderings from the 3D reconstruction. Our experiments verify that our model enhances the rendering better than existing models in the domains of RefSR or style transfer. It is scene and device agnostic, robust to mesh resolution changes, generalizes to greyscale and reliably removes 3D reconstruction artifacts. Possible applications include automatization of manual sanity checks in ground-truthing pipelines, enhancement of synthetic training data and improvement of neural renderings trained with limited data or resources.

\section*{Acknowledgements}
This project is partially funded by the Swiss National Science Foundation (SNSF) Advanced Grant number \textbf{TMAG-2\_216260}.

\section*{Appendix}

\appendix 
\renewcommand{\thefigure}{Sup.~\arabic{figure}}
\renewcommand{\thetable}{Sup.~\arabic{table}}
\setcounter{figure}{0}
\setcounter{table}{0}

The supplementary material contains the following sections:
\vspace{-2pt}
\begin{itemize}
    \item[$\diamond$] \cref{sec:training} provides details about the generation of the training dataset.
    \item[$\diamond$] \cref{sec:visuals} presents additional qualitative results.
    \item[$\diamond$] \cref{sec:Ablation} brings additional ablation studies, notably regarding:
    \begin{itemize}
    \item Weights of perceptual and adversarial losses
    \item Architecture
    \item Data augmentation
    \item Iterations
    \end{itemize}
    \item[$\diamond$] Finally, \cref{sec:metrics} describes the metrics used for evaluation:
    \begin{itemize}
    \item Peak Signal-to-Noise Ratio (PSNR)
    \item Structural Similarity Index Measure (SSIM)
    \item Edge Restoration Quality Assessment (ERQA)
    \item Learned Perceptual Image Patch Similarity (LPIPS)
    \end{itemize}
\end{itemize}
\section{Training dataset}
\label{sec:training}
To render the images from the meshes we use RayBender~\cite{RayBender}. Because the dataset created by LaMAR~\cite{LaMAR} contains large rendering artifacts we filter the data first to remove renderings that are only artifact or not recognizable. We do this by calculating a homography error in a similar way as SuperPoint~\cite{Superpoint}.
While this is not an accurate way of assessing whether the localization was successful, it is enough to filter out the renderings with many artifacts. We estimate the homography based on SuperPoint~\cite{Superpoint} features with SuperGlue~\cite{Superglue} matches. Ideally the homography should be identity. The homography error uses the estimated homography to remap the corners of the image. If the corners end up at their original position, the homography is close to identity and the error is close to zero. Using this method we filter out 32 \% of the data. 

\section{Additional visual results}
\label{sec:visuals}
\begin{figure}[!t]
\verticalspacetable
  \centering
    \newcommand{\sz}{0.14}
    \newcommand{\hw}{50.5pt}
    \renewcommand\fbox{\fcolorbox{black}{black}}
    \setlength{\fboxsep}{0.4pt} 
    \setlength{\fboxrule}{0.4pt}
    \resizebox{\textwidth}{!}{
    \begin{tabular}{ccccccc}
        \scriptsize Reference & \scriptsize Rendering & \scriptsize MASA & \scriptsize DATSR & \scriptsize NNST & \scriptsize \textbf{MaRINeR} & \scriptsize GT
        \\  %
        \fbox{\includegraphics[width=\sz\linewidth, height=\hw]{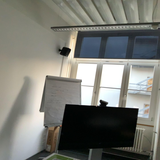}} &
        \fbox{\includegraphics[width=\sz\linewidth, height=\hw]{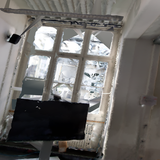}} &
        \fbox{\includegraphics[width=\sz\linewidth, height=\hw]{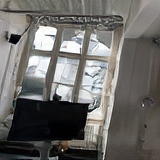}} &
        \fbox{\includegraphics[width=\sz\linewidth, height=\hw]{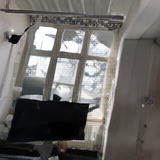}} &
        \fbox{\includegraphics[width=\sz\linewidth, height=\hw]{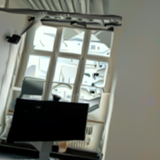}} &
        \fbox{\includegraphics[width=\sz\linewidth, height=\hw]{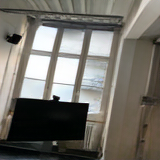}} &
        \fbox{\includegraphics[width=\sz\linewidth, height=\hw]{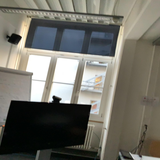}} 
        \\  %
        \fbox{\includegraphics[width=\sz\linewidth, height=\hw]{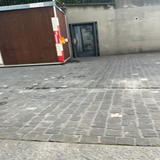}} &
        \fbox{\includegraphics[width=\sz\linewidth, height=\hw]{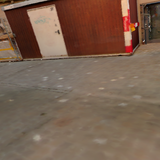}} &
        \fbox{\includegraphics[width=\sz\linewidth, height=\hw]{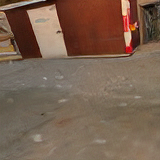}} &
        \fbox{\includegraphics[width=\sz\linewidth, height=\hw]{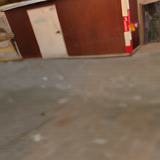}} &
        \fbox{\includegraphics[width=\sz\linewidth, height=\hw]{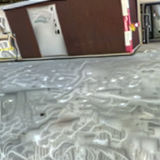}} &
        \fbox{\includegraphics[width=\sz\linewidth, height=\hw]{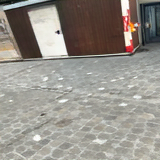}} &
        \fbox{\includegraphics[width=\sz\linewidth, height=\hw]{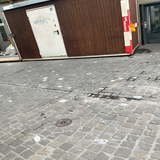}} 
        \\  %
        \fbox{\includegraphics[width=\sz\linewidth, height=\hw]{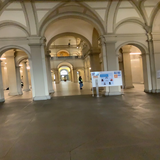}} &
        \fbox{\includegraphics[width=\sz\linewidth, height=\hw]{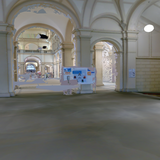}} &
        \fbox{\includegraphics[width=\sz\linewidth, height=\hw]{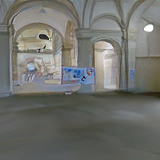}} &
        \fbox{\includegraphics[width=\sz\linewidth, height=\hw]{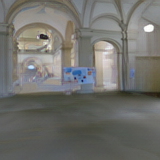}} &
        \fbox{\includegraphics[width=\sz\linewidth, height=\hw]{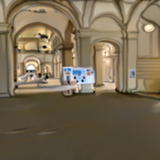}} &
        \fbox{\includegraphics[width=\sz\linewidth, height=\hw]{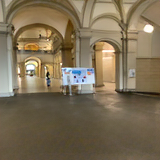}} &
        \fbox{\includegraphics[width=\sz\linewidth, height=\hw]{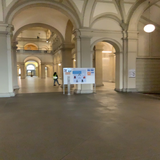}}  
        \\  %
        \fbox{\includegraphics[width=\sz\linewidth, height=\hw]{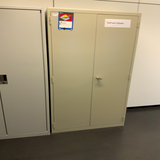}} &
        \fbox{\includegraphics[width=\sz\linewidth, height=\hw]{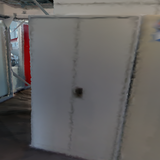}} &
        \fbox{\includegraphics[width=\sz\linewidth, height=\hw]{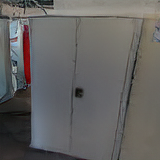}} &
        \fbox{\includegraphics[width=\sz\linewidth, height=\hw]{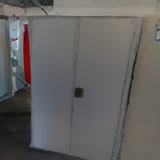}} &
        \fbox{\includegraphics[width=\sz\linewidth, height=\hw]{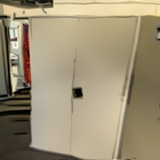}} &
        \fbox{\includegraphics[width=\sz\linewidth, height=\hw]{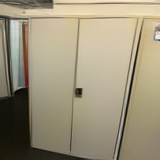}} &
        \fbox{\includegraphics[width=\sz\linewidth, height=\hw]{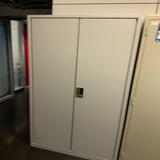}} 
        \\  %
        \fbox{\includegraphics[width=\sz\linewidth, height=\hw]{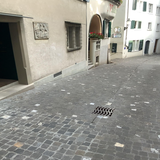}} &
        \fbox{\includegraphics[width=\sz\linewidth, height=\hw]{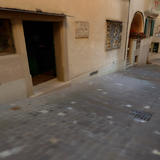}} &
        \fbox{\includegraphics[width=\sz\linewidth, height=\hw]{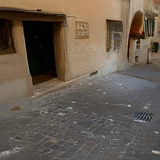}} &
        \fbox{\includegraphics[width=\sz\linewidth, height=\hw]{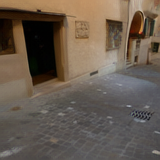}} &
        \fbox{\includegraphics[width=\sz\linewidth, height=\hw]{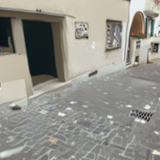}} &
        \fbox{\includegraphics[width=\sz\linewidth, height=\hw]{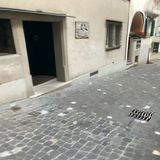}} &
        \fbox{\includegraphics[width=\sz\linewidth, height=\hw]{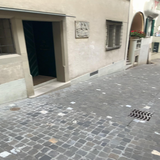}} 
        \\ %
        \fbox{\includegraphics[width=\sz\linewidth, height=\hw]{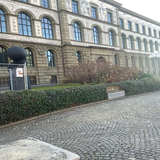}} &
        \fbox{\includegraphics[width=\sz\linewidth, height=\hw]{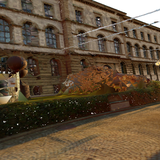}} &
        \fbox{\includegraphics[width=\sz\linewidth, height=\hw]{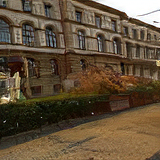}} &
        \fbox{\includegraphics[width=\sz\linewidth, height=\hw]{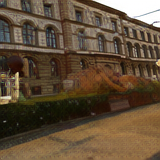}} &
        \fbox{\includegraphics[width=\sz\linewidth, height=\hw]{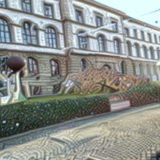}} &
        \fbox{\includegraphics[width=\sz\linewidth, height=\hw]{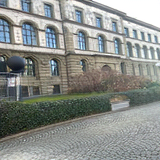}} &
        \fbox{\includegraphics[width=\sz\linewidth, height=\hw]{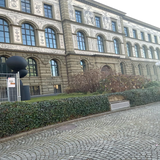}} 
        \\
        \fbox{\includegraphics[width=\sz\linewidth, height=\hw]{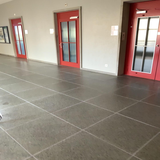}} &
        \fbox{\includegraphics[width=\sz\linewidth, height=\hw]{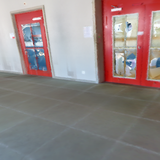}} &
        \fbox{\includegraphics[width=\sz\linewidth, height=\hw]{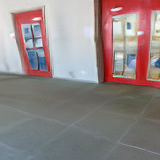}} &
        \fbox{\includegraphics[width=\sz\linewidth, height=\hw]{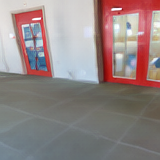}} &
        \fbox{\includegraphics[width=\sz\linewidth, height=\hw]{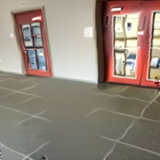}} &
        \fbox{\includegraphics[width=\sz\linewidth, height=\hw]{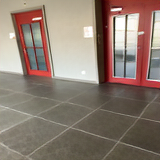}} &
        \fbox{\includegraphics[width=\sz\linewidth, height=\hw]{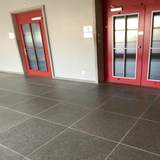}} 
        \\
        \fbox{\includegraphics[width=\sz\linewidth, height=\hw]{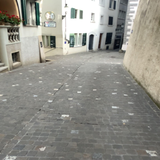}} &
        \fbox{\includegraphics[width=\sz\linewidth, height=\hw]{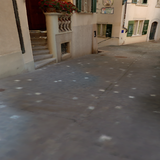}} &
        \fbox{\includegraphics[width=\sz\linewidth, height=\hw]{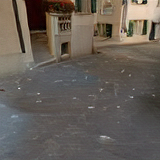}} &
        \fbox{\includegraphics[width=\sz\linewidth, height=\hw]{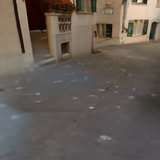}} &
        \fbox{\includegraphics[width=\sz\linewidth, height=\hw]{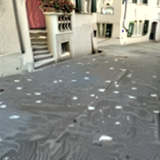}} &
        \fbox{\includegraphics[width=\sz\linewidth, height=\hw]{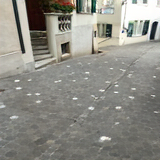}} &
        \fbox{\includegraphics[width=\sz\linewidth, height=\hw]{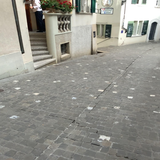}} 
        \\
        \fbox{\includegraphics[width=\sz\linewidth, height=\hw]{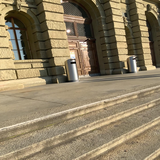}} &
        \fbox{\includegraphics[width=\sz\linewidth, height=\hw]{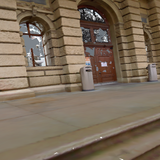}} &
        \fbox{\includegraphics[width=\sz\linewidth, height=\hw]{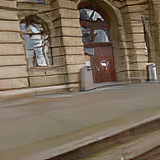}} &
        \fbox{\includegraphics[width=\sz\linewidth, height=\hw]{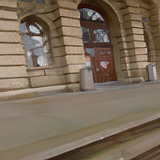}} &
        \fbox{\includegraphics[width=\sz\linewidth, height=\hw]{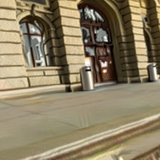}} &
        \fbox{\includegraphics[width=\sz\linewidth, height=\hw]{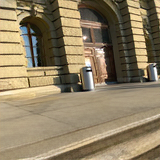}} &
        \fbox{\includegraphics[width=\sz\linewidth, height=\hw]{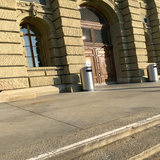}} 
        \\
        \fbox{\includegraphics[width=\sz\linewidth, height=\hw]{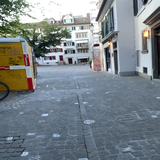}} &
        \fbox{\includegraphics[width=\sz\linewidth, height=\hw]{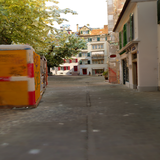}} &
        \fbox{\includegraphics[width=\sz\linewidth, height=\hw]{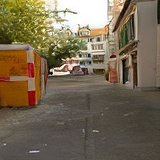}} &
        \fbox{\includegraphics[width=\sz\linewidth, height=\hw]{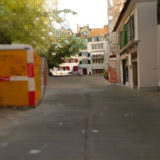}} &
        \fbox{\includegraphics[width=\sz\linewidth, height=\hw]{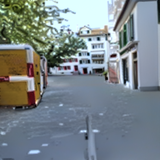}} &
        \fbox{\includegraphics[width=\sz\linewidth, height=\hw]{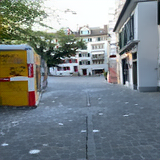}} &
        \fbox{\includegraphics[width=\sz\linewidth, height=\hw]{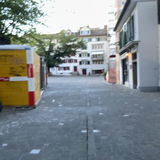}} 
        \\ %
        
    \end{tabular}}
    \caption{\textbf{Qualitative comparison -- 2.} Additional results of comparing MASA~\cite{MASASR}, DATSR~\cite{DATSR} (RefSR) and NNST~\cite{NNST} (ST) with \textbf{MaRINeR} on the task of novel view enhancement.}
\verticalspacetable
\verticalspacetable
  \label{fig:qualitative2}
\end{figure}

\begin{figure}[!t]
\verticalspacetable
    \centering
    \hspace{-14pt}
        \newcommand{\sz}{0.31}

    \hspace*{-10pt}
    \renewcommand\fbox{\fcolorbox{black}{black}}
    \setlength{\fboxsep}{0.4pt} 
    \setlength{\fboxrule}{0.4pt}
    \begin{tabular}{cc}
    \begin{tabular}{cccc}
         & \scriptsize Reference & \scriptsize Rendering & \scriptsize Result
        \\
        \multirow{3}{*}{\vspace{-60pt}\rotatebox[origin=c]{90}{\scriptsize Flt. Tree}} & 
        \fbox{\includegraphics[width=\sz\linewidth]{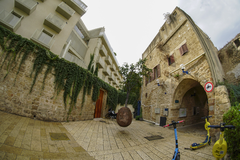}} &
        \fbox{\includegraphics[width=\sz\linewidth]{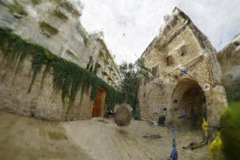}} &
        \fbox{\includegraphics[width=\sz\linewidth]{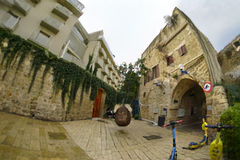}}
        \\
        & \fbox{\includegraphics[width=\sz\linewidth]{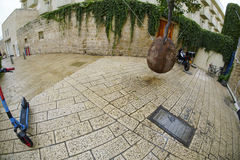}} &
        \fbox{\includegraphics[width=\sz\linewidth]{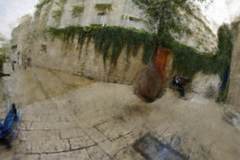}} &
        \fbox{\includegraphics[width=\sz\linewidth]{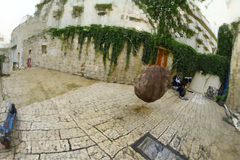}}
        \\
        & \fbox{\includegraphics[width=\sz\linewidth]{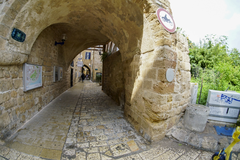}} &
        \fbox{\includegraphics[width=\sz\linewidth]{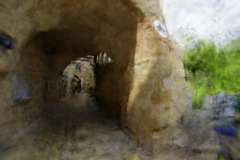}} &
        \fbox{\includegraphics[width=\sz\linewidth]{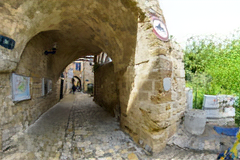}}  
        \\
        
        \multirow{3}{*}{\vspace{-50pt}\rotatebox[origin=c]{90}{\scriptsize Egypt}} & 
        \fbox{\includegraphics[width=\sz\linewidth]{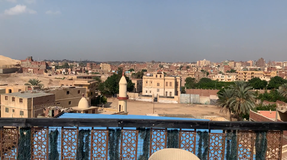}} &
        \fbox{\includegraphics[width=\sz\linewidth]{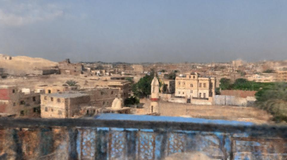}} &
        \fbox{\includegraphics[width=\sz\linewidth]{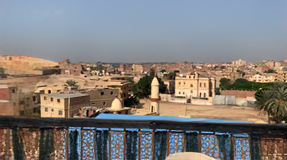}}
        \\
        & \fbox{\includegraphics[width=\sz\linewidth]{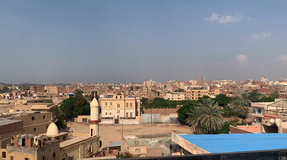}} &
        \fbox{\includegraphics[width=\sz\linewidth]{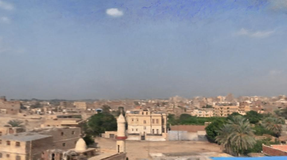}} &
        \fbox{\includegraphics[width=\sz\linewidth]{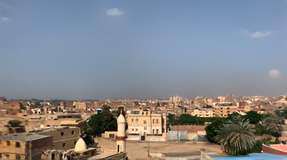}}
        \\
        & \fbox{\includegraphics[width=\sz\linewidth]{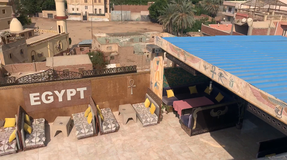}} &
        \fbox{\includegraphics[width=\sz\linewidth]{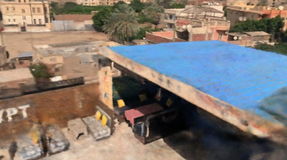}} &
        \fbox{\includegraphics[width=\sz\linewidth]{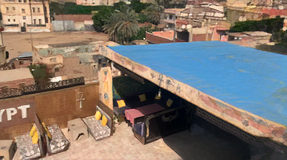}}  
    \\
    \end{tabular}
    \\
    \end{tabular}
    \caption{\textbf{Additional NeRF postprocessing results.} Training a nerfacto~\cite{NerfstudioNerfacto} model on the Floating tree and Egypt data. We use the smallest nerfacto model and the result contains artifacts which our model can successfully remove.}
    \label{fig:nerf}
\end{figure}

\begin{figure}[!htb]
  \centering
     \renewcommand\fbox{\fcolorbox{black}{black}}
    \setlength{\fboxsep}{0.4pt} 
    \setlength{\fboxrule}{0.4pt}
    \newcommand{\sz}{0.239}
    \newcommand{\hw}{56.5pt}
    \setlength{\tabcolsep}{1pt}

    \begin{tabular}{cccc}
        Reference & Rendering & Result & GT
        \\
              
        \fbox{\includegraphics[width=\sz\linewidth]{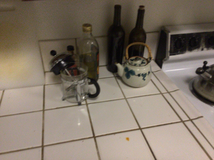}} &
        \fbox{\includegraphics[width=\sz\linewidth]{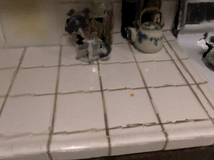}} &
        \fbox{\includegraphics[width=\sz\linewidth]{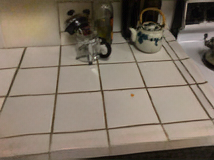}} &
        \fbox{\includegraphics[width=\sz\linewidth]{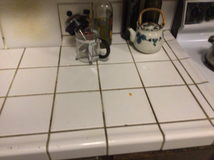}} 

        \\

        \fbox{\includegraphics[width=\sz\linewidth]{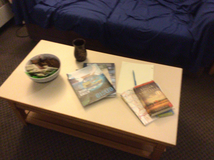}} &
        \fbox{\includegraphics[width=\sz\linewidth]{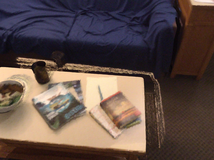}} &
        \fbox{\includegraphics[width=\sz\linewidth]{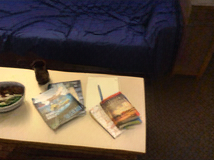}} &
        \fbox{\includegraphics[width=\sz\linewidth]{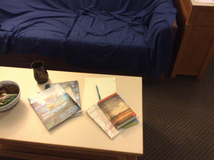}} 

        \\

        \fbox{\includegraphics[width=\sz\linewidth]{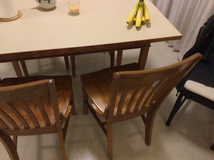}} &
        \fbox{\includegraphics[width=\sz\linewidth]{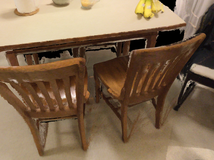}} &
        \fbox{\includegraphics[width=\sz\linewidth]{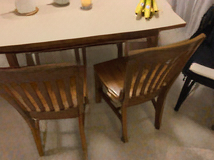}} &
        \fbox{\includegraphics[width=\sz\linewidth]{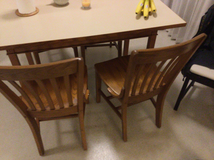}} 
        \\

        \fbox{\includegraphics[width=\sz\linewidth]{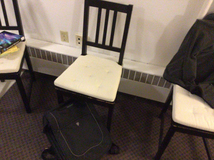}} &
        \fbox{\includegraphics[width=\sz\linewidth]{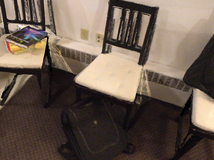}} &
        \fbox{\includegraphics[width=\sz\linewidth]{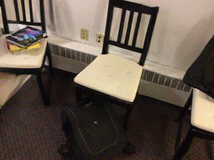}} &
        \fbox{\includegraphics[width=\sz\linewidth]{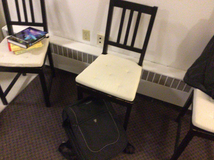}} 

        \\
        \fbox{\includegraphics[width=\sz\linewidth]{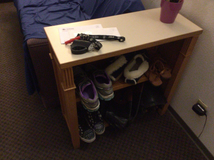}} &
        \fbox{\includegraphics[width=\sz\linewidth]{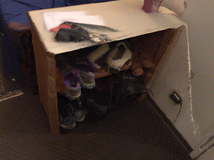}} &
        \fbox{\includegraphics[width=\sz\linewidth]{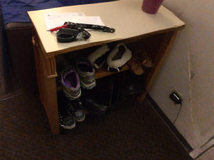}} &
        \fbox{\includegraphics[width=\sz\linewidth]{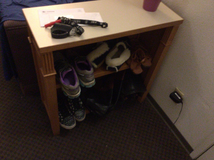}} 
        
        \\
        \fbox{\includegraphics[width=\sz\linewidth]{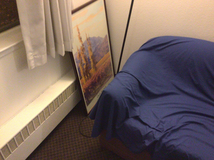}} &
        \fbox{\includegraphics[width=\sz\linewidth]{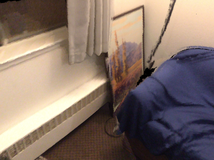}} &
        \fbox{\includegraphics[width=\sz\linewidth]{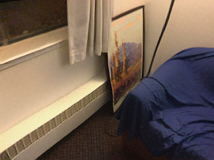}} &
        \fbox{\includegraphics[width=\sz\linewidth]{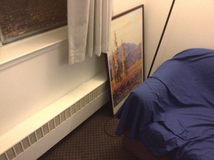}} 

                \\
        \fbox{\includegraphics[width=\sz\linewidth]{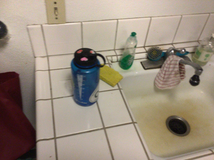}} &
        \fbox{\includegraphics[width=\sz\linewidth]{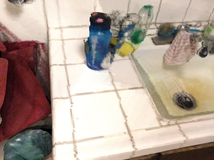}} &
        \fbox{\includegraphics[width=\sz\linewidth]{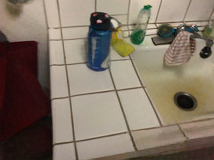}} &
        \fbox{\includegraphics[width=\sz\linewidth]{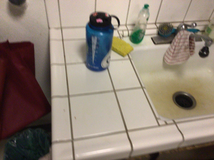}} 
        
    \end{tabular}
  \caption{\textbf{12 Scenes results.} Evaluating the model on the apartment 1 kitchen and living scenes of the 12 Scenes~\cite{12scenes} dataset shows that \textbf{MaRINeR} also enhances renderings of 3D reconstructions created by a different algorithm than the one used by LaMAR~\cite{LaMAR}.}
  \label{fig:12scenes}
\end{figure}

\begin{figure}[!htb]
  \centering
     \renewcommand\fbox{\fcolorbox{black}{black}}
    \setlength{\fboxsep}{0.4pt} 
    \setlength{\fboxrule}{0.4pt}
    \newcommand{\sz}{0.239}
    \newcommand{\hw}{56.5pt}
    \setlength{\tabcolsep}{1pt}

    \begin{tabular}{cccc}
        Reference & Rendering & Result & GT
        \\
              
        \fbox{\includegraphics[width=\sz\linewidth]{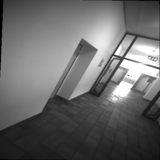}} &
        \fbox{\includegraphics[width=\sz\linewidth]{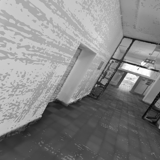}} &
        \fbox{\includegraphics[width=\sz\linewidth]{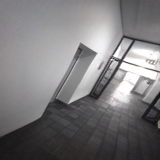}} &
        \fbox{\includegraphics[width=\sz\linewidth]{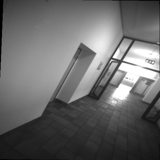}} 

        \\

        \fbox{\includegraphics[width=\sz\linewidth]{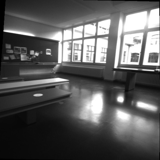}} &
        \fbox{\includegraphics[width=\sz\linewidth]{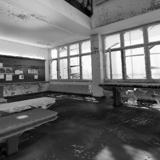}} &
        \fbox{\includegraphics[width=\sz\linewidth]{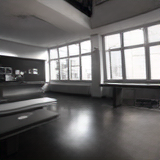}} &
        \fbox{\includegraphics[width=\sz\linewidth]{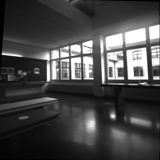}} 

        \\

        \fbox{\includegraphics[width=\sz\linewidth]{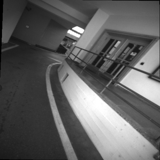}} &
        \fbox{\includegraphics[width=\sz\linewidth]{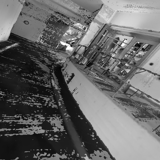}} &
        \fbox{\includegraphics[width=\sz\linewidth]{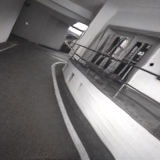}} &
        \fbox{\includegraphics[width=\sz\linewidth]{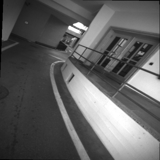}} 

        \\

        \fbox{\includegraphics[width=\sz\linewidth]{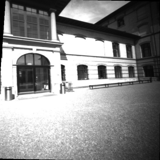}} &
        \fbox{\includegraphics[width=\sz\linewidth]{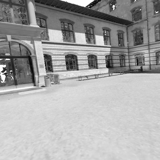}} &
        \fbox{\includegraphics[width=\sz\linewidth]{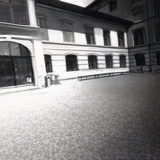}} &
        \fbox{\includegraphics[width=\sz\linewidth]{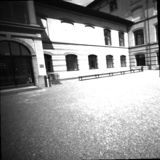}}

        \\
        \fbox{\includegraphics[width=\sz\linewidth]{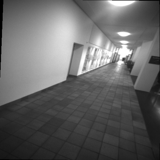}} &
        \fbox{\includegraphics[width=\sz\linewidth]{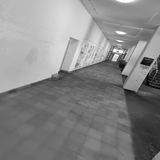}} &
        \fbox{\includegraphics[width=\sz\linewidth]{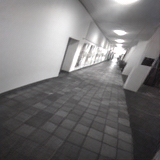}} &
        \fbox{\includegraphics[width=\sz\linewidth]{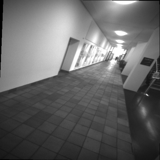}}

        \\
        \fbox{\includegraphics[width=\sz\linewidth]{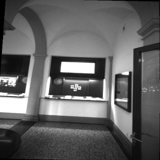}} &
        \fbox{\includegraphics[width=\sz\linewidth]{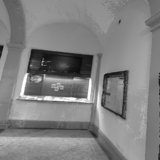}} &
        \fbox{\includegraphics[width=\sz\linewidth]{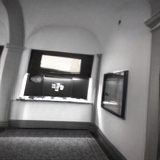}} &
        \fbox{\includegraphics[width=\sz\linewidth]{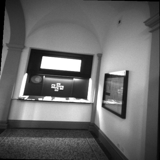}}

    \end{tabular}
  \caption{\textbf{HoloLens 2 results.} Enhancing greyscale renderings using references recorded by a HoloLens 2 device.}
  \label{fig:greyscale}
\end{figure}

\begin{figure}[!t]
  \centering
    \setlength{\fboxsep}{0.4pt} 
    \setlength{\fboxrule}{0.4pt}
    \newcommand{\sz}{0.239}
    \newcommand{\hw}{56.5pt}
    \begin{tabular}{cccc}
         Reference & Rendering  &  Result & GT
        \\
         \fbox{\includegraphics[width=\sz\linewidth]{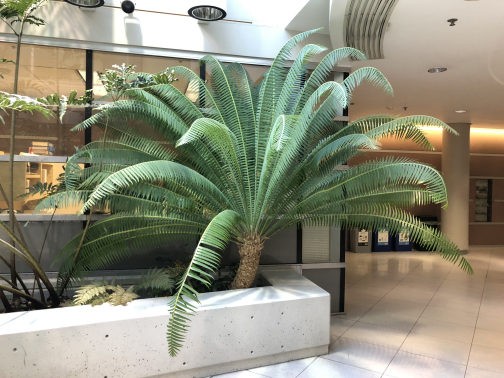}} &
         \fbox{\includegraphics[width=\sz\linewidth]{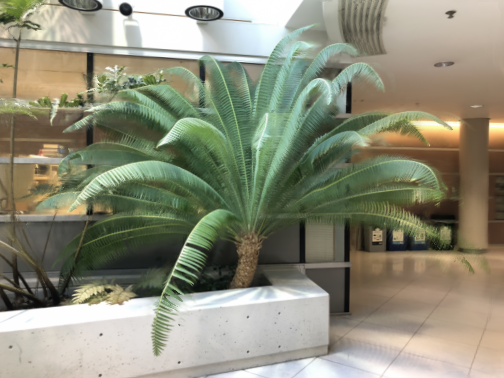}} &
         \fbox{\includegraphics[width=\sz\linewidth]{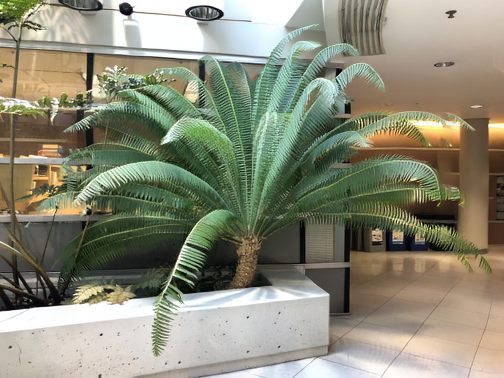}} &
         \fbox{\includegraphics[width=\sz\linewidth]{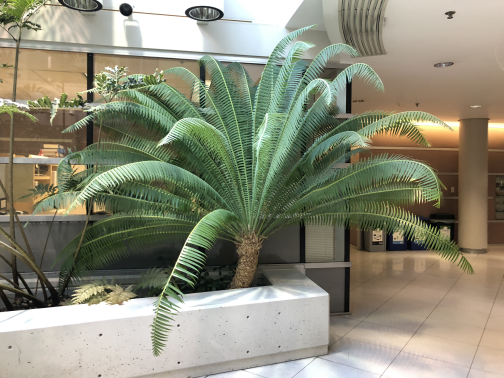}} 
         \\
         \fbox{\includegraphics[width=\sz\linewidth]{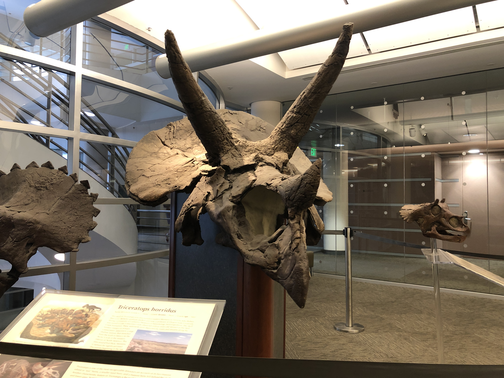}} &
         \fbox{\includegraphics[width=\sz\linewidth]{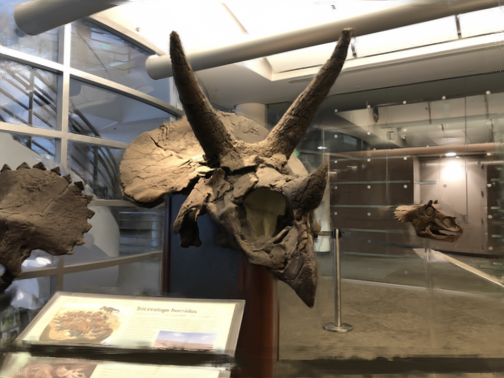}} &
         \fbox{\includegraphics[width=\sz\linewidth]{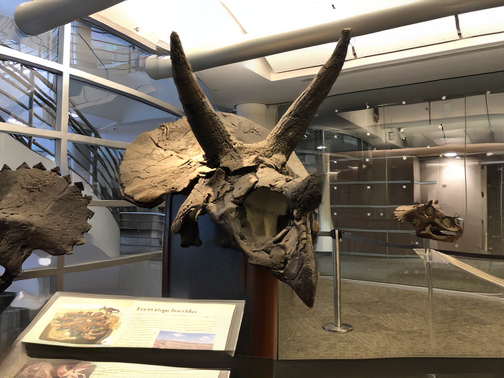}} &
         \fbox{\includegraphics[width=\sz\linewidth]{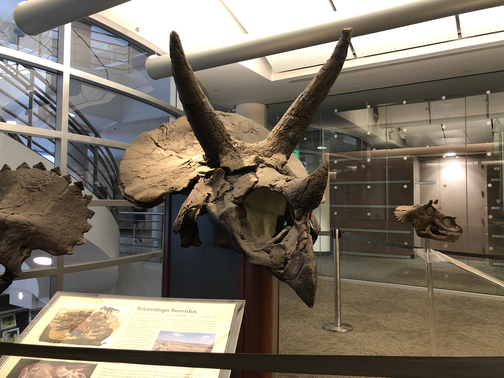}}
        \\
        \fbox{\includegraphics[width=\sz\linewidth]{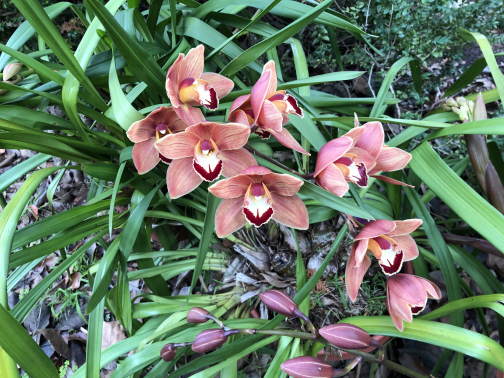}} &
        \fbox{\includegraphics[width=\sz\linewidth]{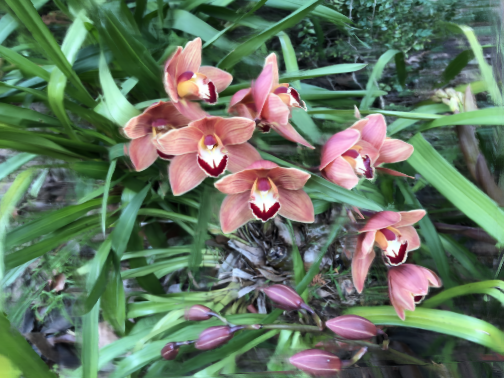}} &
        \fbox{\includegraphics[width=\sz\linewidth]{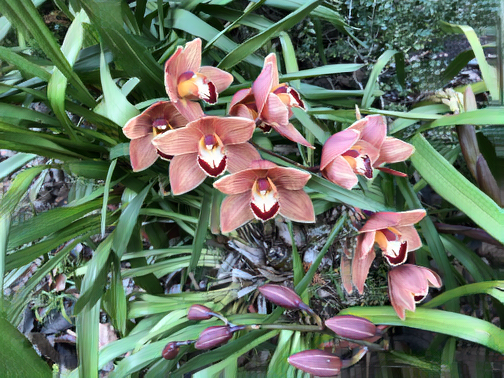}} &
        \fbox{\includegraphics[width=\sz\linewidth]{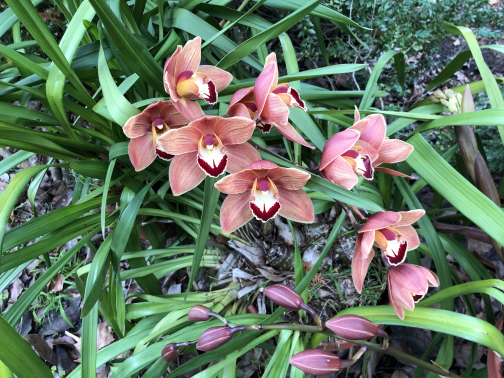}} 
        \\
        \fbox{\includegraphics[width=\sz\linewidth]{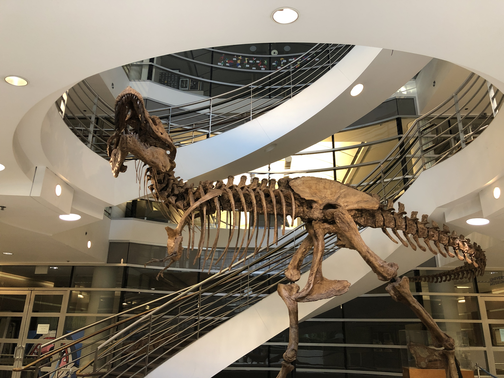}} &
        \fbox{\includegraphics[width=\sz\linewidth]{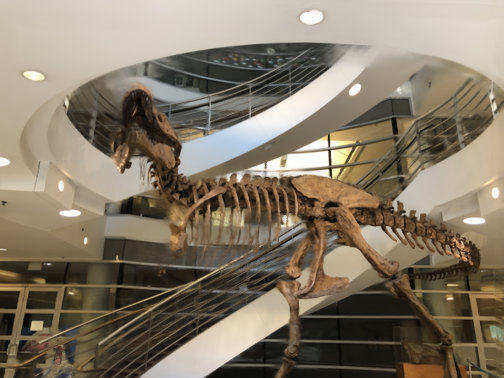}} &
        \fbox{\includegraphics[width=\sz\linewidth]{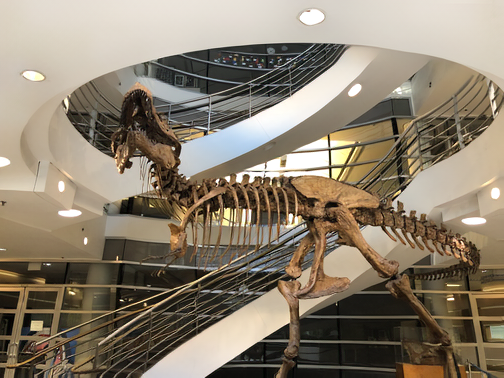}} &
        \fbox{\includegraphics[width=\sz\linewidth]{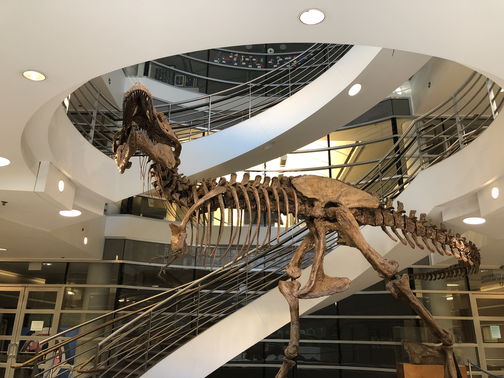}}
        \\
        \fbox{\includegraphics[width=\sz\linewidth]{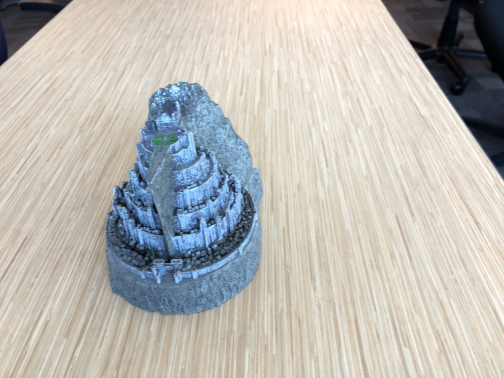}} &
        \fbox{\includegraphics[width=\sz\linewidth]{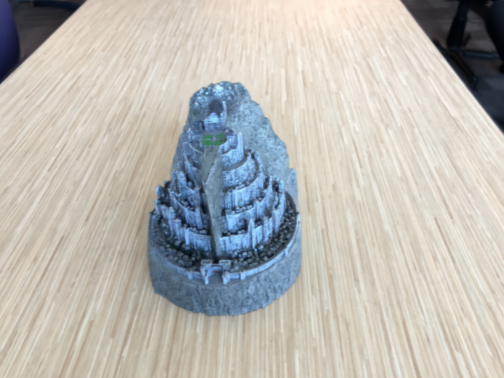}} &
        \fbox{\includegraphics[width=\sz\linewidth]{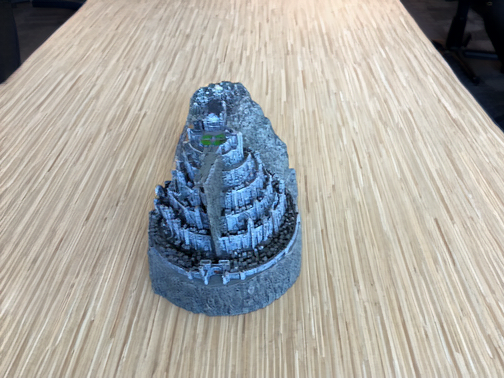}} &
        \fbox{\includegraphics[width=\sz\linewidth]{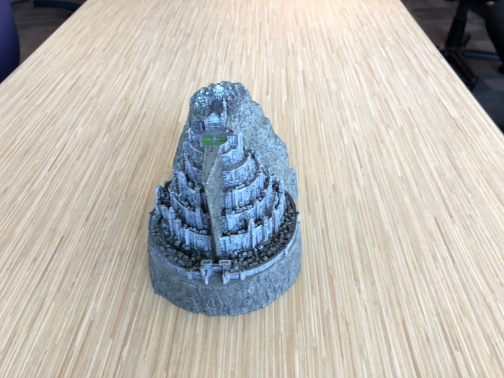}}
        \\
        \fbox{\includegraphics[width=\sz\linewidth]{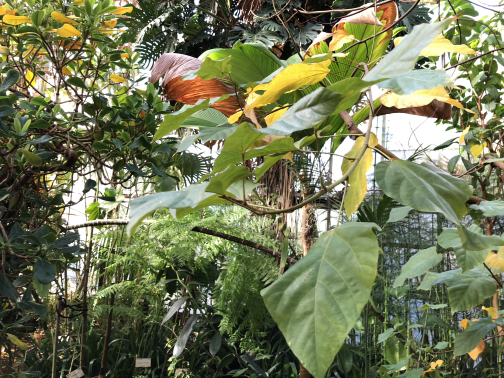}} &
        \fbox{\includegraphics[width=\sz\linewidth]{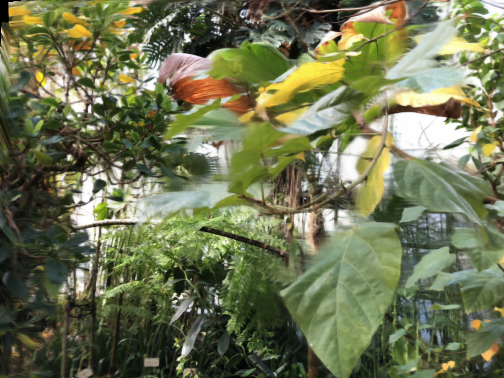}} &
        \fbox{\includegraphics[width=\sz\linewidth]{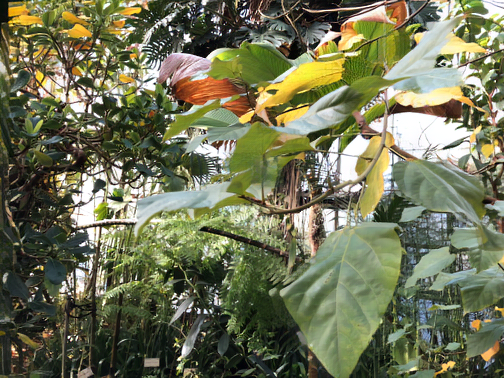}} &
        \fbox{\includegraphics[width=\sz\linewidth]{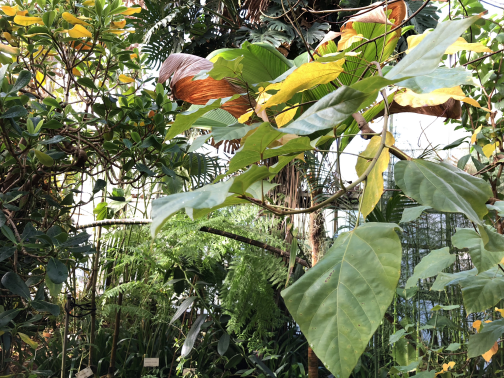}}
        \\
        \fbox{\includegraphics[width=\sz\linewidth]{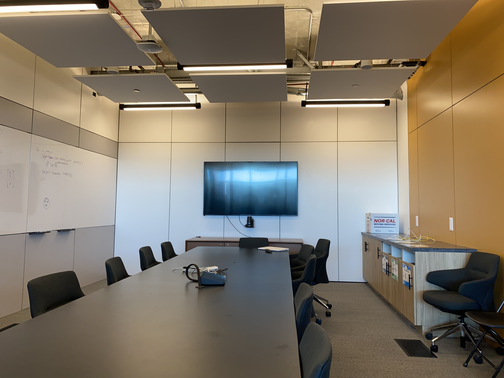}} &
        \fbox{\includegraphics[width=\sz\linewidth]{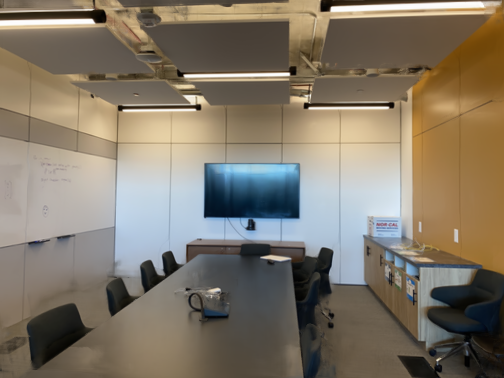}} &
        \fbox{\includegraphics[width=\sz\linewidth]{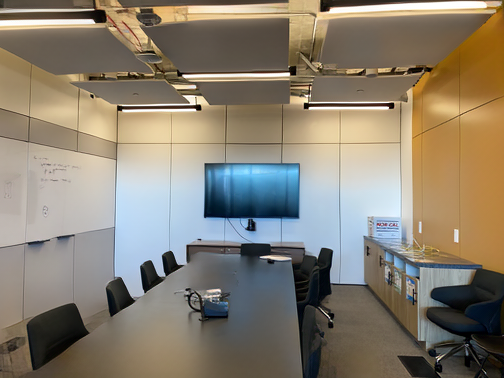}} &
        \fbox{\includegraphics[width=\sz\linewidth]{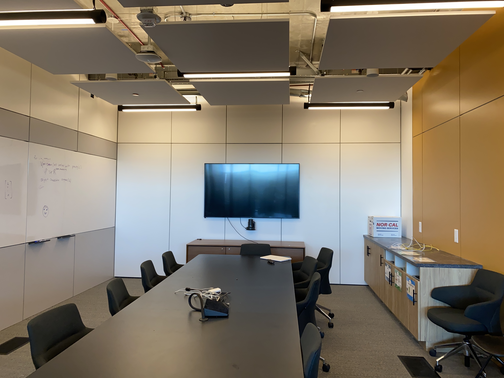}}
        
    \end{tabular}

  \caption{Enhancing novel view renderings created by the image-based rendering method IBRNet~\cite{IBRnet} using our model without retraining.}
  \label{fig:IBRnet}
\end{figure}

We show additional results for the qualitative comparison in \cref{fig:qualitative} and \cref{fig:qualitative2}. \cref{fig:homography_app} shows further results of the validation of localization pseudo-ground-truth. \cref{fig:domain_gap} shows more enhanced synthetic trajectories and \cref{fig:nerf} shows further results of enhanced NeRF renderings. \cref{fig:12scenes} shows additional results on the 12 Scenes~\cite{12scenes} dataset and \cref{fig:greyscale} shows additional results of enhancing greyscale renderings using references captured by a HoloLens 2 device. \cref{fig:IBRnet} shows zero-shot prediction results for renderings of the image-based rendering method IBRnet~\cite{IBRnet}.
\begin{figure}[!t]
\verticalspacetable
  \centering
    \newcommand{\sz}{0.14}
    \newcommand{\hw}{50.5pt}
    \renewcommand\fbox{\fcolorbox{black}{black}}
    \setlength{\fboxsep}{0.4pt} 
    \setlength{\fboxrule}{0.4pt}
    \resizebox{\textwidth}{!}{
    \begin{tabular}{ccccccc}
        \scriptsize Reference & \scriptsize Rendering & \scriptsize MASA & \scriptsize DATSR & \scriptsize NNST & \scriptsize \textbf{MaRINeR} & \scriptsize GT
        \\  %
        \fbox{\includegraphics[width=\sz\linewidth, height=\hw]{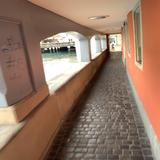}} &
        \fbox{\includegraphics[width=\sz\linewidth, height=\hw]{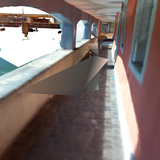}} &
        \fbox{\includegraphics[width=\sz\linewidth, height=\hw]{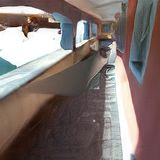}} &
        \fbox{\includegraphics[width=\sz\linewidth, height=\hw]{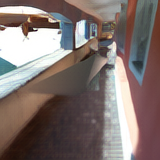}} &
        \fbox{\includegraphics[width=\sz\linewidth, height=\hw]{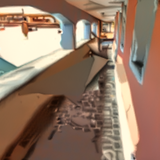}} &
        \fbox{\includegraphics[width=\sz\linewidth, height=\hw]{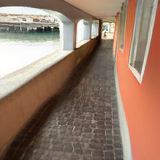}} &
        \fbox{\includegraphics[width=\sz\linewidth, height=\hw]{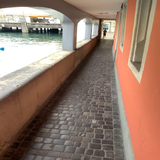}} 
        \\  %
        \fbox{\includegraphics[width=\sz\linewidth, height=\hw]{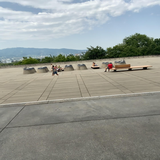}} &
        \fbox{\includegraphics[width=\sz\linewidth, height=\hw]{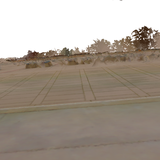}} &
        \fbox{\includegraphics[width=\sz\linewidth, height=\hw]{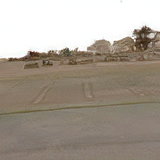}} &
        \fbox{\includegraphics[width=\sz\linewidth, height=\hw]{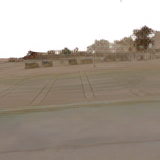}} &
        \fbox{\includegraphics[width=\sz\linewidth, height=\hw]{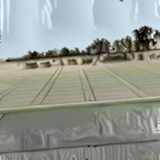}} &
        \fbox{\includegraphics[width=\sz\linewidth, height=\hw]{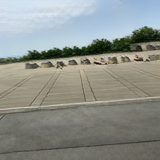}} &
        \fbox{\includegraphics[width=\sz\linewidth, height=\hw]{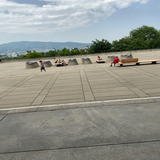}} 
        \\  %
        \fbox{\includegraphics[width=\sz\linewidth, height=\hw]{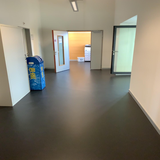}} &
        \fbox{\includegraphics[width=\sz\linewidth, height=\hw]{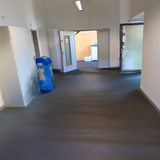}} &
        \fbox{\includegraphics[width=\sz\linewidth, height=\hw]{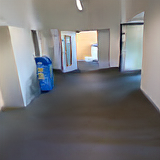}} &
        \fbox{\includegraphics[width=\sz\linewidth, height=\hw]{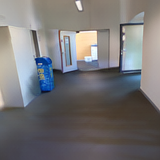}} &
        \fbox{\includegraphics[width=\sz\linewidth, height=\hw]{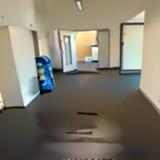}} &
        \fbox{\includegraphics[width=\sz\linewidth, height=\hw]{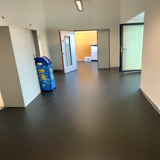}} &
        \fbox{\includegraphics[width=\sz\linewidth, height=\hw]{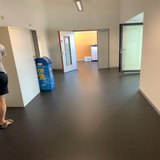}}  
        \\  %
        \fbox{\includegraphics[width=\sz\linewidth, height=\hw]{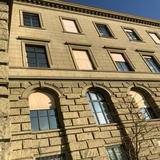}} &
        \fbox{\includegraphics[width=\sz\linewidth, height=\hw]{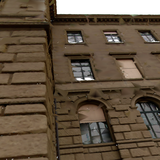}} &
        \fbox{\includegraphics[width=\sz\linewidth, height=\hw]{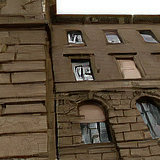}} &
        \fbox{\includegraphics[width=\sz\linewidth, height=\hw]{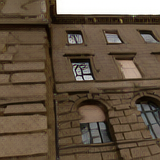}} &
        \fbox{\includegraphics[width=\sz\linewidth, height=\hw]{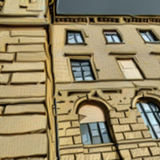}} &
        \fbox{\includegraphics[width=\sz\linewidth, height=\hw]{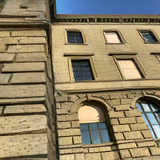}} &
        \fbox{\includegraphics[width=\sz\linewidth, height=\hw]{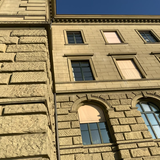}} 
        \\  %
        \fbox{\includegraphics[width=\sz\linewidth, height=\hw]{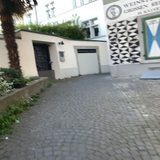}} &
        \fbox{\includegraphics[width=\sz\linewidth, height=\hw]{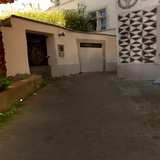}} &
        \fbox{\includegraphics[width=\sz\linewidth, height=\hw]{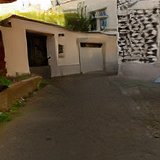}} &
        \fbox{\includegraphics[width=\sz\linewidth, height=\hw]{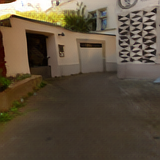}} &
        \fbox{\includegraphics[width=\sz\linewidth, height=\hw]{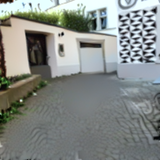}} &
        \fbox{\includegraphics[width=\sz\linewidth, height=\hw]{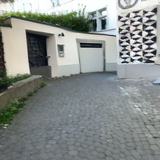}} &
        \fbox{\includegraphics[width=\sz\linewidth, height=\hw]{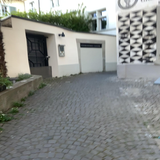}} 
        \\ %
        \fbox{\includegraphics[width=\sz\linewidth, height=\hw]{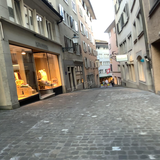}} &
        \fbox{\includegraphics[width=\sz\linewidth, height=\hw]{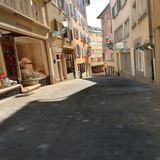}} &
        \fbox{\includegraphics[width=\sz\linewidth, height=\hw]{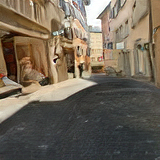}} &
        \fbox{\includegraphics[width=\sz\linewidth, height=\hw]{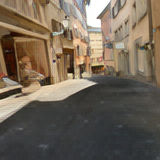}} &
        \fbox{\includegraphics[width=\sz\linewidth, height=\hw]{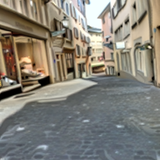}} &
        \fbox{\includegraphics[width=\sz\linewidth, height=\hw]{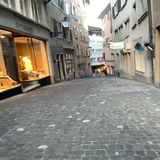}} &
        \fbox{\includegraphics[width=\sz\linewidth, height=\hw]{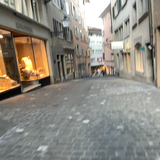}} 
        \\
        \fbox{\includegraphics[width=\sz\linewidth, height=\hw]{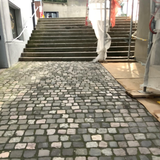}} &
        \fbox{\includegraphics[width=\sz\linewidth, height=\hw]{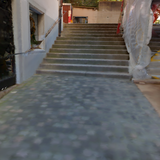}} &
        \fbox{\includegraphics[width=\sz\linewidth, height=\hw]{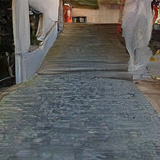}} &
        \fbox{\includegraphics[width=\sz\linewidth, height=\hw]{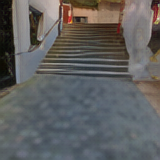}} &
        \fbox{\includegraphics[width=\sz\linewidth, height=\hw]{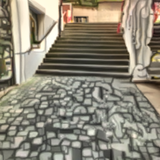}} &
        \fbox{\includegraphics[width=\sz\linewidth, height=\hw]{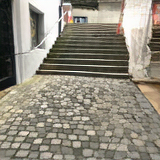}} &
        \fbox{\includegraphics[width=\sz\linewidth, height=\hw]{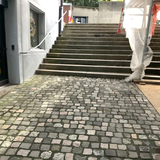}} 
        \\
        \fbox{\includegraphics[width=\sz\linewidth, height=\hw]{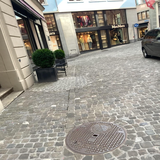}} &
        \fbox{\includegraphics[width=\sz\linewidth, height=\hw]{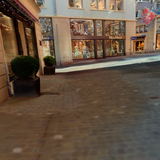}} &
        \fbox{\includegraphics[width=\sz\linewidth, height=\hw]{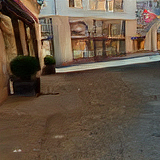}} &
        \fbox{\includegraphics[width=\sz\linewidth, height=\hw]{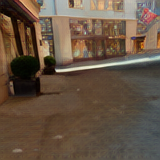}} &
        \fbox{\includegraphics[width=\sz\linewidth, height=\hw]{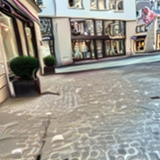}} &
        \fbox{\includegraphics[width=\sz\linewidth, height=\hw]{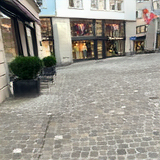}} &
        \fbox{\includegraphics[width=\sz\linewidth, height=\hw]{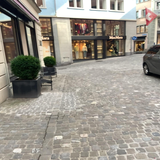}} 
        \\
        \fbox{\includegraphics[width=\sz\linewidth, height=\hw]{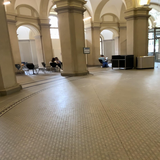}} &
        \fbox{\includegraphics[width=\sz\linewidth, height=\hw]{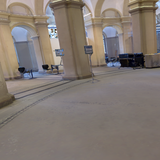}} &
        \fbox{\includegraphics[width=\sz\linewidth, height=\hw]{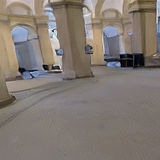}} &
        \fbox{\includegraphics[width=\sz\linewidth, height=\hw]{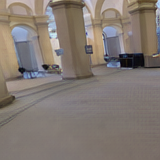}} &
        \fbox{\includegraphics[width=\sz\linewidth, height=\hw]{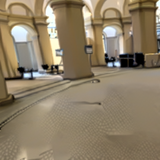}} &
        \fbox{\includegraphics[width=\sz\linewidth, height=\hw]{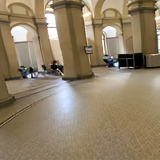}} &
        \fbox{\includegraphics[width=\sz\linewidth, height=\hw]{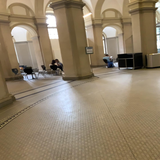}} 
        \\
        \fbox{\includegraphics[width=\sz\linewidth, height=\hw]{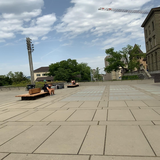}} &
        \fbox{\includegraphics[width=\sz\linewidth, height=\hw]{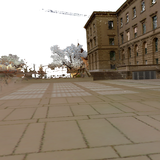}} &
        \fbox{\includegraphics[width=\sz\linewidth, height=\hw]{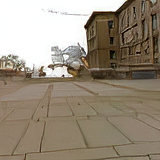}} &
        \fbox{\includegraphics[width=\sz\linewidth, height=\hw]{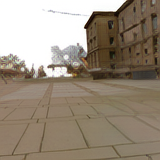}} &
        \fbox{\includegraphics[width=\sz\linewidth, height=\hw]{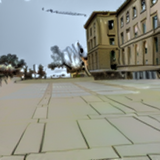}} &
        \fbox{\includegraphics[width=\sz\linewidth, height=\hw]{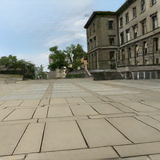}} &
        \fbox{\includegraphics[width=\sz\linewidth, height=\hw]{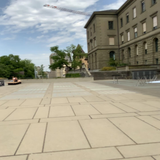}} 
        \\ %
        
    \end{tabular}}
    \caption{\textbf{Qualitative comparison -- 1.} Additional results of comparing MASA~\cite{MASASR}, DATSR~\cite{DATSR} (RefSR) and NNST~\cite{NNST} (ST) with \textbf{MaRINeR} on the task of novel view enhancement.}
\verticalspacetable
\verticalspacetable
  \label{fig:qualitative}
\end{figure}

\begin{figure}[!t]
  \centering
    \newcommand{\sz}{0.305}
    \renewcommand\fbox{\fcolorbox{black}{black}}
    \setlength{\fboxsep}{0.4pt} 
    \setlength{\fboxrule}{0.4pt}
    \resizebox{0.9\textwidth}{!}{
    \begin{tabular}{cccc}
        \rotatebox[origin=c]{90}{\tiny Render vs GT} &       
        \makecell{\fbox{\includegraphics[width=\sz\linewidth]{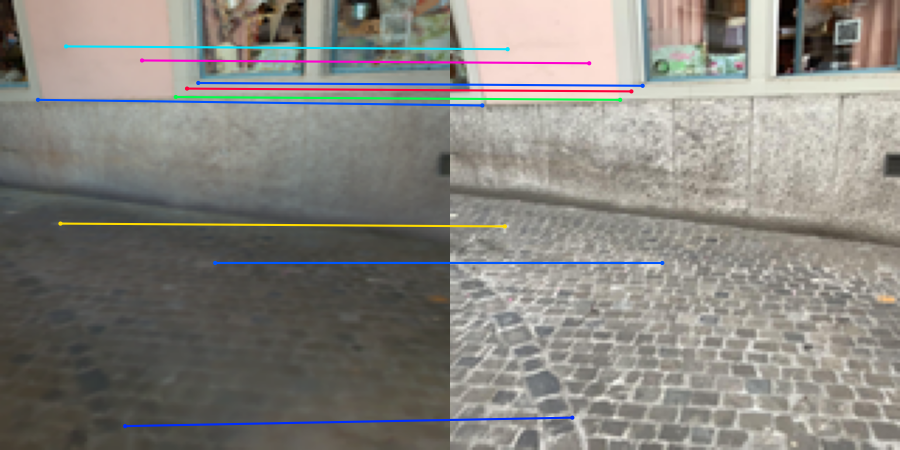}}} &
        \makecell{\fbox{\includegraphics[width=\sz\linewidth]{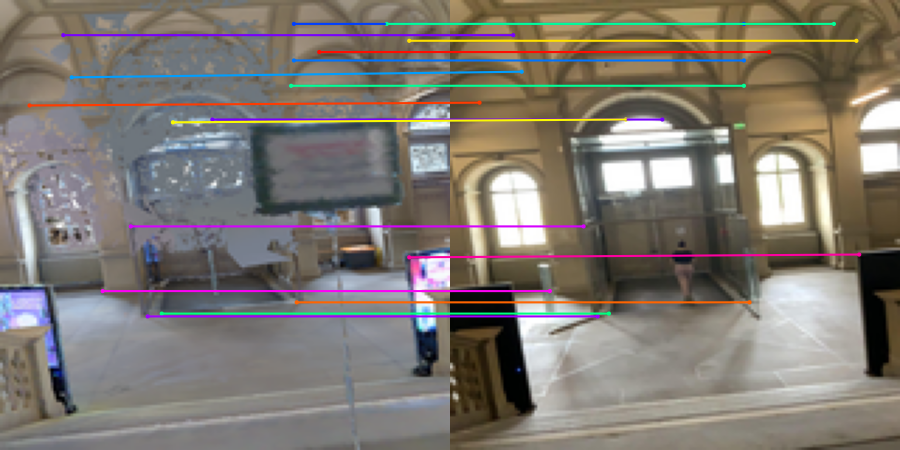}}} &
        \makecell{\fbox{\includegraphics[width=\sz\linewidth]{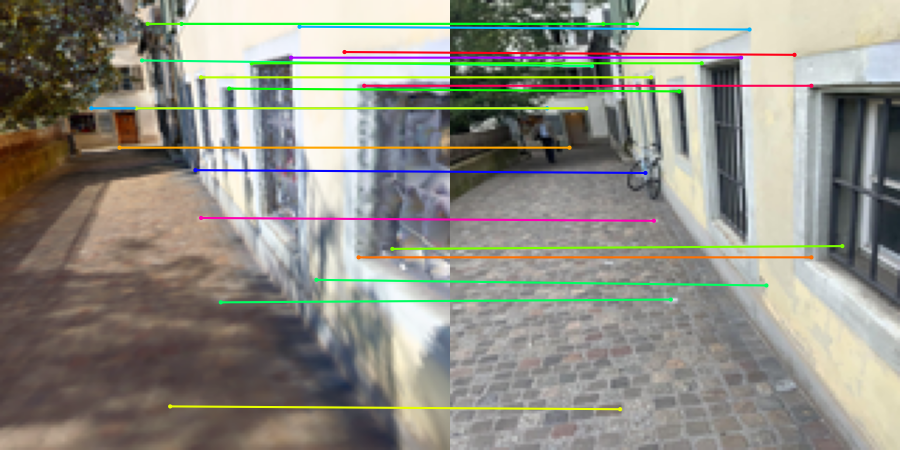}}} 
        \\
        \rotatebox[origin=c]{90}{\tiny MaRINeR vs GT} &
        \makecell{\fbox{\includegraphics[width=\sz\linewidth]{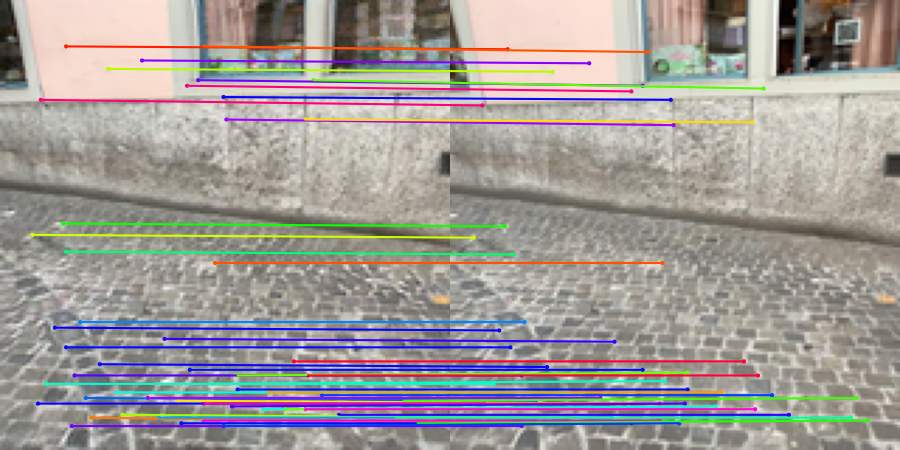}}} &
        \makecell{\fbox{\includegraphics[width=\sz\linewidth]{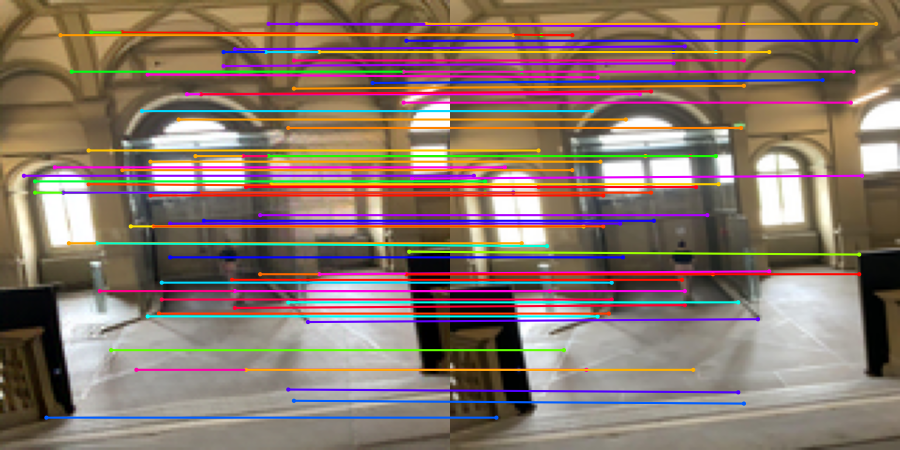}}} &
        \makecell{\fbox{\includegraphics[width=\sz\linewidth]{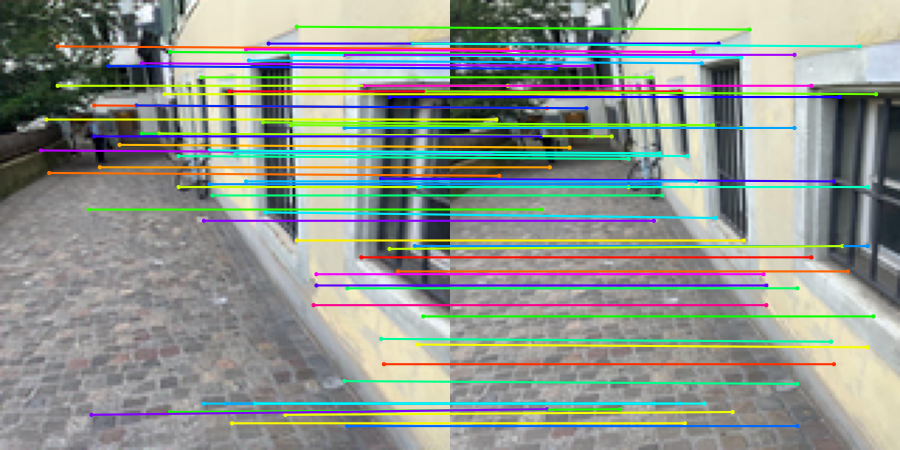}}} 
        \\
        \rotatebox[origin=c]{90}{\tiny Render vs GT} &       
        \makecell{\fbox{\includegraphics[width=\sz\linewidth]{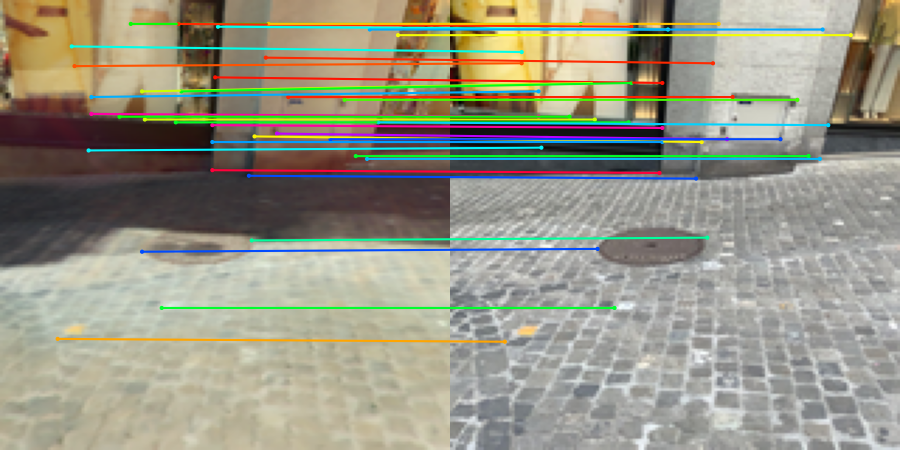}}} &
        \makecell{\fbox{\includegraphics[width=\sz\linewidth]{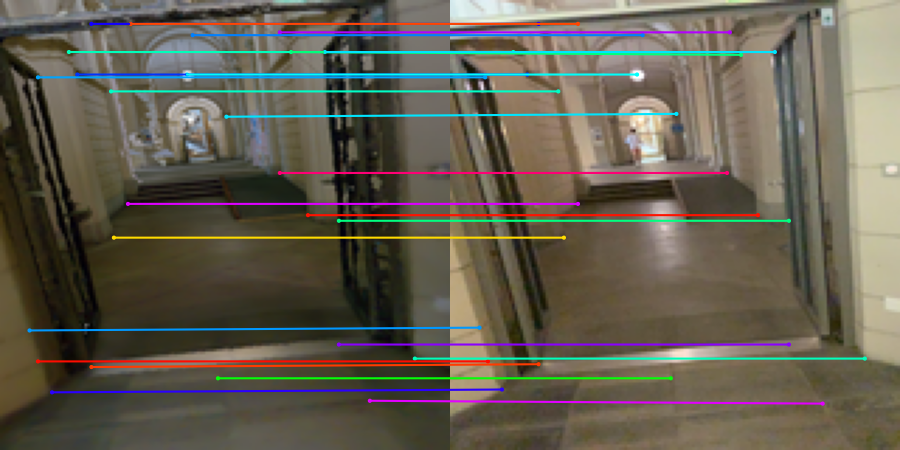}}} &
        \makecell{\fbox{\includegraphics[width=\sz\linewidth]{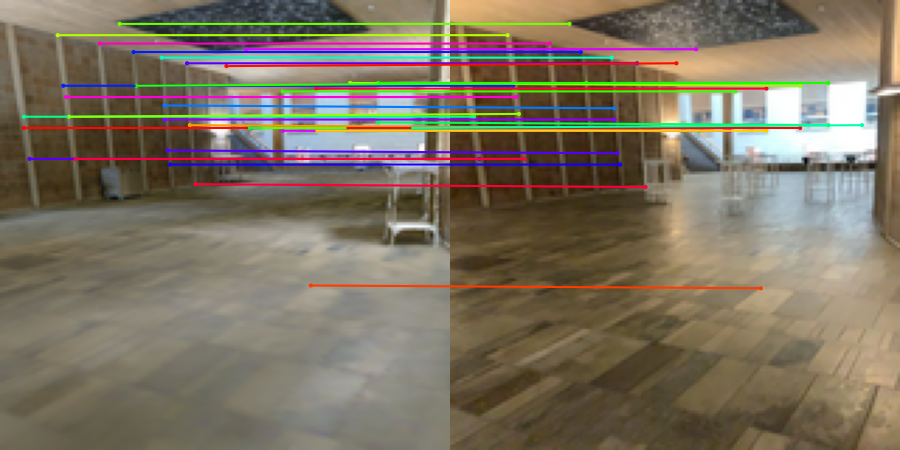}}} 
        \\
        \rotatebox[origin=c]{90}{\tiny MaRINeR vs GT} &
        \makecell{\fbox{\includegraphics[width=\sz\linewidth]{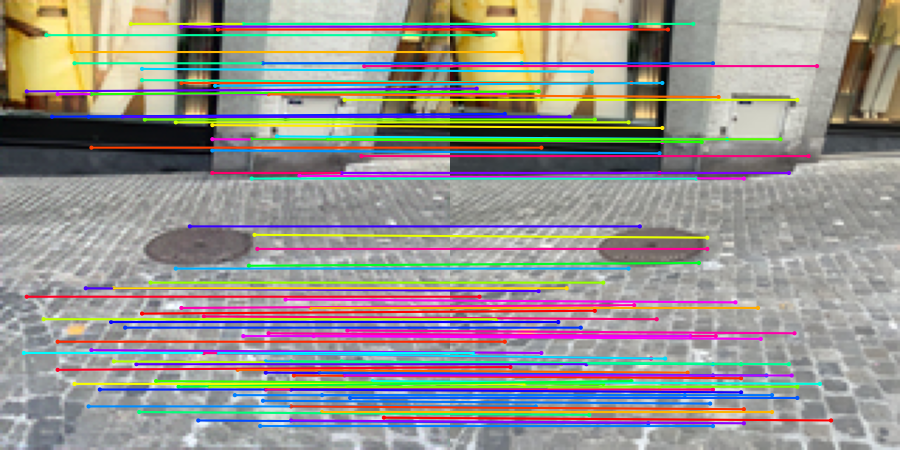}}} &
        \makecell{\fbox{\includegraphics[width=\sz\linewidth]{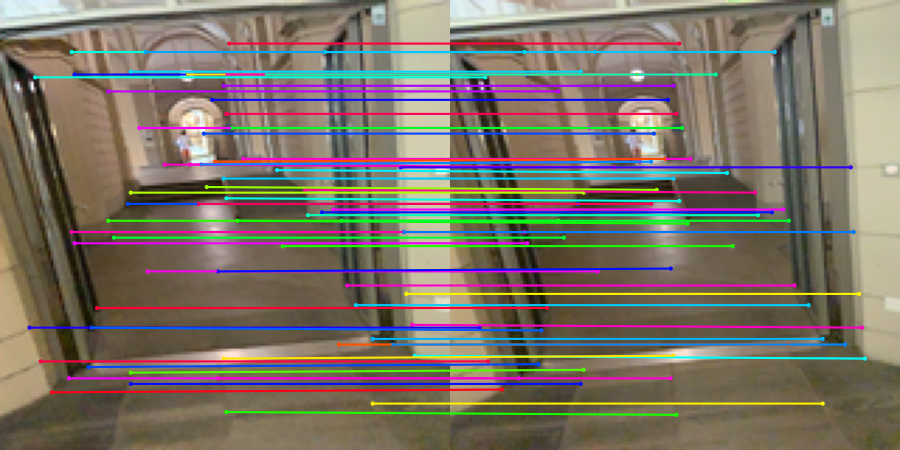}}} &
        \makecell{\fbox{\includegraphics[width=\sz\linewidth]{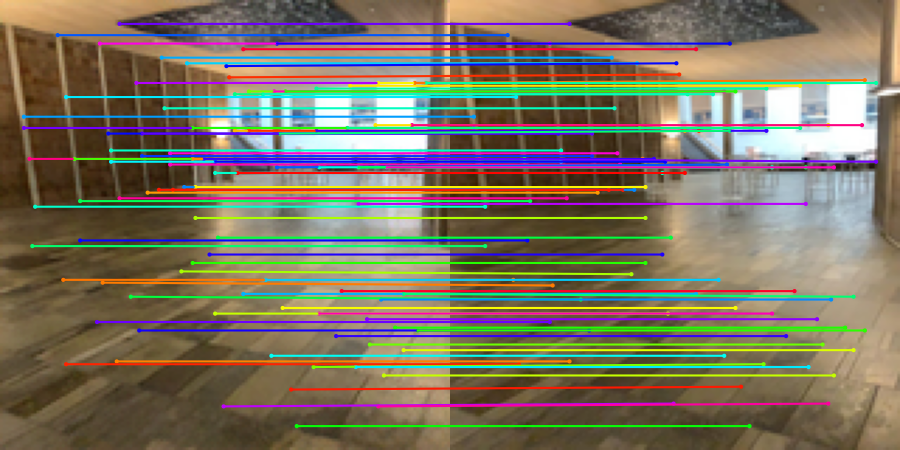}}} 
        \\
        
    \end{tabular}}
    \caption{\textbf{Further homography estimation results.} Using enhanced renderings of MARINeR, estimating a homography to the aligned source image is more accurate and can be used to automate manual sanity checks in the LaMAR~\cite{LaMAR} pipeline.}
    \label{fig:homography_app}
\end{figure}

\begin{figure}[!b]
    \vspace{-23pt}
    \newcommand{\sz}{0.155}
    \renewcommand\fbox{\fcolorbox{black}{black}}
    \setlength{\fboxsep}{0.4pt} 
    \setlength{\fboxrule}{0.4pt}
    \centering
    \hspace*{-10pt}
    \resizebox{0.9\textwidth}{!}{
    \begin{tabular}{ccccccc}
        & \scriptsize t=0 & \scriptsize t=1 & \scriptsize t=2 & \scriptsize t=3 & \scriptsize t=4 & \scriptsize t=5
        \\
        \scriptsize \rotatebox[origin=c]{90}{Synthetic} &
        \makecell{\fbox{\includegraphics[width=\sz\linewidth]{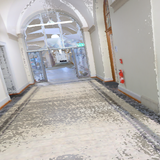}}}  &
        \makecell{\fbox{\includegraphics[width=\sz\linewidth]{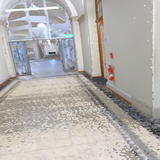}}}  &
        \makecell{\fbox{\includegraphics[width=\sz\linewidth]{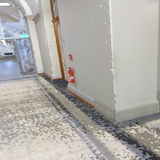}}}  &
        \makecell{\fbox{\includegraphics[width=\sz\linewidth]{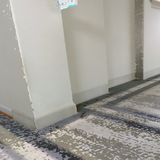}}}  &
        \makecell{\fbox{\includegraphics[width=\sz\linewidth]{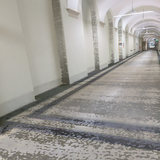}}}  &
        \makecell{\fbox{\includegraphics[width=\sz\linewidth]{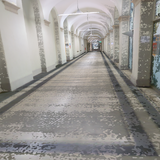}}} 
        \\
        \scriptsize \rotatebox[origin=c]{90}{Enhanced} &
        \makecell{\fbox{\includegraphics[width=\sz\linewidth]{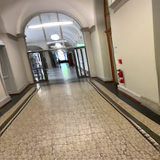}}}  &
        \makecell{\fbox{\includegraphics[width=\sz\linewidth]{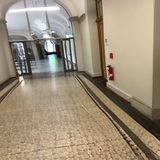}}}  &
        \makecell{\fbox{\includegraphics[width=\sz\linewidth]{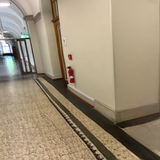}}}  &
        \makecell{\fbox{\includegraphics[width=\sz\linewidth]{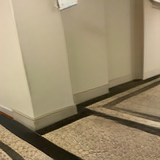}}}  &
        \makecell{\fbox{\includegraphics[width=\sz\linewidth]{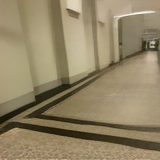}}}  &
        \makecell{\fbox{\includegraphics[width=\sz\linewidth]{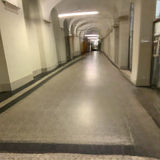}}}     
        \\ 
        \scriptsize \rotatebox[origin=c]{90}{Synthetic} &
        \makecell{\fbox{\includegraphics[width=\sz\linewidth]{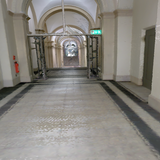}}}  &
        \makecell{\fbox{\includegraphics[width=\sz\linewidth]{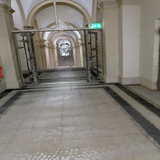}}}  &
        \makecell{\fbox{\includegraphics[width=\sz\linewidth]{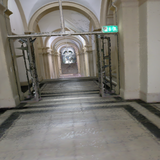}}}  &
        \makecell{\fbox{\includegraphics[width=\sz\linewidth]{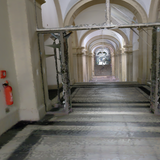}}}  &
        \makecell{\fbox{\includegraphics[width=\sz\linewidth]{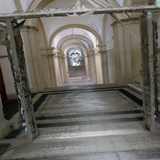}}}  &
        \makecell{\fbox{\includegraphics[width=\sz\linewidth]{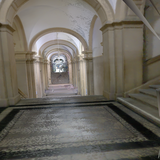}}} 
        \\
        \scriptsize \rotatebox[origin=c]{90}{Enhanced} &
        \makecell{\fbox{\includegraphics[width=\sz\linewidth]{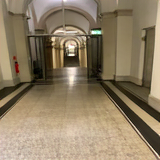}}}  &
        \makecell{\fbox{\includegraphics[width=\sz\linewidth]{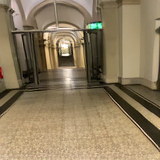}}}  &
        \makecell{\fbox{\includegraphics[width=\sz\linewidth]{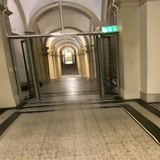}}}  &
        \makecell{\fbox{\includegraphics[width=\sz\linewidth]{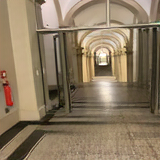}}}  &
        \makecell{\fbox{\includegraphics[width=\sz\linewidth]{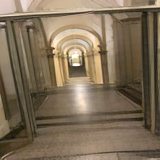}}}  &
        \makecell{\fbox{\includegraphics[width=\sz\linewidth]{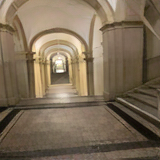}}}     
        \\
        
    \end{tabular}}
    \vspace{-3pt}
    \caption{\textbf{Enhancing synthetic trajectories with nearby localized images.} Additional results showing that because of the increased realism, the results from \textbf{MaRINeR} can extend the current dataset without introducing a gap between synthetic and human recorded trajectories.}
    \label{fig:domain_gap}\
\end{figure}

\section{Additional ablation studies}
\label{sec:Ablation}
Further ablation studies are performed on the influence of the perceptual and adversarial loss weights and the impact of different encoders and decoders. Finally, we provide further results extending the data-augmentation and iterative refinement ablations.
\begin{figure}[tb]
  \centering
  \begin{subfigure}{0.45\linewidth}
    \includegraphics[height=3.5cm]{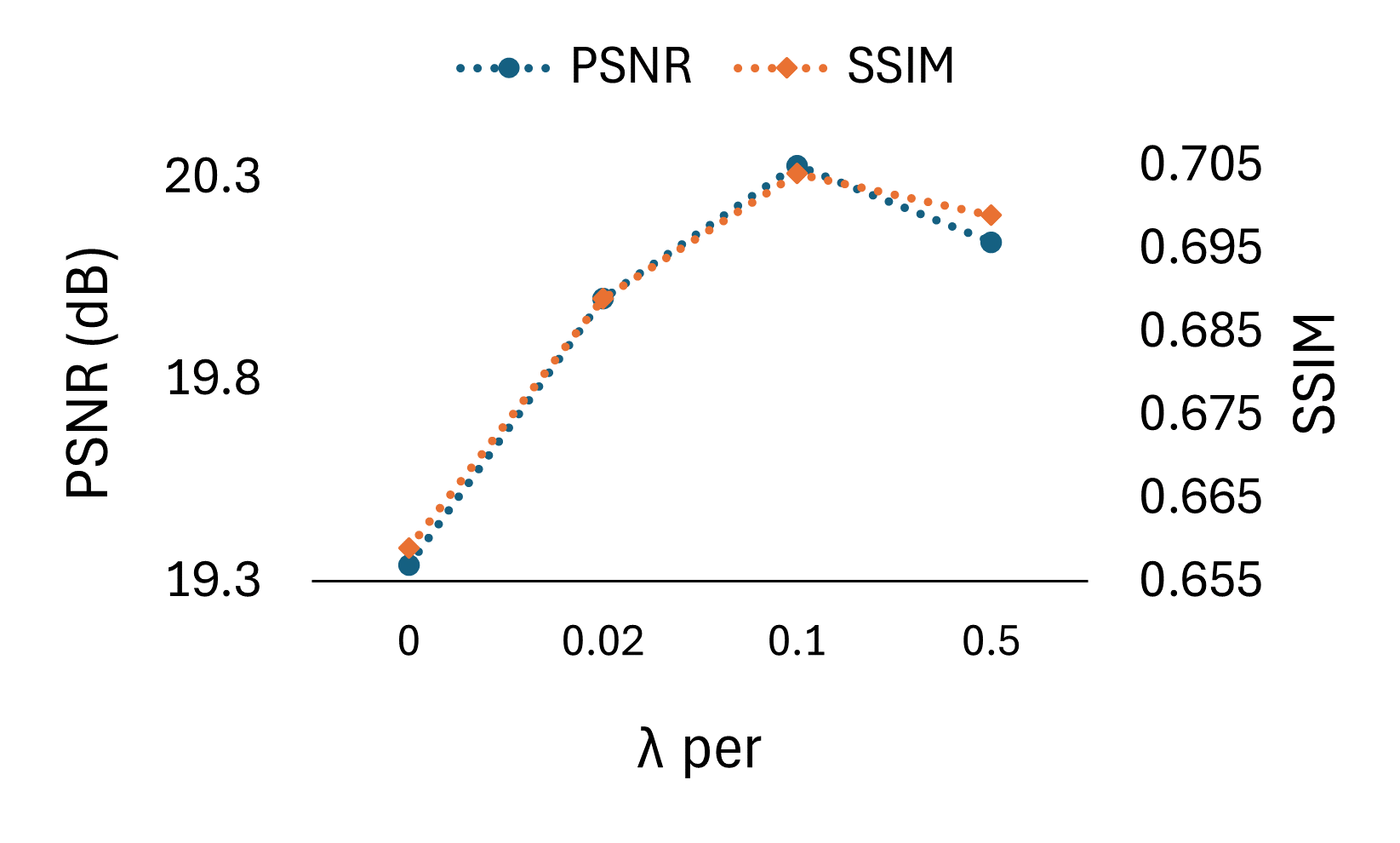}
    \caption{Perceptual loss weight influence}
    \label{fig:perc_loss_diag}
  \end{subfigure}  
  \hfill
\begin{subfigure}{0.45\linewidth}
    \includegraphics[height=3.5cm]{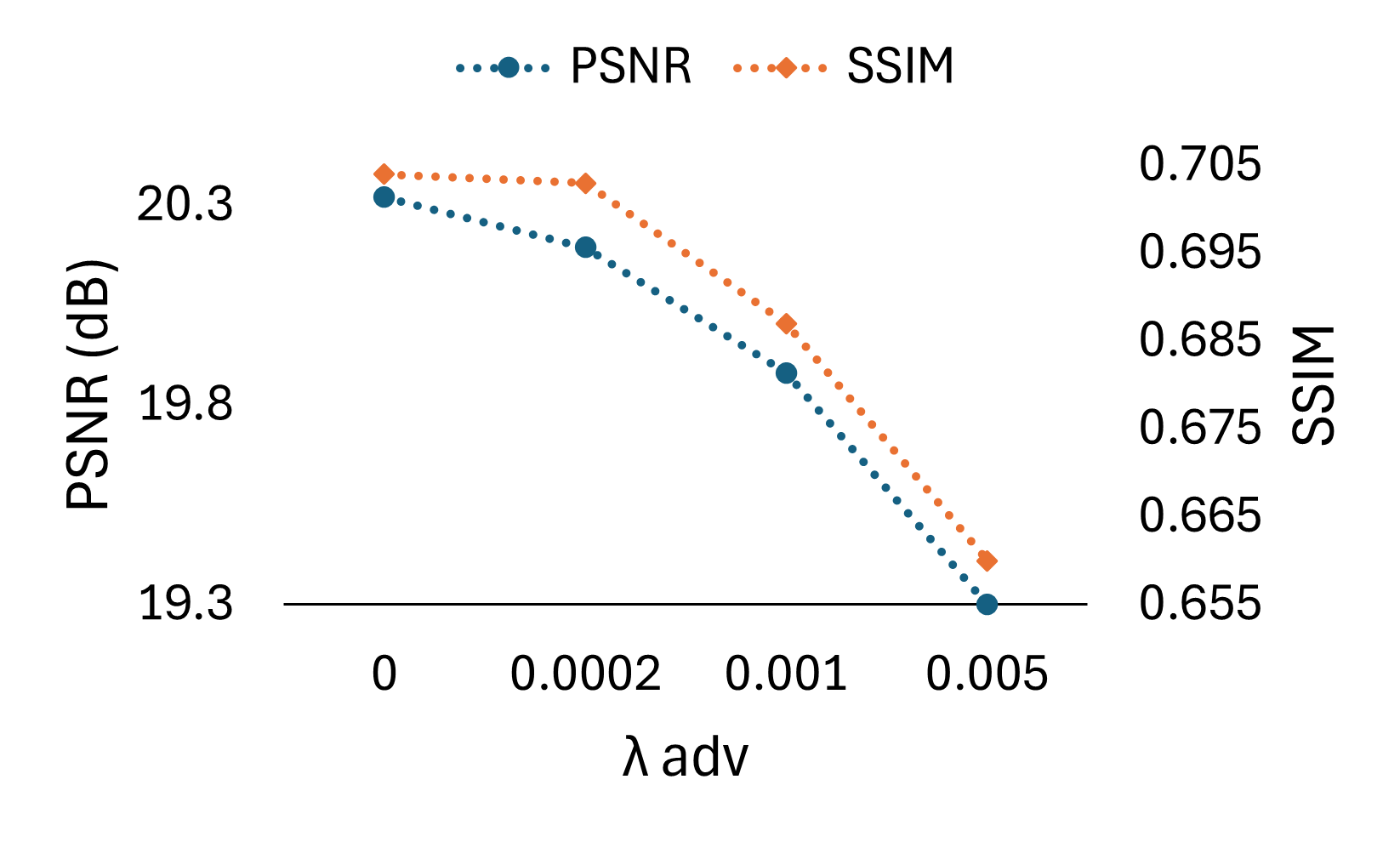}
    \caption{Adversarial loss weight influence}
    \label{fig:adv_loss_diags}
  \end{subfigure}
  \caption{\textbf{Influence of the loss weights.} The results of our experiments finding the optimal weights for a the perceptual loss and b the adversarial loss.}
  \label{fig:loss_diags}
 \verticalspace
\end{figure}
\begin{figure}[!htb]
\centering
    \newcommand{\hw}{56pt}
    \renewcommand\fbox{\fcolorbox{red}{red}}
    \setlength{\fboxsep}{0.4pt} 
    \setlength{\fboxrule}{0.4pt}
    \begin{tabular}{cccccc}
        Rendering & $\lambda_{\text{per}} = 0$ & $ \lambda_{\text{per}} = 0.02$  & \makecell[c]{\begin{tikzpicture} 
        \path[use as bounding box, draw=inchworm, thick] (0, 0) rectangle (1.7,0.4) node[pos=0.5] {$ \lambda_{\text{per}} = 0.1$}; 
        \end{tikzpicture}} & $ \lambda_{\text{per}} = 0.5$ & $ \mathcal{L}_{\text{per}}^{\text{MASA}}$
        \\
        \includegraphics[width=\hw, height=\hw]{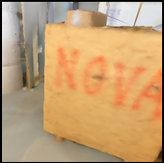} &
        \begin{tikzpicture}
        \node[anchor=south west,inner sep=0] at (0,0) {\includegraphics[width=\hw, height=\hw]{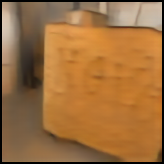}};
        \draw[red,ultra thick] (1.35,1.42) rectangle (0.65,0.7);
        \end{tikzpicture} &
        \begin{tikzpicture}
        \node[anchor=south west,inner sep=0] at (0,0) {\includegraphics[width=\hw, height=\hw]{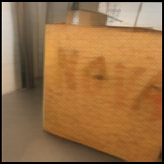}};
        \draw[red,ultra thick] (1.35,1.42) rectangle (0.65,0.7);
        \end{tikzpicture} &
        \begin{tikzpicture}
        \node[anchor=south west,inner sep=0] at (0,0) {\includegraphics[width=\hw, height=\hw]{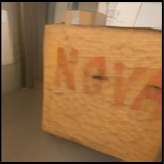}};
        \draw[red,ultra thick] (1.35,1.42) rectangle (0.65,0.7);
        \end{tikzpicture} &
        \begin{tikzpicture}
        \node[anchor=south west,inner sep=0] at (0,0) {\includegraphics[width=\hw, height=\hw]{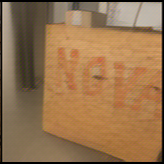}};
        \draw[red,ultra thick] (1.35,1.42) rectangle (0.65,0.7);
        \end{tikzpicture} &
        \begin{tikzpicture}
        \node[anchor=south west,inner sep=0] at (0,0) {\includegraphics[width=\hw, height=\hw]{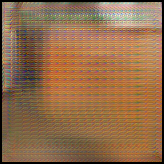}};
        \draw[red,ultra thick] (1.35,1.42) rectangle (0.65,0.7);
        \end{tikzpicture} 
        \\
        \rotatebox[origin=l]{90}{ Reference} \includegraphics[width=0.14\linewidth]{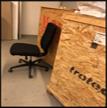} &
        
        \fbox{\includegraphics[width=54pt, height=\hw]{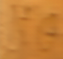}} &
        \fbox{\includegraphics[width=54pt, height=\hw]{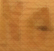}} &
        \fbox{\includegraphics[width=54pt, height=\hw]{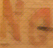}} &
        \fbox{\includegraphics[width=54pt, height=\hw]{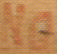}} &
        \fbox{\includegraphics[width=54pt, height=\hw]{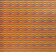}} 
        \\ %
        
    \end{tabular}
  \caption{\textbf{Impact of the perceptual loss.} Increased weight enhances details and the visual quality but also introduces perceptual loss specific grid-like artifacts.}
    \verticalspace
\label{fig:perc_loss_impact}
\end{figure}
\PAR{Influence of the perceptual loss.}
Because the task of novel view enhancement is different from RefSR, we investigate the effectiveness of the commonly used perceptual loss on our task. \cref{fig:perc_loss_impact} $\lambda_{\text{per}} = 0$ shows that without the perceptual loss, fine geometric structure like the texture of the box are not correctly transferred. Increasing the weight to $\lambda_{\text{per}} = 0.02$ and $\lambda_{\text{per}} = 0.1$, we observe increased texture details. A higher perceptual weight  $\lambda_{\text{per}} = 0.5$ leads to grid like artifacts~\cite{perceptualLossartifacts} which are more visible in image regions where the correspondence matching is less confident. The extreme case can be observed for $\lambda_{\text{per}}^{\text{MASA}}$ using the same perceptual loss as MASA-SR~\cite{MASASR}. \cref{fig:perc_loss_impact} shows that $\lambda_{\text{per}} = 0.1$ increases the details optimally while introducing minimal artifacts which is also confirmed numerically in \cref{fig:perc_loss_diag}.

\begin{figure}[!htb]
\centering
    \newcommand{\sz}{0.19}
    \renewcommand\fbox{\fcolorbox{red}{red}}
    \setlength{\fboxsep}{0.4pt} 
    \setlength{\fboxrule}{0.4pt}
    \begin{tabular}{ccccc}
        Rendering & $\lambda_{\text{adv}} = 0$ & $\lambda_{\text{adv}} = 0.0002$ & \makecell[c]{\begin{tikzpicture}
        \draw[draw=inchworm, thick] (0,0) rectangle (2,0.5) node[pos=.5] {$\lambda_{\text{adv}} = 0.001$}; 
        \end{tikzpicture}} & $\lambda_{\text{adv}} = 0.005$
        \\
        \includegraphics[width=67pt, height=67pt]{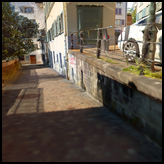} &
        \begin{tikzpicture}
        \node[anchor=south west,inner sep=0] at (0,0) {\includegraphics[width=67pt, height=67pt]{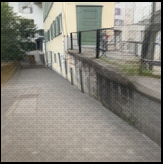}};
        \draw[red,ultra thick] (1.65,0.85) rectangle (0.8,0.05);
        \end{tikzpicture} &
        \begin{tikzpicture}
        \node[anchor=south west,inner sep=0] at (0,0) {\includegraphics[width=67pt, height=67pt]{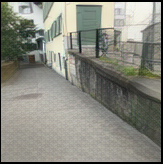}};
        \draw[red,ultra thick] (1.65,0.85) rectangle (0.8,0.05);
        \end{tikzpicture} &
        \begin{tikzpicture}
        \node[anchor=south west,inner sep=0] at (0,0) {\includegraphics[width=67pt, height=67pt]{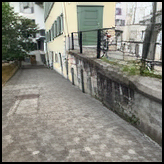}};
        \draw[red,ultra thick] (1.65,0.85) rectangle (0.8,0.05);
        \end{tikzpicture} &
        \begin{tikzpicture}
        \node[anchor=south west,inner sep=0] at (0,0) {\includegraphics[width=67pt, height=67pt]{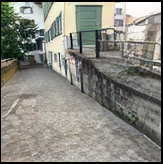}};
        \draw[red,ultra thick] (1.65,0.85) rectangle (0.8,0.05);
        \end{tikzpicture} 
        \\
        \rotatebox[origin=l]{90}{ Reference} \includegraphics[width=0.16\linewidth]{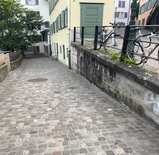} &
        
        \fbox{\includegraphics[width=65pt, height=67pt]{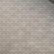}} &
        \fbox{\includegraphics[width=65pt, height=67pt]{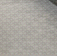}} &
        \fbox{\includegraphics[width=65pt, height=67pt]{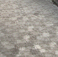}} &
        \fbox{\includegraphics[width=65pt, height=67pt]{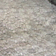}} 
        \\ [-5pt]
        
    \end{tabular}
  \caption{\textbf{Impact of the adversarial loss. }Increased weight removes the perceptual loss artifacts and keeps the underlying texture. Increasing the weight too much leads to the introduction of hallucinated details not present in the reference.}
\label{fig:adv_loss_impact}
  \vspace{-8pt}
\end{figure}

\begin{figure}[tb]
  \centering
  \includegraphics[width=0.5\textwidth]{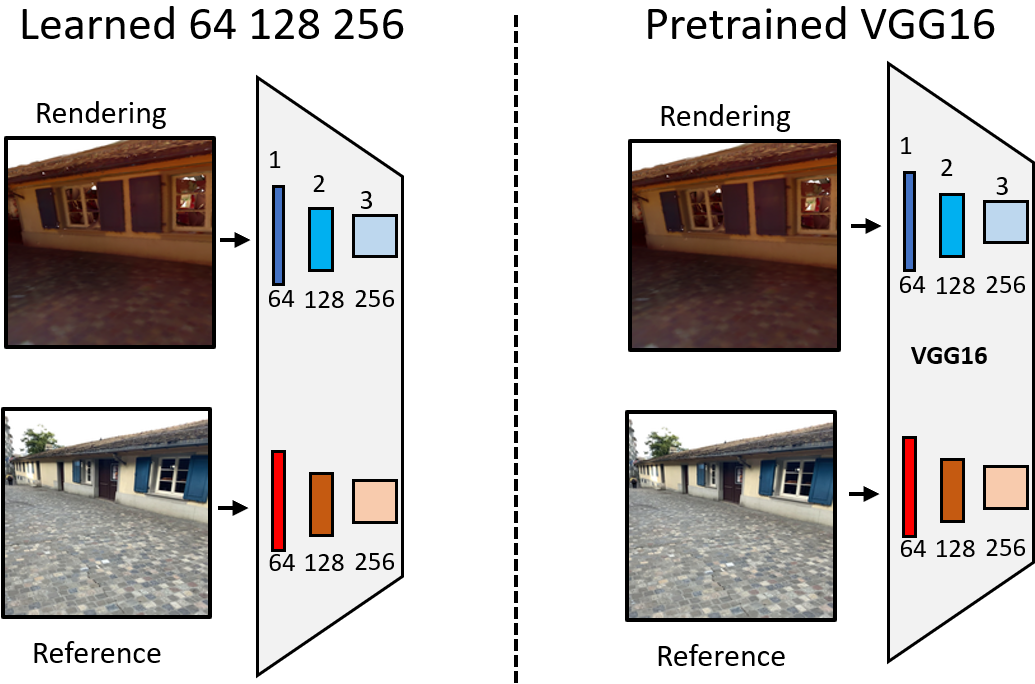}
  \caption{\textbf{Architecture of alternative encoders.} We validate the choice of our encoder by comparing it against an end-to-end trained encoder with larger feature channels and an encoder using pre-trained VGG16~\cite{VGG} features.}
\label{fig:encoders}
  \verticalspace
\end{figure}

\PAR{Influence of the adversarial loss.}
Using the perceptual loss can lead to grid-like artifacts~\cite{perceptualLossartifacts}. To remove those and make the images more visually pleasing~\cite{MASASR, DATSR} we use the adversarial loss. \cref{fig:adv_loss_impact} shows the impact of different weights for the loss. $\lambda_{\text{adv}} = 0$ contains the artifacts from the perceptual loss. $\lambda_{\text{adv}} = 0.005$ removes those artifacts completely but introduces high frequency details not present in the reference. \cref{fig:adv_loss_diags} shows that also the scores decrease with higher adversarial loss weight. We found that with $\lambda_{\text{adv}} = 0.001$ the perceptual loss artifacts are removed while minimal new details are wrongly introduced.
\begin{table}[tb]
 
  \caption{\textbf{Left - Encoders} Results of using different encoders. \textbf{Middle - Decoders} Results of replacing SAMs and DRAMs in the decoder. \textbf{Right - Inference time} Impact of the number of iterations on the inference time.}
  \centering
    \begin{minipage}{0.33\textwidth}
      \label{tab:encoder_ablation}
      \centering
      \scriptsize
      \begin{tabular}{ccc}
        \toprule
        \multirow{2}{*}{Encoder} & \multicolumn{2}{c}{CAB} \\ \cline{2-3}
        & PSNR & SSIM\\
        \midrule
          Learned 64 &  \fs{19.88} & \fs{0.687}\\
          Learned 64 128 256 &  19.66 & 0.685\\
          VGG &  19.08 & 0.653\\
      \bottomrule
      \end{tabular}
    \end{minipage} 
    \begin{minipage}{0.32\textwidth}
     \label{tab:decoder_ablation}
      \centering
      \scriptsize
      \hspace*{8pt}
      \begin{tabular}{ccc}
        \toprule
        \multirow{2}{*}{Decoder} & \multicolumn{2}{c}{CAB} \\ \cline{2-3}
        & PSNR & SSIM\\
        \midrule
         SAM + DRAM &  \fs{19.88} & \fs{0.687}\\
          No SAM &  19.73 & 0.685\\
          No DRAM &  19.71 & 0.676\\
      \bottomrule
      \end{tabular}
    \end{minipage}
     \begin{minipage}{0.32\textwidth}
      \label{tab:inference_time_ablation}
      \centering
      \scriptsize
      \begin{tabular}{cc}
        \toprule
         iterations \# & Runtime (ms) \\ 
        \midrule
        1 & 49.2\\
        2 & 66.3\\
        3 & 88.7\\
        4 & 110.5\\
      \bottomrule
      \end{tabular}
    \end{minipage}
 \verticalspace
\end{table}
\PAR{Encoder.}
An important part of the model performance is whether the matching between rendering and real image is successful. This matching is performed on the features from the encoder. MASA-SR~\cite{MASASR} uses features trained end-to-end with the super resolution task which has the advantage that the features are tailored for the task. Another option is to use pre-trained features~\cite{SRNTT, WTRNTIP}. If we use for example VGG features, we can leverage that those models were trained on a much larger dataset and the features potentially generalize better. \cref{fig:encoders} shows an overview over alternative encoders. The first encoder is trained end-to-end like ours but increases the feature dimension with each stage. The second one uses a pre-trained VGG16~\cite{VGG} encoder where we use the \texttt{relu1\_1}, \texttt{relu2\_2} and \texttt{relu3\_3} features. \cref{tab:encoder_ablation} validates the choice of the encoder used in our architecture.

\PAR{Decoder.}
We show that the SAM and DRAM blocks are also applicable for the task of novel view enhancement. For this we train two models, where in the first one the decoder has no SAMs. In the second model, the DRAMs are replaced by simply concatenating the features and merging them using a convolution. \cref{tab:decoder_ablation} shows that the scores are the best using both DRAMs and SAMs.
\begin{figure}[!htb]
  \centering
    \newcommand{\sz}{0.22}
    \renewcommand\fbox{\fcolorbox{black}{black}}
    \setlength{\fboxsep}{0.4pt} 
    \setlength{\fboxrule}{0.4pt}
    \hspace*{-12pt}
    \begin{tabular}{ccccc}
        & Reference & Rendering & No Augment. & Augment.
        \\
        \rotatebox{90}{\hspace*{28pt}level 1} &       
        \fbox{\includegraphics[width=\sz\linewidth]{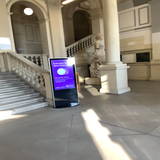}} &
        \fbox{\includegraphics[width=\sz\linewidth]{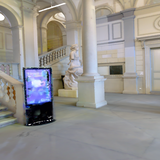}} &
        \fbox{\includegraphics[width=\sz\linewidth]{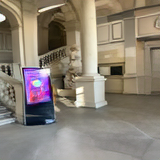}} &
        \fbox{\includegraphics[width=\sz\linewidth]{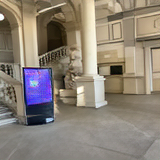}}
        \\
        \rotatebox{90}{\hspace*{28pt} level 7} &
        \fbox{\includegraphics[width=\sz\linewidth]{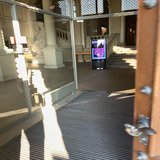}} &
        \fbox{\includegraphics[width=\sz\linewidth]{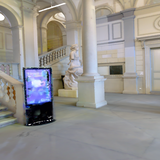}} &
        \fbox{\includegraphics[width=\sz\linewidth]{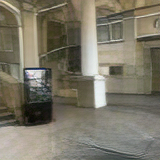}} &
        \fbox{\includegraphics[width=\sz\linewidth]{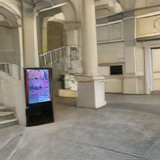}}
        \\
        
    \end{tabular}
  \caption{\textbf{Ref. level data augmentation.} Impact of using random close-by images as reference instead of only the closest one. While the model performs similar for ref. level 1, the correspondence matching fails for ref. level 7 leading to worse results.}
\label{fig:ref_level}
  \verticalspace
\end{figure}

\begin{figure}[!htb]
  \centering
    \newcommand{\sz}{0.22}
    \renewcommand\fbox{\fcolorbox{black}{black}}
    \setlength{\fboxsep}{0.4pt} 
    \setlength{\fboxrule}{0.4pt}
    \hspace*{-14pt}
    \begin{tabular}{ccccc}
        & Reference & Rendering & No Augment. & Augment.
        \\
        \makecell[c]{\rotatebox{90}{100\%}} &       
        \makecell[c]{\fbox{\includegraphics[width=\sz\linewidth]{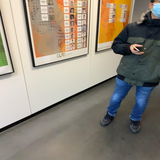}}} &
        \makecell[c]{\fbox{\includegraphics[width=\sz\linewidth]{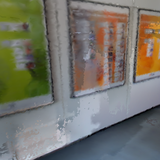}}} &
        \makecell[c]{\fbox{\includegraphics[width=\sz\linewidth]{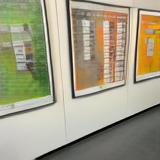}}} &
        \makecell[c]{\fbox{\includegraphics[width=\sz\linewidth]{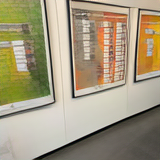}}}
        \\
        \makecell[c]{\rotatebox{90}{10\%}} &
        \makecell[c]{\fbox{\includegraphics[width=\sz\linewidth]{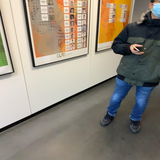}}} &
        \makecell[c]{\fbox{\includegraphics[width=\sz\linewidth]{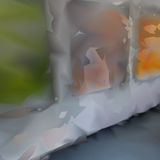}}} &
        \makecell[c]{\fbox{\includegraphics[width=\sz\linewidth]{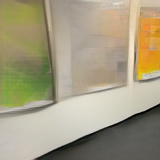}}} &
        \makecell[c]{\fbox{\includegraphics[width=\sz\linewidth]{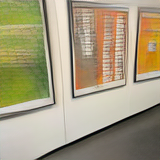}}}
        \\
        
    \end{tabular}
  \caption{\textbf{Mesh quality data-augmentation} Comparison of the model with and without augmenting the data with renderings from a down-sampled mesh. While for a mesh size of 100\% the results are visually similar, for the mesh size of 10\% the non augmented model fails to find correspondences and the result lacks the details from the reference.}
\label{fig:low_res_mesh_aug}
  \verticalspace
\end{figure}

\begin{figure}[!htb]
  \centering
    \newcommand{\sz}{0.16}
    \newcommand{\hw}{56.5pt}
    \renewcommand\fbox{\fcolorbox{black}{black}}
    \setlength{\fboxsep}{0.4pt} 
    \setlength{\fboxrule}{0.4pt}
    \hspace{-10pt}
    \resizebox{1.01\textwidth}{!}{
    \begin{tabular}{ccccccc}
        & Reference & Rendering & Result & Reference & Rendering & Result
        \\
        \rotatebox{90}{\hspace*{20pt} 75\%} &       
        \fbox{\includegraphics[width=\sz\linewidth, height=\hw]{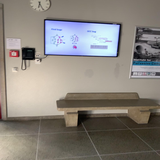}} &
        \fbox{\includegraphics[width=\sz\linewidth, height=\hw]{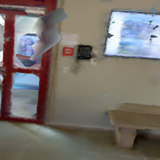}} &
        \fbox{\includegraphics[width=\sz\linewidth, height=\hw]{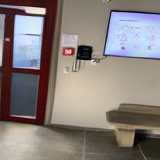}} &
        \hspace*{3pt}
        \fbox{\includegraphics[width=\sz\linewidth, height=\hw]{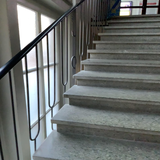}} &
        \fbox{\includegraphics[width=\sz\linewidth, height=\hw]{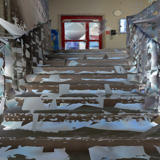}} &
        \fbox{\includegraphics[width=\sz\linewidth, height=\hw]{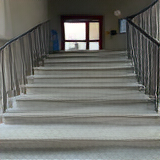}} 
        \\
        \rotatebox{90}{\hspace*{20pt} 50\%} &
        \fbox{\includegraphics[width=\sz\linewidth, height=\hw]{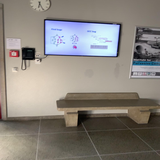}} &
        \fbox{\includegraphics[width=\sz\linewidth, height=\hw]{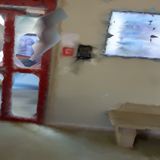}} &
        \fbox{\includegraphics[width=\sz\linewidth, height=\hw]{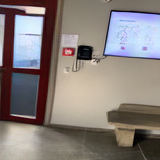}} &
        \hspace*{3pt}
        \fbox{\includegraphics[width=\sz\linewidth, height=\hw]{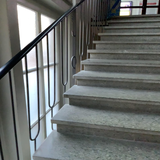}} &
        \fbox{\includegraphics[width=\sz\linewidth, height=\hw]{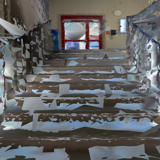}} &
        \fbox{\includegraphics[width=\sz\linewidth, height=\hw]{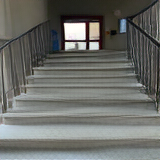}} 
        \\
        \rotatebox{90}{\hspace*{20pt} 25\%} &
        \fbox{\includegraphics[width=\sz\linewidth, height=\hw]{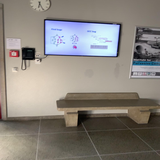}} &
        \fbox{\includegraphics[width=\sz\linewidth, height=\hw]{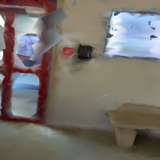}} &
        \fbox{\includegraphics[width=\sz\linewidth, height=\hw]{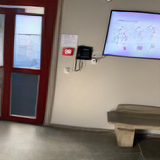}} &
        \hspace*{3pt}
        \fbox{\includegraphics[width=\sz\linewidth, height=\hw]{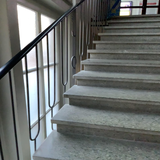}} &
        \fbox{\includegraphics[width=\sz\linewidth, height=\hw]{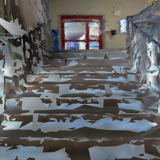}} &
        \fbox{\includegraphics[width=\sz\linewidth, height=\hw]{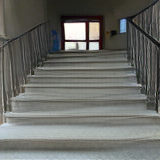}}
        \\
        
    \end{tabular}}
  \caption{ \textbf{Results of our model on low quality meshes.} The visual quality of the results stays high even if the mesh size is reduced to 75\% and 25\% of the original mesh triangles.}
  \label{fig:mesh_res}
  \verticalspace
\end{figure}

\PAR{Data augmentation.}
We show the visual impact of the random reference level data-augmentation in \cref{fig:ref_level}. The impact on the visual results of the mesh quality data-augmentation is shown in \cref{fig:low_res_mesh_aug}. This leads to increased robustness against meshes of various qualities, as  v   visualized in ~\cref{fig:mesh_res}.

\begin{figure}[tb]
  \centering
  \includegraphics[height=4cm]{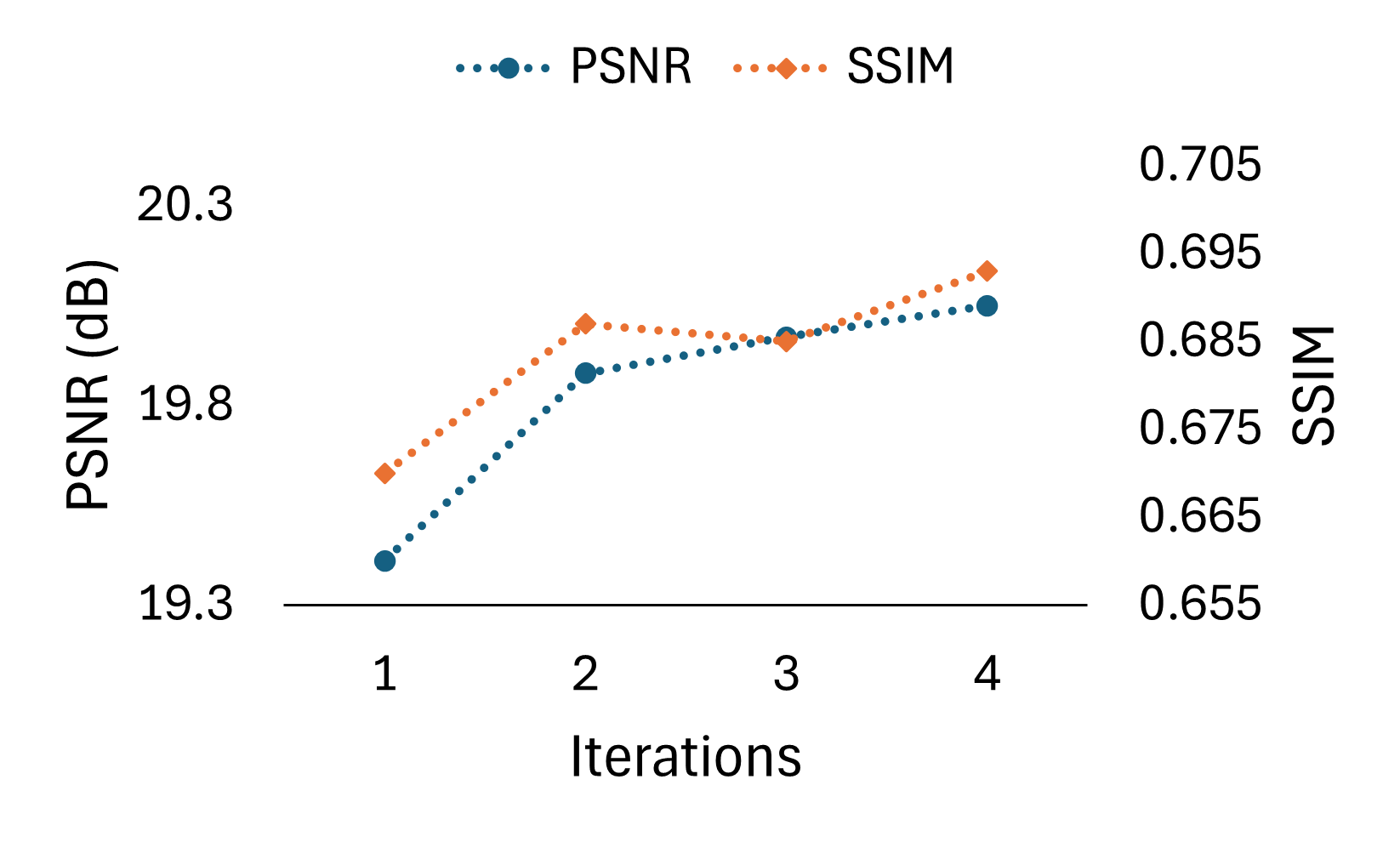}
  \caption{\textbf{Impact of the number of iterations.} Increasing the number of iterations leads to better results. The largest improvement can be seen between 1 and 2 iterations.}
\label{fig:diagram_iterations}
\verticalspace
\end{figure}

\PAR{Iterations.}
We show the effect on the PSNR and SSIM scores of different numbers of iterations in the iterative refinement process in \cref{fig:diagram_iterations}. The impact on the inference time is shown in \cref{tab:inference_time_ablation}.

\section{Metrics}
\label{sec:metrics}
\newcommand{\er}{\mathbf{I}_{\text{ER}}} 
\newcommand{\gt}{\mathbf{I}_{\text{GT}}} 
We provide the definitions of the metrics used to evaluate our model. The metrics are calculated between the ground truth image $\gt$ and the enhanced rendering $\er$.
\PAR{Peak Signal-to-Noise Ratio (PSNR).}
 The PSNR~\cite{PsnrSsim} is defined as

\begin{equation}
\begin{split}
        \text{PSNR}(\gt, \er) & = 10\text{log}_{10}\left(\frac{255^2}{\text{MSE}(\gt, \er)}\right) \\
        \text{MSE}(\gt, \er) &= \frac{1}{HW}\underset{i=1}{\overset{H}{\sum}}\underset{j=1}{\overset{W}{\sum}}(\gt(i, j) - \er(i, j))^2
\end{split}
\end{equation}
where the MSE is the mean squared error. 

\PAR{Structural Similarity Index Measure (SSIM).} 
The SSIM~\cite{SSIM} is calculated on two equally sized windows $x\subset\gt$ and $y\subset\er$

\begin{equation}
\begin{split}
        \text{SSIM}(x, y) & = \frac{(2\mu_x\mu_y + c_1)(2\sigma_{xy} + c_2)}{(\mu_x^2 + \mu_y^2 + c_1)(\sigma_x^2 + \sigma_y^2 + c_2)} \\
\end{split}
\end{equation}
where $c_1 = (0.01\cdot255)^2$ and $c_2 = (0.03 \cdot 255)^2$. The formula is based on three components that measure the difference between $x$ and $y$ in terms of luminace, constrast and structure.
\PAR{Edge Restoration Quality Assessment (ERQA).}
The ERQA~\cite{ERQA} finds edges in $\gt$ and $\er$ using the Canny algorithm~\cite{CannyAlgo}. Those edges are compared using the F1 score
\begin{equation}
        \text{precision}  = \frac{\text{TP}}{\text{TP} + \text{FP}}, \text{recall}  = \frac{\text{TP}}{\text{TP} + \text{FN}} \\
\end{equation}
\begin{equation}
        \text{F}_1  = 2\frac{\text{precision} \cdot \text{recall}}{\text{precision} + \text{recall}}
\end{equation}
where $\text{TP}$ (True Positive) are the number of pixels detected as edge in both $\gt$ and $\er$. $\text{FP}$ (False Positive) is the number of pixels detected only in $\er$, $\text{FN}$ (False Negative) are pixels only detected in $\gt$. To account for networks that produce small edge shifts either globally over the entire image or locally, ERQA builds in compensations to match the pixels of those edges before calculating the $F1$ score. 
\PAR{Learned Perceptual Image Patch Similarity (LPIPS).}
The LPIPS~\cite{LPIPS} uses deep neural networks as feature extractor and trains a similarity predictor network based on the feature difference of the images at several layers.
\begin{equation}
        \text{LPIPS}(\gt, \er)  = \underset{l}{\sum} \mathcal{G}_l\left( \frac{1}{H_l W_l}\underset{i}{\overset{H_l}{\sum}}\underset{j}{\overset{W_l}{\sum}}\|w_l \odot(\phi_l(\gt)_{i, j} - \phi_l(\er)_{i, j}\|_2^2)\right)
\end{equation}
where $\phi_l$ denotes the output of layer $l$ of the pretrained AlexNet~\cite{AlexNet}. LPIPS uses layers \texttt{conv\_1} to \texttt{conv\_5}. $\mathcal{G}_l$ is the trained prediction network for layer $l$, $\odot$ stands for scaling the activations channel-wise by a vector $w_l$.

\clearpage

\bibliographystyle{splncs04}
\bibliography{egbib}

\end{document}